\newif\ifdraft\draftfalse
\newif\iffull\fulltrue
\newif\iffinal\finaltrue

\iffinal

\else
\fi

\documentclass[runningheads]{llncs}
\usepackage{graphicx}
\usepackage{tikz}
\usepackage{comment}
\usepackage{amsmath,amssymb} % define this before the line numbering.
\usepackage{color}

\usepackage[accsupp]{axessibility}  % Improves PDF readability for those with disabilities.

\ifdefined\FinalVersion%
        \includecomment{FinalVersionBlock}
        \excludecomment{InitialVersionBlock}
\else
        \excludecomment{FinalVersionBlock}
        \includecomment{InitialVersionBlock}
\fi

\iffinal
\else
\begin{InitialVersionBlock}
  \usepackage{ruler}
  \usepackage[width=122mm,left=12mm,paperwidth=146mm,height=193mm,top=12mm,paperheight=217mm]{geometry} 
\end{InitialVersionBlock}
\fi

\usepackage{wrapfig}
\usepackage{booktabs}
\usepackage{xr}
\usepackage{subcaption}
\usepackage{paralist}
\usepackage{algorithm}
\usepackage{algorithmic}
\usetikzlibrary{arrows,automata,positioning,math,calc}

\usepackage[pagebackref,breaklinks,colorlinks]{hyperref}

\usepackage[capitalize]{cleveref}
\crefname{section}{Section}{Sections}
\Crefname{section}{Section}{Sections}
\Crefname{table}{Table}{Tables}
\crefname{table}{Table}{Tables}

\def\mkDraftFn#1#2{%
  \expandafter\def\csname #1\endcsname##1{\ifdraft\textcolor{#2}{[#1: ##1]}\marginpar[$\longrightarrow$]{$\longleftarrow$}\fi}%
}
\mkDraftFn{KS}{red}
\mkDraftFn{MW}{blue}

\usepackage{bm}

\usepackage{ifthen}
\newcommand{\colorR}[1]{\textcolor{red}{#1}}
\newcommand{\pagelimitmarker}[1]{~\\ {\colorR{\ifthenelse{\thepage>#1}{\Huge Exceeding the page limit}{\huge Within the page limit}}}~\\ {\huge{\colorR{~~Page Limit\,\,\,\,\, = #1}}}~\\ {\huge{\colorR{~~Current Page = $\thepage$}}}}

\newcommand\EXPECT{\mathbb{E}}
\newcommand\set[1]{\{{#1}\}}

\newcommand\REAL{\mathbb{R}}
\newcommand\PN{\mathrm{PN}}
\newcommand\POS{\mathrm{P}}
\newcommand\NEG{\mathrm{N}}
\newcommand\MASKDIST{\mathcal{M}}
\newcommand\flip[1]{\overline{{#1}}}

\renewcommand\vec[1]{\mathbf{{#1}}}
\newcommand\GP{\mathcal{GP}}

\DeclareMathOperator*{\argmax}{arg\,max}

\newcommand\SAL{\mathit{sal}}

\usepackage{url}	% required for `\url' (yatex added)

\newcommand{\imagewidth}{.6\linewidth}

\begin{document}
\pagestyle{headings}
\mainmatter%

\iffinal
\else
\def\ACCV22SubNumber{398}  % Insert your submission number here
\fi

\title{BOREx: Bayesian-Optimization--Based Refinement of Saliency Map for Image- and Video-Classification Models\thanks{We thank Atsushi Nakazawa for his fruitful comments on this work.
KS is partially supported by JST, CREST Grant Number JPMJCR2012, Japan.
MW is partially supported by JST, ACT-X Grant Number JPMJAX200U, Japan.
}}
\iffinal
\titlerunning{BOREx: Bayesian-Optimization--Based Refinement of Saliency Map}
\authorrunning{A. Kikuchi et al.}
\else
\titlerunning{ACCV-22 submission ID \ACCV22SubNumber}
\authorrunning{ACCV-22 submission ID \ACCV22SubNumber}
\fi

\iffinal
\author{Atsushi Kikuchi
  \and
  Kotaro Uchida
  \and
  Masaki Waga\orcidID{0000-0001-9360-7490}
  \and
  Kohei Suenaga\orcidID{0000-0002-7466-8789}}
\institute{Kyoto University}
\else
\author{Anonymous ACCV 2022 submission}
\institute{Paper ID \ACCV22SubNumber}
\fi

\maketitle

\begin{abstract}
  Explaining a classification result produced by an image- and video-classification model is one of the important but challenging issues in computer vision.
  Many methods have been proposed for producing heat-map--based explanations for this purpose, including ones based on the white-box approach that uses the internal information of a model (e.g., LRP, Grad-CAM, and Grad-CAM++) and ones based on the black-box approach that does not use any internal information (e.g., LIME, SHAP, and RISE).

  We propose a new black-box method \emph{BOREx} (\textbf{B}ayesian \textbf{O}ptimization for \textbf{R}efinement of visual model \textbf{Ex}planation) to refine a heat map produced by any method.
  Our observation is that a heat-map--based explanation can be seen as a prior for an explanation method based on Bayesian optimization.
  Based on this observation, BOREx conducts Gaussian process regression (GPR) to estimate the saliency of each pixel in a given image starting from the one produced by another explanation method.
  Our experiments statistically demonstrate that the refinement by BOREx improves low-quality heat maps for image- and video-classification results.
\end{abstract}
\section{Introduction}
\label{sec:introduction}

% \KS{Check missing references and citations.}
% \KS{Check well-formedness of the reference section.}
% \KS{Grammar and spell check.}
% \KS{Update the abstract in CMT.}

\begin{figure}[t]
 \begin{subfigure}[t]{.24\linewidth}
  \centering
  \includegraphics[trim=26 35 635 34,clip,width=.5\linewidth]{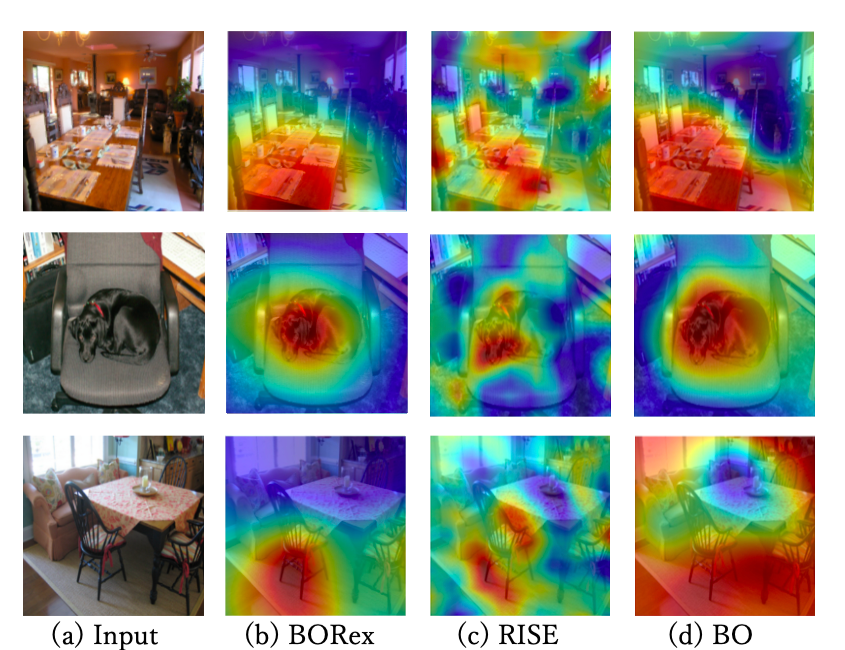}
  \caption{Input}%
  \label{fig:intro:saliency:input}
 \end{subfigure}
 \hfill
 \begin{subfigure}[t]{.24\linewidth}
  \centering
  \includegraphics[trim=228 35 433 34,clip,width=.5\linewidth]{image_square/top_image/top1.png}
  \caption{BOREx (Our method)}%
  \label{fig:intro:saliency:borex}
 \end{subfigure}
 \hfill
 \begin{subfigure}[t]{.24\linewidth}
  \centering
  \includegraphics[trim=432 35 228 34,clip,width=.5\linewidth]{image_square/top_image/top1.png}
  \caption{RISE}%
  \label{fig:intro:saliency:rise}
 \end{subfigure}
 \hfill
 \begin{subfigure}[t]{.24\linewidth}
  \centering
  \includegraphics[trim=636 35 24 34,clip,width=.5\linewidth]{image_square/top_image/top1.png}
  \caption{BO}%
  \label{fig:intro:saliency:bo}
 \end{subfigure}
  \caption{Example of the saliency maps generated by our method BOREx (in column (b)), RISE~\cite{DBLP:conf/bmvc/PetsiukDS18} (in column (c)), and the Bayesian-optimization-based method~\cite{DBLP:conf/ijcnn/MokuweBB20} (in column (d)); the input images to each method are presented in column (a).  The classification label used in the saliency maps in each row is ``dining table'', ``Labrador retriever'', and ``folding chair'' from the first row.}%
  \label{fig:intro:saliency}
\end{figure}
% \begin{figure}[t]
%   \includegraphics[width=0.48\textwidth]{image_square/top_image/top1.png}
%   \caption{Example of the saliency maps generated by our method BOREx (in column (b)), RISE~\cite{DBLP:conf/bmvc/PetsiukDS18}, and the method proposed by \cite{DBLP:conf/ijcnn/MokuweBB20} (in column (c)); the input images to each method are presented in column (a).  The classification label used in the saliency maps in each row is ``dining table'', ``labrador retriever'', and ``folding chair'' from the first row.}
%   \label{fig:intro:saliency}
% \end{figure}

Many image- and video-classification methods based on machine learning have been developed and are widely used. %~\cite{papersThatDemonstrateMachineLearningApplicationForImageAndVideoClassification}.
However, many of these methods (e.g., DNN-based ones) are not interpretable to humans.
The lack of interpretability is sometimes problematic in using an ML-based classifier under a safety-critical system such as autonomous driving. %~\cite{papersThatArgueThatBlackBoxModelIsHardToEmployInSafetyCriticalSystems}.

\sloppy
To address this problem, various methods to explain the result of image and video classification in the form of a heatmap called \emph{saliency map}~\cite{DBLP:conf/cvpr/ZhouKLOT16,DBLP:conf/bmvc/PetsiukDS18,DBLP:journals/ijcv/SelvarajuCDVPB20,DBLP:conf/wacv/ChattopadhyaySH18,DBLP:conf/icpr/ShiKLG20,DBLP:conf/accv/HatakeyamaSKS20,DBLP:conf/ijcnn/MokuweBB20} have been studied.
\cref{fig:intro:saliency} shows examples of saliency maps synthesized by several methods, including ours.
A saliency map for an image-classification result is an image of the same size as the input image.
Each pixel in the saliency map shows the contribution of the corresponding pixel in the input image to the classification result.
In each saliency map, the part that positively contributes to the classification result is shown in red, whereas the negatively-contributing parts are shown in blue.
The notion of saliency maps is extended to explain the results produced by a video-classification model, e.g., in~\cite{DBLP:conf/wacv/ChattopadhyaySH18} and~\cite{DBLP:conf/icip/StergiouKKCVP19}.

These saliency-map generation techniques can be classified into two groups: the \emph{white-box approach} and the \emph{black-box approach}.
A technique in the former group uses internal information (e.g., gradient computed inside DNN) to generate a saliency map; Grad-CAM~\cite{DBLP:journals/ijcv/SelvarajuCDVPB20} and Grad-CAM++~\cite{DBLP:conf/wacv/ChattopadhyaySH18} are representative examples of this group.
A technique in the latter group does not use internal information.
Instead, it repeatedly perturbs the input image by occluding several parts randomly and synthesizes a saliency map based on the change in the outputs of the model to the masked images from that of the original one.
The representative examples of this group are LIME~\cite{DBLP:conf/kdd/Ribeiro0G16}, SHAP~\cite{DBLP:conf/nips/LundbergL17}, and RISE~\cite{DBLP:conf/bmvc/PetsiukDS18}.

Although these methods provide valuable information to interpret many classification results,  the generated saliency maps sometimes do not correctly localize the regions that contribute to a classification result~\cite{DBLP:conf/eccv/BrunkeAG20,DBLP:conf/aaai/TomsettHCGP20,DBLP:conf/aaai/GhorbaniAZ19}.
Such a low-quality saliency map cannot be used to interpret a classification result correctly.

% ----------------------------------------------------------
%%%% KS: Old tikz picture: I revised this since the current explanation of BOREx is the "latter part" in the old picture.
% \begin{figure*}[tbp]
%  \centering
%  \begin{tikzpicture}[shorten >=1pt, scale=2, yscale=1, auto,rounded corners=1mm, node distance=3.5cm]
%   %% nodes
%   \node (Input) [align=center]{};
%   \node[rectangle,draw,node distance=2.5cm] (Initial) [right=of Input, align=center] {
%   Existing saliency map\\
%   generation method, \\
%   e.g., RISE~\cite{DBLP:conf/bmvc/PetsiukDS18} or\\
%   Grad-CAM~\cite{DBLP:journals/ijcv/SelvarajuCDVPB20}};
%   \node[rectangle,draw] (Bayesian) [right=of Initial, align=center]{Refinement of the\\ current saliency map $i_j$\\ by Bayesian optimization};
%   \node[node distance=2.5cm] (Result) [right=of Bayesian, align=center]{};

%   %% Edges
%   \path[->] 
%   (Input) edge node[align=center] {image (or video) $i$,\\ classifier $M$,\\ and label $l$} (Initial)
%   (Initial) edge node[align=center] {(Potentially low-quality)\\ saliency map $i_0$} (Bayesian)
%   (Bayesian) edge[loop above] node[align=center] {Refined saliency map $i_{j+1}$} (Bayesian)
%   (Bayesian) edge node[align=center] {Better-quality\\ saliency map $i_N$} (Result)
%   ;
%  \end{tikzpicture}
%  \caption{Our saliency map generation scheme via refinement. Starting from a potentially low-quality saliency map $i_0$ generated by an existing method, we refine the saliency map using Bayesian optimization and obtain a better-quality saliency map $i_N$.}%
%  \label{figure:refinement_scheme}
% \end{figure*}

Mokuwe et al.~\cite{DBLP:conf/ijcnn/MokuweBB20} recently proposed another black-box saliency map generation method using \emph{Bayesian optimization} based on the theory of \emph{Gaussian processes regression (GPR)}~\cite{gpml}.
Their method maintains (1) the estimated saliency value of each pixel and (2) the estimated variance of the saliency values during an execution of their procedure, assuming that a Gaussian process can approximate the saliency map; this assumption is indeed reasonable in many cases because a neighbor of an important pixel is often also important.
Using this information, their method iteratively generates the most effective mask to refine the estimations and observes the saliency value using the generated mask instead of randomly generating masks.
Then, the estimations are updated with the observation using the theory of Gaussian processes.

% Compared with the other black-box approaches such as RISE~\cite{DBLP:conf/bmvc/PetsiukDS18}, their approach can generate a meaningful saliency map with fewer masks by generating masks that are estimated to be most effective for refining the currently saliency estimation with the help of GPR.
% %
% However, the total time spent before a saliency map is generated is longer than RISE due to the computation required for GPR.

%----------------------------------------------------------

% One crucial observation concerning our method is the \emph{locality} of the salient pixels.
% Namely, \emph{a neighbor of an important pixel is often also important}.
% %
% Therefore, if the saliency values of several pixels are known, then one can estimate the saliency of the other part based on the known values.
%

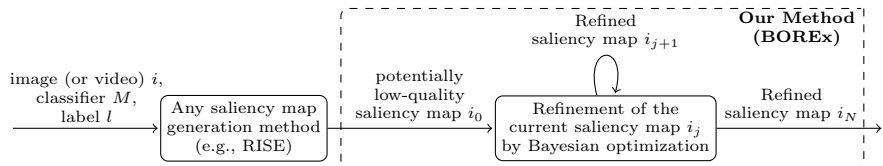
\begin{figure}[tbp]
 \centering
 \scriptsize
 \begin{tikzpicture}[shorten >=1pt, scale=0.89, auto,rounded corners=1mm, node distance=2.2cm,every node/.style={transform shape}]
  %% nodes
  % \node[rectangle,draw,node distance=2.5cm] (Initial) [right=of Input, align=center] {
  % Existing saliency map\\
  % generation method, \\
  % e.g., RISE~\cite{DBLP:conf/bmvc/PetsiukDS18} or\\
  % Grad-CAM~\cite{DBLP:journals/ijcv/SelvarajuCDVPB20}};
   \node[rectangle,draw] (Bayesian) [align=center]{Refinement of the\\ current saliency map $i_j$\\ by Bayesian optimization};
   \node[rectangle,draw,node distance=2.5cm] (RISE) [left=of Bayesian, align=center]{Any saliency map\\ generation method\\ (e.g., RISE)};
   \node (Input) [left=of RISE, align=center]{};
   \node[node distance=2.5cm] (Result) [right=of Bayesian, align=center]{};
   \draw[dashed] ($(Bayesian.north east) + (2.2,1.3)$)   --
                 ($(Bayesian.south east) + (2.2,-0.1)$)  --
                 ($(Bayesian.south west) + (-2.3,-0.1)$) --
                 ($(Bayesian.north west) + (-2.3,1.3)$)  --
                 ($(Bayesian.north east) + (2.2,1.3)$);
   \node at ($(Bayesian.north east) + (1.2,1.0)$)[align=center] {\textbf{Our Method}\\ \textbf{(BOREx)}};

  %% Edges
  \path[->] 
  (Input) edge node[align=center] {image (or video) $i$,\\ classifier $M$,\\ label $l$} (RISE) % (Initial)
  (RISE) edge node[align=center,pos=0.55] {potentially \\ low-quality\\ saliency map $i_0$} (Bayesian)
  % (Initial) edge node[align=center] {(Potentially low-quality)\\ saliency map $i_0$} (Bayesian)
  (Bayesian) edge[loop above] node[align=center] {Refined\\ saliency map $i_{j+1}$} (Bayesian)
  (Bayesian) edge node[align=center,pos=0.45] {Refined\\ saliency map $i_N$} (Result)
  ;
 \end{tikzpicture}
 \caption{Our saliency map generation scheme via refinement. Starting from a potentially low-quality saliency map $i_0$ generated by an existing method, we refine the saliency map using Bayesian optimization and obtain a better-quality saliency map $i_N$.}%
 \label{figure:refinement_scheme}
\end{figure}

% \KS{Editing this paragraph.}
Inspired by the method by Mokuwe et al., we propose a method to \emph{refine} the quality of a (potentially low-quality) saliency map.
Our idea is that the GPR-based optimization using a low-quality saliency map $i_0$ as prior can be seen as a procedure to iteratively refine $i_0$.
Furthermore, even if a saliency map $i_0$ generated by certain method is of low quality, it often captures the characteristic of the real saliency of the input image; therefore, using $i_0$ as prior is helpful to guide the optimization.
% it often captures the characteristic of the the real saliency of the input image; therefore, $i_0$ is helpful to guide the Bayesian optimization conducted in Mokuwe et al.~\cite{DBLP:conf/ijcnn/MokuweBB20}.
%

Based on this idea, we extend their approach so that it uses $i_0$ as prior information for their Bayesian optimization; see \cref{figure:refinement_scheme} for an overview of our saliency map generation scheme via refinement.
Our method can be applied to a saliency map $i_0$ generated by \emph{any} method; by the iterative refinement conducted by GPR, $i_0$ is refined to a better-quality saliency map as \cref{fig:intro:saliency} presents.
Each saliency map in \cref{fig:intro:saliency:borex} is generated by refining the one generated by RISE~\cite{DBLP:conf/bmvc/PetsiukDS18} presented in \cref{fig:intro:saliency:rise}; each saliency map in \cref{fig:intro:saliency} generated by our method localizes important parts better than that by RISE.\@

% black-box saliency-map generation method to take the best of Mokuwe et al.~\cite{DBLP:conf/ijcnn/MokuweBB20} and the other saliency-map generation method.
%
%
% Based on this idea, we extend the method by \cite{DBLP:conf/ijcnn/MokuweBB20} to use $i_0$ as the initial estimation of the saliency value of each pixel.
%

In addition to this extension, we improve their method to generate better saliency maps in a nontrivial way; these improvements include the way a saliency value is observed using a mask and the way a saliency map is generated from the final estimation of GPR.
With these extensions, our method \emph{BOREx} (\textbf{B}ayesian \textbf{O}ptimization for \textbf{R}efinement of visual model \textbf{Ex}planation) can generate better-quality saliency maps as presented in \cref{fig:intro:saliency}.

% BOREx can also be viewed as a method to \emph{refine} the quality of a (potentially low-quality) saliency map.
% %
% \emph{Any} saliency map generated by any method can be given as the prior $i_0$ in Figure~\ref{figure:refinement_scheme}; by the iterative refinement conducted by GPR, $i_0$ is refined to a better-quality saliency map as \cref{fig:intro:saliency} presents.
% %
% Each saliency map in \cref{fig:intro:saliency:borex} is generated by refining the one generated by RISE~\cite{DBLP:conf/bmvc/PetsiukDS18} presented in \cref{fig:intro:saliency:rise}; each saliency map in \cref{fig:intro:saliency} generated by our method localizes important parts better than that by RISE.\@

We also present an extension of BOREx to video-classification models.
Given a video-classification result, the resulting extension produces a video that indicates the saliency of each pixel in each frame using colors.
Combined with a naively extended RISE for video-classification models, BOREx can generate a saliency map for a video-classification result without using any internal information of the classification model.  % ; we are not aware of any other black-box--style explanation method for video-classification results at the time of writing.

We implemented BOREx and experimentally evaluated the effectiveness of BOREx.
The result confirms that BOREx effectively improves the quality of low-quality saliency maps, both for images and for videos, in terms of several standard metrics for evaluating saliency maps with statistical significance ($p < 0.001$).
We also conducted an ablation study, which demonstrates that the additional improvements to the method by Mokuwe et al.~\cite{DBLP:conf/ijcnn/MokuweBB20} mentioned above are paramount for this effectiveness.

% The experiments confirm the effectiveness of BOREx over the extension of RISE for video classifiers in terms of the standard metrics with statistical significance ($p < 0.001$).

Our contribution can be summarized as follows.
\begin{itemize}
\item
  We propose a new black-box method to refine a saliency map generated by any method.
  BOREx is an extension of the method by Mokuwe et al.~\cite{DBLP:conf/ijcnn/MokuweBB20} so that it uses a saliency map to be refined as prior in its Bayesian-optimization phase.
  % ; by this extension, BORex takes the best of Mokuwe et al.~\cite{DBLP:conf/ijcnn/MokuweBB20} and the other black-box saliency-map generation methods.
  %
  % BOREx can also be viewed as a method to \emph{refine} a potentially low-quality saliency map generated by any methods.
  %
  Besides the extension to take a saliency map as a prior, BOREx also enhances Mokuwe et al.~\cite{DBLP:conf/ijcnn/MokuweBB20} in several features, including how saliency values are evaluated using masks and how a saliency map is calculated from the final estimation obtained by the Bayesian optimization.
\item
  We present an extension of BOREx to explain video-classification results.
  The resulting extended BOREx produces a saliency map in the form of a video in a black-box manner.
  %
  % We are not aware of any other black-box explanation method for video-classification results at the time of writing.
\item
  We implemented BOREx and empirically evaluated its effectiveness.
  The experimental results statistically confirm the effectiveness of BOREx as a method for refining saliency-map--based explanation for image and video classifiers.
  We also conducted an ablation study, which demonstrates that the enhancement added to the method by Mokuwe et al.~\cite{DBLP:conf/ijcnn/MokuweBB20} is essential for the effectiveness.
  %
  % We also discuss the limitation of BOREx and potential extensions to enhance BOREx.
\end{itemize}

\paragraph{Related Work.}%
\label{section:related_work}
For both \emph{white-box} and \emph{black-box} approaches, various techniques have been proposed to explain a classification result of an image classifier by generating a saliency map.
% These techniques can be categorized into  techniques. 
The white-box approach exploits the internal information of the classifier, e.g., the network architecture and the parameters, and generates a saliency map, typically without using the inference result. 
Zhou et al.~\cite{DBLP:conf/cvpr/ZhouKLOT16} introduce \emph{class activation maps (CAM)} that generate a saliency map exploiting the global average pooling layer in the classification model.
Grad-CAM~\cite{DBLP:journals/ijcv/SelvarajuCDVPB20} and Grad-CAM++~\cite{DBLP:conf/wacv/ChattopadhyaySH18} generalize CAM by focusing on the gradient during back propagation to relax the requirements on the architecture of the classification model.
Zoom-CAM~\cite{DBLP:conf/icpr/ShiKLG20} is a variant of Grad-CAM that utilizes the feature map of the intermediate convolutional layers as well as the last convolutional layer.
Although these techniques are efficient since an inference is not necessary, gradient-based methods do not always generate a faithful explanation because the inference result is ignored in a saliency-map generation~\cite{DBLP:conf/nips/AdebayoGMGHK18,DBLP:conf/nips/HeoJM19,DBLP:conf/nips/DombrowskiAAAMK19,DBLP:conf/iccv/SubramanyaPP19}.

In contrast, the black-box approach treats a classifier as a black-box function without using its internal information.
These techniques typically perturb the given image and explain the classifier utilizing the difference in inference results between the original and the perturbed images.
For example, RISE~\cite{DBLP:conf/bmvc/PetsiukDS18} and PN-RISE~\cite{DBLP:conf/accv/HatakeyamaSKS20} randomly generate a mask by the Monte-Carlo method and perturb the image by occluding the pixels using the mask.
%
%% Mokuwe et al.~\cite{DBLP:conf/ijcnn/MokuweBB20} propose a saliency map generation technique with masks generated by the Gaussian process; see \cref{sec:background} for detail.
%
% These techniques utilize the prediction result in the saliency map generation we expect the resulting saliency maps to be faithful to the behavior of the classifier.
Although these techniques can be applied to a model whose internal information is not available, it requires many inferences to obtain a high-quality saliency map.

As shown in \cref{figure:refinement_scheme}, our technique, saliency map refinement by Bayesian optimization, requires an initial saliency map $i_0$ generated by an explanation technique mentioned above and refines it to improve its quality.
Thus, our technique allows combining one of the techniques above and the Bayesian optimization to balance various tradeoffs.
Typically, one can balance the tradeoff between the number of inferences and quality by feeding a saliency map that is not necessarily of high quality but requires less number of inferences.

Saliency-based explanation methods have also been investigated for video classifiers.
Stergiou et al.~\cite{DBLP:conf/icip/StergiouKKCVP19} propose an explanation of a 3D CNN model for video classification by generating a saliency \emph{tube} that is a 3D generalization of a saliency map.
They use the white-box approach based on the idea of CAM~\cite{DBLP:conf/cvpr/ZhouKLOT16}.
Chattopadhyay et al.~\cite{DBLP:conf/wacv/ChattopadhyaySH18} show that Grad-CAM++ outperforms in the explanation of a 3D CNN model for action recognition compared to Grad-CAM.\@
Bargal et al.~\cite{DBLP:conf/cvpr/BargalZKZMS18} propose an explanation technique for recurrent neural networks (RNNs) with convolutional layers utilizing excitation backpropagation~\cite{DBLP:journals/ijcv/ZhangBLBSS18}.
Perturbation-based black-box approaches have also been investigated to explain a video classifier by presenting salient frames~\cite{DBLP:conf/accv/PriceD20} or a 3D generalization of a saliency map~\cite{DBLP:conf/wacv/LiWLHS21}.
Same as the explanation of image classifiers,
our technique allows combining the techniques above and the Bayesian optimization to balance various tradeoffs.

% Overall, all of these techniques are white-box, and our work is the first explanation technique for video classifiers utilizing the black-box approach to the best of our knowledge.

The rest of the paper is organized as follows.
\Cref{sec:background} defines saliency maps and reviews the saliency-map generation method by Mokuwe et al.~\cite{DBLP:conf/ijcnn/MokuweBB20};
\cref{section:borex} introduces BOREx and an extension for video classifiers;
\cref{sec:experiments} explains the experiments;
\cref{sec:conclusion} concludes.

% 定義とか記法
% \label{section:notation}
We write $\Lambda$ for a set of \emph{pixels}; we write $\lambda$ for an element of $\Lambda$.
An \emph{image} is a map from $\Lambda$ to $\mathbb{N}^3$; we write $i$ for an image and $\mathcal{I}$ for the set of images.
The value $i(\lambda)$ represents the RGB value of pixel $\lambda$ in image $i$.
%
% If $i$ is a video, then $X$ is set to $\mathbb{N} \times \mathbb{N}^3$; 
% , which represents the RGB value of each pixel; we write $i$ for an image and $\mathcal{I}$ for the set of images.
%
We write $\mathcal{L}$ for the finite set of \emph{labels}.
A \emph{classification model} is a function from $\mathcal{I}$ to a probability distribution over $\mathcal{L}$; we write $M$ for a model.
For a model $M$ and an image $i$, the distribution $M(i)$ represents the confidence of $M$ in classifying $i$ to each label.
We write $M(i,l)$ for the confidence of $M$ classifying $i$ to $l$.

% Let the size of the image $i$ be width $U$ and height $V$.

\section{Background}%
\label{sec:background}

\subsection{Saliency}
\label{sec:saliency}

% RISE randomly masks the input image and generates a heat map from the information about how it affects the prediction results.

% % RISE とはなにか（input, output）
% This section explains RISE~\cite{rise}, a method for explaining an image-classification result returned by a a black-box model.
% %
% RISE takes an image-classification model $M$, an image $i$, and a label $l$; it returns a saliency map as a result.
% %
% The saliency map expresses the contribution of each pixel in the decision of $M$ in classifying $i$ to $l$.
% %
% RISE generates a saliency map observing only the perturbation in the output added by perturbing an input image; it does not use the internal information of $M$.

% RISE is a black-box method that generates saliency maps using only the input and output of $M$ without accessing the internals of $M$ such as gradient information if $M$ is implemented by DNN.

% Notations like $M(i,l)$ and $i \odot m$.

% RISE の背後にあるアイデア，RISE における saliency の定義

% \paragraph{Definition in Prior Work.}

Petsiuk et al.~\cite{DBLP:conf/bmvc/PetsiukDS18} define the saliency of each part in an image $i$ based on the following idea: \emph{A part in $i$ is important for a model $M$ classifying $i$ as $l$ if the confidence remains high even the other part in $i$ is masked.}
This intuition is formulated as follows by using the notion of \emph{masks}.
A mask $m$ is a function $m\colon\Lambda \rightarrow \set{0,1}$ that expresses how the value of each pixel of an image $i$ is diminished; the value of pixel $\lambda$ in the masked image---written $i \odot m$---is obtained by occluding the pixel $\lambda$ if $m(\lambda) = 0$.
Then, given a model $M$, an image $i$, and a label $l$, the \emph{saliency} $S_{i,l}(\lambda)$ of pixel $\lambda$ in image $i$ in $M$ classifying $i$ to $l$ is defined as follows:
\begin{eqnarray}
  S_{i,l}(\lambda) := \EXPECT[M(i \odot m, l) \mid m(\lambda) = 1]\label{def:saliency:rise}.
  % \EXPECT_{m \sim \MASKDIST}[M(i \odot m, l) \mid m(\lambda) = 0].\nonumber
\end{eqnarray}
In the above definition and in the following, the expectation $\EXPECT[M(i \odot m, l) \mid m(\lambda) = 1]$ is taken over a given distribution $\MASKDIST$ of masks.
Notice that the above formula defines saliency only by the input-output relation of $M$.
We call $S_{i,l}$ a \emph{saliency map}.

In \eqref{def:saliency:rise}, $m$ is randomly taken from a distribution $\MASKDIST$ over masks that models the assumption on how a salient part tends to distribute in an image.
$\MASKDIST$ is typically designed so that it gives higher probabilities to a mask in which masked regions form lumps, rather than the one in which masked pixels are scattered around the image; this design reflects that if a pixel is salient in an image, then the neighborhoods of the pixel are often also salient.

The definition of saliency we use in this paper is the refinement of $S_{i,l}$ by Hatakeyama et al.~\cite{DBLP:conf/accv/HatakeyamaSKS20} so that it takes \emph{negative saliency} into account.
Concretely, their definition of saliency $S_{i,l}^{\PN}$ is as follows.
\begin{eqnarray}
  S_{i,l}^{\POS}(\lambda) &:=& S_{i,l}(\lambda). \label{def:saliency:pnrise:positive}\\
 S_{i,l}^{\NEG}(\lambda) &:=& \EXPECT[M(i \odot m, l) \mid m(\lambda) = 0]. \label{def:saliency:pnrise:negative}\\
  S_{i,l}^{\PN}(\lambda) &:=& S_{i,l}^{\POS}(\lambda) - S_{i,l}^{\NEG}(\lambda).\label{def:saliency:pnrise:pn}
\end{eqnarray}
Their saliency $S^{\PN}_{i,l}(\lambda)$ is defined as the difference between the positive saliency $S^{\POS}_{i,l}(\lambda)$ and the negative saliency $S^{\NEG}_{i,l}(\lambda)$.
The latter is the expected confidence $M(i \odot m, l)$ conditioned by $m(\lambda) = 0$; therefore, a pixel $\lambda$ is negatively salient if masking out $\lambda$ contributes to increasing confidence in classifying the image as $l$.
% The first term of \eqref{def:saliency:pnrise} is the same as \eqref{def:saliency:rise}.
% %
% The second term $\EXPECT[M(i \odot m, l) \mid m(\lambda) = 0]$ is high if occluding $\lambda$ contributes to $M$ classifying the image to $l$; therefore, $\lambda$ is negatively salient for the classification.
% %
Hatakeyama et al.~\cite{DBLP:conf/accv/HatakeyamaSKS20} show that the saliency of an irreverent pixel calculated by $S_{i,l}^{\PN}(\lambda)$ is close to $0$, making the generated saliency map easier to interpret.

Evaluating $S_{i,l}$ and $S_{i,l}^{\PN}$ requires exhausting all masks, which is prohibitively expensive.
Petsuik et al.~\cite{DBLP:conf/bmvc/PetsiukDS18} and Hatakeyama et al.~\cite{DBLP:conf/accv/HatakeyamaSKS20} propose a method to approximate these saliency values using the Monte-Carlo method.
Their implementations draw masks $\set{m_1,\dots,m_N}$ from $\MASKDIST$ and approximate $S_{i,l}$ and $S_{i,l}^{\PN}$ using the following formulas, which are derived from the definitions of $S_{i,l}$ and $S_{i,l}^{\PN}$~\cite{DBLP:conf/accv/HatakeyamaSKS20,DBLP:conf/bmvc/PetsiukDS18} where $p = P[m(\lambda) = 1]$:
\begin{eqnarray}
  S_{i,l}(\lambda) &\approx& \frac{1}{N} \sum_n \frac{m_n(\lambda)}{p} M(i \odot m_n,l) \label{eq:montecarlo:pnrise}\\
  S_{i,l}^{\PN}(\lambda) &\approx& \frac{1}{N} \sum_n \frac{m_n(\lambda) - p}{p(1-p)} M(i \odot m_n, l).
\end{eqnarray}

\subsection{Saliency Map Generation using Gaussian Process Regression}%
\label{sec:gp}

Mokuwe et al.~\cite{DBLP:conf/ijcnn/MokuweBB20} propose another approach to generate saliency maps for black-box classification models.
Their approach uses Bayesian optimization, in particular \emph{Gaussian process regression (GPR)}~\cite{gpml} for this purpose.
We summarize the theory of GPR and how it serves for saliency-map generation in this section; for a detailed exposition, see~\cite{gpml}.

In general, a Gaussian process is a set of random variables, any finite number of which constitute a joint Gaussian distribution.
In our context, Gaussian process is a distribution over functions; each $f$ drawn from a Gaussian process maps $(\lambda,\vec{r})$ to a saliency value $f(\lambda,\vec{r}) \in \REAL$, where $\vec{r} \in \REAL^p$ is a vector of auxiliary parameters for determining a mask.
% \MW{It is nice to explicitly write down $f$ and $r$ in BO (maybe as an instance of the general framework).}
The $\vec{r}$ expresses, for example, the position and the size of a generated mask.
A Gaussian process is completely determined by specifying (1) a mean function $\mu(\lambda,\vec{r})$ that maps a pixel $\lambda$ and mask parameters $\vec{r}$ to their expected value $\EXPECT[f(\lambda,\vec{r})]$ and (2) a covariance function $k((\lambda,\vec{r}),(\lambda',\vec{r'}))$ that maps $(\lambda,\vec{r})$ and $(\lambda',\vec{r'})$ to their covariance $\EXPECT[(f(\lambda,\vec{r}) - \mu(\lambda,\vec{r}))(f(\lambda',\vec{r'}) - \mu(\lambda',\vec{r'}))]$.
We write $\GP(\mu,k)$ for the Gaussian process with $\mu$ and $k$.

\begin{algorithm}[t]
  \begin{algorithmic}[1]
   \scriptsize
  \caption{GPR-based saliency-map generation~\cite{DBLP:conf/ijcnn/MokuweBB20}.  The function $k$ is used in Line~\ref{alg:gp:updateMu}, which is kept implicit there.}%
  \label{alg:gp}
  \REQUIRE{Model $M$; Image $i$; Label $l$; Function $k$; Upperbound of iterations $N$; Set of mask size $L := \set{r_1,\dots,r_q}$.}
  \ENSURE{Saliency map that explains the classification of $i$ to $l$ by $M$.}
  \STATE{$D \leftarrow []$}
  \STATE{Set $\mu(\lambda,r) \leftarrow 0$ for every pixel $\lambda$ and $r \in L$}\label{alg:gp:initializeMu}
  \STATE{$j \leftarrow 0$}
  
  \WHILE{$j < N$}

  \STATE{$(\lambda,r) \leftarrow \argmax u_{\mu,D}$}\label{alg:gp:choosePixel}
  \STATE{Set $m$ to a square mask whose center is $\lambda$, whose side length is $r$, and $m(\lambda') = 0$ if $\lambda'$ is in the square}\label{alg:gp:generateMask}
  \STATE{$s \leftarrow M(i, l) - M(i \odot m, l)$}\label{alg:gp:observeSaliency}
  \STATE{Add $(\lambda,s)$ at the end of $D$}
  \STATE{Update $\mu$ using Bayes' law}\label{alg:gp:updateMu}
  \STATE{$j \leftarrow j + 1$}
  
  \ENDWHILE{}
  \STATE{$i_{\SAL}(\lambda) \leftarrow \frac{1}{q} \sum_{i} \mu(\lambda,r_i)$ for every $\lambda$.}
  \RETURN{$i_{\SAL}$}
  
  \end{algorithmic}
\end{algorithm}

GPR is a method to use Gaussian processes for regression.
Suppose we observe the saliency at several points in an image as $D := \set{((\lambda_1,\vec{r_1}),s_1),\dots,((\lambda_n,\vec{r_n}),s_n)}$.
For an unseen $(\lambda,\vec{r})$, its saliency conditioned by $D$ is obtained as a Gaussian distribution whose mean and variance can be computed by $D$, $\mu$, and $k$.
Furthermore, once a new observation is obtained, the optimization procedure can update $\mu$ using the Bayes' law.
These properties allow Gaussian processes to explore new observations and predict the saliency at unseen points.

Using these properties of GPs, Mokuwe et al.~\cite{DBLP:conf/ijcnn/MokuweBB20} propose \cref{alg:gp} for saliency-map generation.
Their method models a saliency map as a Gaussian process with mean function $\mu$ and covariance function $k$.
Under this model, \cref{alg:gp} iteratively chooses $(\lambda,r)$ (Line~\ref{alg:gp:choosePixel}), observe the saliency evaluated with $(\lambda,r)$ by using a mask whose center is at $\lambda$ and with side length $r$ (Lines~\ref{alg:gp:generateMask} and~\ref{alg:gp:observeSaliency}), and update $\mu$ using Bayes' law (Line~\ref{alg:gp:updateMu}).
To detect the most positively salient part with a small number of inferences, \cref{alg:gp} uses an \emph{acquisition function} $u_{\mu,D}(\lambda,r)$.
This function is designed to evaluate to a larger value if (1) $|\mu(\lambda,r)|$ or (2) the expected variance of the saliency at $\lambda$ estimated from $D$ is high; therefore, choosing $\lambda$ and $r$ such that $u_{\mu,D}(\lambda,r)$ is large balances exploiting the current estimation of the saliency value $\mu(\lambda,r)$ and exploring pixels whose saliency values are uncertain.
To keep the search space reasonably small, we keep the shape of the generated masks simple; in \cref{alg:gp}, to a finite set of square masks.

Various functions that can be used as a covariance function $k$ have been proposed; see~\cite{gpml} for detail.
Mokuwe et al.~\cite{DBLP:conf/ijcnn/MokuweBB20} use \emph{Mat\'ern kernel}~\cite{gpml}.

\cref{alg:gp} returns the saliency map $i_{\SAL}$ by $i_{\SAL}(\lambda) := \frac{1}{q} \sum_{i} \mu(\lambda,r_i)$.
The value of $i_{\SAL}$ at $\lambda$ is the average of $\mu(\lambda,r)$ over $r \in L$.

% \subsection{Comparison}
% \label{sec:comparison}

% %
% The time spent for the Monte-Carlo method described in \cref{sec:saliency} is dominated by inference time; computation of the approximated saliency can be done efficiently.
% %
% Therefore, this method is fast if an inference by the given model is fast.
% %
% However, it requires many iterations to obtain a high-quality saliency map; therefore, this method is costly if an inference takes time.

% The merit and the demerit of the GPR-based saliency-map generation in \cref{sec:gp} is complementary to the Monte-Carlo--based method.
% %
% Each iteration of \cref{alg:gp} requires much time compared to one iteration of the Monte-Carlo method since it requires evaluating Equation~\ref{eq:gpUpdate}, which involves matrix operations.
% %
% However, a small number of iterations is enough to obtain a high-quality saliency map in many cases.

% BO の特徴（利点欠点）
% The BO method can generate a highly accurate saliency map with a small number of inferences because it uses optimization to sample intelligently.
% However, the BO method has the disadvantage that it takes time to perform optimization to determine the next mask.

\section{BOREx}
\label{section:borex}

\begin{algorithm}[t]
  \begin{algorithmic}[1]
   \scriptsize
  \caption{GPR-based refinement of a saliency map.}
  \label{alg:gpBOREx}
  \REQUIRE{Model $M$; Image $i$; Initial saliency map $i_0$; Label $l$; Function $k((\lambda,r),(\lambda',r')$; Upperbound of iterations $N$; List of the side length of a mask $L := \set{r_1,\dots,r_p}$.}
  \ENSURE{Refined saliency map obtained with GP.}
  \STATE{$D \leftarrow []$}
  \STATE{Set $\mu(\lambda,r) \leftarrow i_0(\lambda)$ for every pixel $\lambda$ and side size $r \in L$.}\label{alg:gpBOREx:initializeMu}
  \STATE{$j \leftarrow 0$}
  
  \WHILE{$j < N$}

  \STATE{$(\lambda,r) \leftarrow \argmax u_{\mu,D}$}\label{alg:gpBOREx:choosePixel}
  \STATE{Set $m$ to a square mask with side length $r$, whose center is $\lambda$, and $m(\lambda') = 0$ if $\lambda'$ is inside the rectangle}\label{alg:gpBOREx:generateMask}
  \STATE{$s \leftarrow M(i \odot \flip{m}, l) - M(i \odot m, l)$}\label{alg:gpBOREx:observeSaliency}
  \STATE{Add $((\lambda,r),s)$ at the end of $D$}
  \STATE{Update $\mu$ using Bayes' law}\label{alg:gpBOREx:updateMu}
  \STATE{$j \leftarrow j + 1$}
  
  \ENDWHILE
  \STATE{$i_{\SAL}(\lambda) \leftarrow \frac{1}{p} \sum_{i} \frac{1}{r_i^2} \mu(\lambda,r_i)$ for every $\lambda$.}\label{alg:gpBOREx:computeSaliency}
  \RETURN{$i_{\SAL}$}
  
  \end{algorithmic}
\end{algorithm}

\subsection{GPR-based Refinement of Saliency Map}
\label{sec:borex}

\cref{alg:gpBOREx} is the definition of BOREx.
The overall structure of the procedure is the same as that of \cref{alg:gp}.
The major differences are the following: (1) the input given to the procedures; (2) how the saliency is evaluated; and (3) how a saliency map is produced from the resulting $\mu$.
We explain each difference in the following.

\noindent\textbf{Input to the algorithm.}
\cref{alg:gpBOREx} takes the initial saliency map $i_0$, which is used as prior information for GPR.
Concretely, this $i_0$ is used to initialize $\mu(\lambda, r)$ in Line~\ref{alg:gpBOREx:initializeMu}.
To generate $i_0$, one can use \emph{any} saliency-map generation methods, including ones based on black-box approach~\cite{DBLP:conf/accv/HatakeyamaSKS20,DBLP:conf/bmvc/PetsiukDS18,DBLP:conf/kdd/Ribeiro0G16,DBLP:conf/nips/LundbergL17} and ones based on white-box approach~\cite{DBLP:journals/ijcv/SelvarajuCDVPB20,DBLP:conf/wacv/ChattopadhyaySH18}.

\noindent\textbf{Saliency evaluation.}
\cref{alg:gp} evaluates the saliency by calculating $M(i, l) - M(i \odot m, l)$.
This value corresponds to the value of $-S_{i,l}^{\NEG}$ around the pixel $\lambda$ defined in \cref{sec:saliency} since it computes how much the confidence \emph{drops} if a neighborhood of $\lambda$ is masked out.

To estimate $S_{i,l}^{\PN}$ instead of $-S_{i,l}^{\NEG}$, \cref{alg:gpBOREx} calculates $M(i \odot \flip{m}, l) - M(i \odot m, l)$ in Line~\ref{alg:gpBOREx:observeSaliency}, where $\flip{m}$ is the \emph{flipped} mask obtained by inverting the value at each pixel (i.e., $\flip{m}(\lambda') = 1 - m(\lambda')$ for any $\lambda'$).
Since $\flip{m}(\lambda) = 1$ if and only if $m(\lambda) = 0$, the value of $M(i \odot \flip{m}, l) - M(i \odot m, l)$ is expected to be close to $S_{i,l}^{\PN}(\lambda')$ if $\lambda'$ is near $\lambda$.

\begin{figure}[t]
 \begin{subfigure}[t]{.30\linewidth}
  \centering
  \includegraphics[trim=0 0 0 0,width=.5\linewidth,height=.5\linewidth,clip]{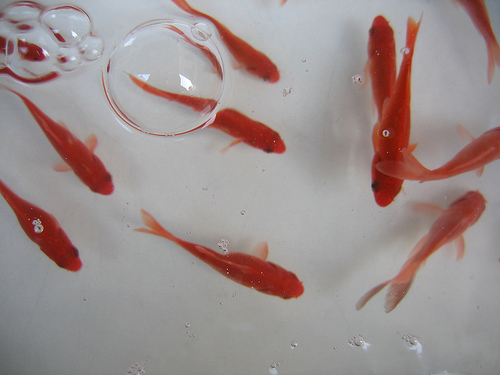} 
  \caption{Image of goldfish.}%
  \label{fig:multipleSalient:image}
 \end{subfigure}
 \hfill
 \begin{subfigure}[t]{.33\linewidth}
  \centering
  \includegraphics[trim=75 40 115 15,clip,width=.5\linewidth,height=.5\linewidth]{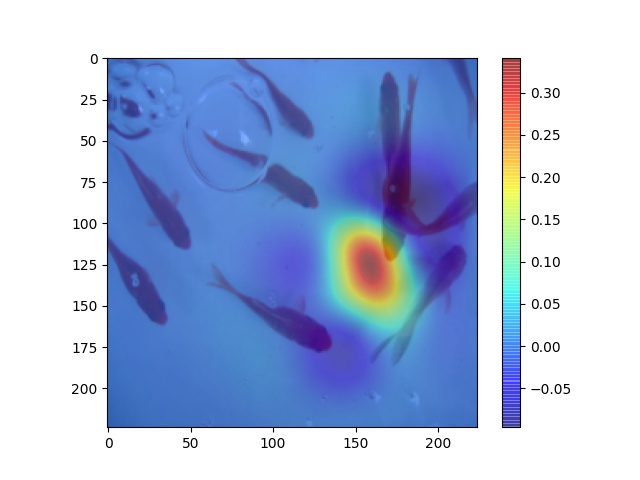} 
  \caption{Saliency map generated by \cref{alg:gp}.}%
  \label{fig:multipleSalient:bo}
 \end{subfigure}
 \hfill
 \begin{subfigure}[t]{.33 \linewidth}
  \centering
  \includegraphics[trim=75 40 115 15,width=.5\linewidth,height=.5\linewidth,clip]{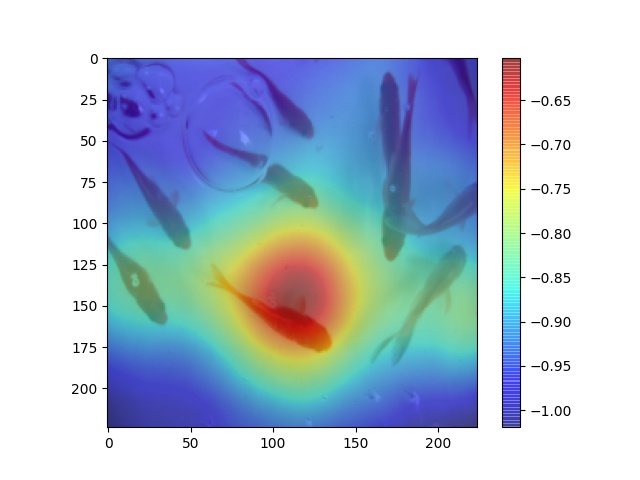}
  \caption{Saliency map generated by \cref{alg:gpBOREx}.}%
  \label{fig:multipleSalient:borex}
 \end{subfigure}
 \caption{Example of an image with multiple salient regions.}%
 \label{fig:multipleSalient}
\end{figure}

Another reason of using a flipped mask in the saliency observations of \cref{alg:gpBOREx} is to handle images in which there are multiple salient parts.
For example, the image of goldfish in \cref{fig:multipleSalient:image} has multiple salient regions, namely, multiple goldfish.
If we apply \cref{alg:gp}, which does not use flipped masks, to this image, we obtain the saliency map in \cref{fig:multipleSalient:bo}; obviously, the saliency map does not capture the salient parts in the image.
This is because the value of $M(i, l) - M(i \odot m, l)$ in Line~\ref{alg:gp:observeSaliency} of \cref{alg:gp} is almost same everywhere; this value becomes high for this image only if $m$ hides \emph{every} goldfish in the image, which is difficult using only a single mask.
Our method generates the saliency map in \cref{fig:multipleSalient:borex}; an observed saliency value $M(i \odot \flip{m}, l) - M(i \odot m, l)$ in \cref{alg:gpBOREx} is higher if $m$ hides \emph{at least} one goldfish than if $m$ does not hide any goldfish.
%
% This difference helps our method generate a better saliency map in our case.

\noindent\textbf{Generating saliency map from resulting $\mu$.}
\cref{alg:gpBOREx} returns the saliency map $i_{\SAL}$ defined by $i_{\SAL}(\lambda) = \frac{1}{p} \sum_{i} \frac{1}{r_i^2} \mu(\lambda,r_i)$.
Instead of the saliency map computed by taking the simple average over every mask in \cref{alg:gp}, the saliency map map returned by \cref{alg:gpBOREx} is the average weighted by the inverse of the area $\frac{1}{r_i^2}$ of each mask with the side size $r$.
This weighted average gives more weight to the saliency values obtained by smaller masks.
Using the weighted average helps a saliency map produced by \cref{alg:gpBOREx} localizes salient parts better than \cref{alg:gp}.

% ステップ２の大まかな説明
% In the second step, we refine the saliency map obtained in the first step with BO method. 
% Most of the methods we use are the same as those described in section \ref{BO chap}.
% But we've added some ingenuities.

% 具体的な説明
% 変数の定義
% Let $S_{i,l}(\lambda)$ be the saliency map obtained in the first step.
% We define $\mu(Q)$ and $k(Q, Q')$ as in Chapter \ref{BO chap}.
% We define the input to Gaussian Process model as $\bm{x} = (u, v, w, h)$, where $u$ is the position in the horizontal axis to which the mask is applied, $v$ is the position in the vertical axis to which the mask is applied, $w$ is the width of the mask, and $h$ is the height of the mask.
% Let $\mathcal{W} = \{w_1, .., w_W\}$ be the set of variable mask widths of length $W$ and $\mathcal{H} = \{h_1, .., h_H\}$ be the set of variable mask heights of length $H$.
% $Q$ is set of the observations and $Q = \{\bm{x_1},...,\bm{x_p}\}$, $p = U \times V \times W \times H$.

% 長方形のマスクを取り入れている点に触れる

%  inside mask と outside maskの説明

% 提案手法におけるガウス過程の出力

% inside mask と outside maskを使用している理由
% 対象オブジェクトが複数写ってる画像を載せる

% % その後の進め方は元BOと同じ
% The second and subsequent observation points are determined in the same way as in the original BO method.
% Past observed data $D=\{(\bm{x_{1:j}},y_{1:j})\}$ are provided to a Gaussian process and predicts the approximated saliency map for other unseen observations $Q$.
% Based on the results of this prediction, We search for the location and size of the mask that will reduce the confidence of classifying image $i$ into label $l$ the most.

% 最終的なサリエンシーマップの決め方
% マスクサイズを考慮した補正についても書く必要がある

\subsection{Extension for Video-Classification Models}
\label{sec:video}

\cref{alg:gpBOREx} can be naturally extended for a video classifier with the following changes.
\begin{itemize}
\item
  The set of masks is extended, from 2D squares specified by their side length, to 3D rectangles specified by the side length of the square in a frame, and the number of frames that they hide.
  Suppose a mask $m$ with side length $r$ and the number of frames $t$ is applied to the pixel $\lambda$ at coordinate $(x,y)$ and at $n$-th frame of a video $i$.
  Then, $i \odot m$ is obtained by hiding the pixel at $(x,y)$ in each of the $n$-th to $(n+t-1)$-th frame with the 2D square mask specified by $r$.
\item
  The type of functions drawn from the Gaussian process is changed to $f(\lambda,r,t)$ from $f(\lambda,r)$ in \cref{alg:gpBOREx} reflecting the change of the definition of masks.
\item
  The algorithm takes $T := \set{t_1,\dots,t_k}$ in addition to $L$; the set $T$ expresses the allowed variation of parameter $t$ of a mask.
\item
  The expression to update $i_{\SAL}$ in Line~\ref{alg:gpBOREx:computeSaliency} of \cref{alg:gpBOREx} is changed to $\frac{1}{pk} \sum_{i} \sum_{s} \frac{1}{r_i^2 t_s} \mu(\lambda,r_i,t_s)$; the weight is changed to the reciprocal of the volume of each mask.
\end{itemize}

\section{Experiments}
\label{sec:experiments}

We implemented \cref{alg:gpBOREx} and conducted experiments to evaluate the effectiveness of BOREx.%
Due to the limited space, we report a part of the experimental results. %in the paper.
\iffull
See the appendix for the experimental environment and more results and discussions, particularly on video classification.
\else
See the supplementary material for the experimental environment and more results and discussions, particularly on video classification.
\fi

% RQ1: BO に比べてどのくらい早いか
% TODO: 縦軸に BO の実行時間，横軸に BO-Rex の実行時間をとって，各画像についてそれぞれの手法でどのくらい時間がかかったかを散布図で示す
% RQ2: RISE に比べてどのくらい推論回数が少なくなるか
% ?
% RQ3: 生成されるヒートマップの質は BO と RISE に比べてどのくらい良くなっているか
% TODO: BOとBO-Rex と，RISEとBO-Rex との f-value AUC を散布図にする
% RQ4: 速さ，推論回数，ヒートマップの質は事前分布のとり方にどの程度依存するか
% ?
% RISE の iteration 回数
% GradCAM とか使って生成したヒートマップ
% RQ5: RISE と BO の注力配分に質や速度がどの程度依存するか
% ?
% RQ6: PascalVOC を使って f-value を求める手法は良い尺度になっているか
% insertion/deletion との相関を見てみる？
% TODO: insertion–f-valueとdeletion–f-valueの散布図を書く
% RISE論文のpointing gameとの関係は？

The research questions that we are addressing are the following.
\begin{description}
\item[RQ1:]
  \textbf{Does BOREx improve the quality of an input saliency map?}
  This is to evaluate that BOREx is useful to refine a potentially low-quality saliency map, which is the main claim of this paper.
\item[RQ2:]
  \textbf{Does Algorithm~\ref{alg:gpBOREx} produce a better saliency map than one produced by Algorithm~\ref{alg:gp} by Mokuwe et al.~\cite{DBLP:conf/ijcnn/MokuweBB20}?}
  This is to demonstrate the merit of BOREx over the algorithm by Mokuwe et al.
\item[RQ3:]
  \textbf{Does the extension in Section~\ref{sec:video} useful as a saliency-map generation for video classifiers?}
  This is to evaluate the competency of BOREx to explain a video-classification result.
\end{description}

\paragraph{Evaluation metrics.}
To quantitatively evaluate the quality of a saliency map, we used the
following three measures.
\begin{description}
\item[Insertion:] For a saliency map $i_{\SAL}$ explaining a
  classification of an image $i$ to label $l$, the \emph{insertion} metric is
  defined as $\sum_{k} M(i^{(k)},l)$, where $i^{(k)}$ is the image
  obtained by masking all the pixels other than those with top-$k$
  saliency values in $i_{\SAL}$ to black.
\item[Deletion:] The \emph{deletion} metric is defined as
  $\sum_{k} M(i^{(-k)},l)$, where $i^{(-k)}$ is the image obtained by
  masking all the pixels with top-$k$ saliency values in $i_{\SAL}$ to
  black.
% \item[F-measure:] The \emph{F-measure} in our experiments is defined
%   as $\sum_{k} F(i^{(k)},l,B_{i,l})$, where $B_i$ is the
%   human-annotated bounded region in $i$ that indicates an object of
%   label $l$ and $F(i^{(k)},l,B_{i,l})$ is the F-measure calculated
%   from the recall and the precision of the pixels in $i^{(k)}$ with
%   respect to the region $B_{i,l}$.
\item[F-measure:] The \emph{F-measure} in our experiments is defined
  as $\sum_{k} F(i^{(k)},l,B_{i,l})$, where 
  $F(i^{(k)},l,B_{i,l})$ is the F-measure calculated
  from the \emph{recall} and the \emph{precision} of the pixels in $i^{(k)}$ against
  the human-annotated bounded region $B_{i,l}$ in $i$ that indicates
  an object of label $l$.
\end{description}

The insertion and the deletion metrics are introduced by~\cite{DBLP:conf/bmvc/PetsiukDS18} to quantitatively evaluate how well a saliency map localizes a region that is important for a decision by a model.
The higher value of the insertion metric is better; the lower value of the deletion metric is better.
The higher insertion implies that $i_{\SAL}$ localizes regions in $i$ that are enough for classifying $i$ to $l$.
The lower deletion implies that $i_{\SAL}$ localizes regions that are indispensable for classifying $i$ to $l$.
The F-measure is an extension of their pointing-game metric also to consider recall, not only the precision.
The higher value of F-measure is better, implying $i_{\SAL}$ points out more of an important region correctly.

%% According to Li et al.~\cite{9435317}, the most reliable metric for a saliency-based interpretation method is the insertion metric.
%% %
%% Following their argument, we consider the insertion metric as the most important metric.

% \paragraph{Statistical hypothesis test.}
In what follows, we use a statistical hypothesis test called the \emph{one-sided Wilcoxon signed-rank test}~\cite{woolson2007wilcoxon} (or, simply \emph{Wilcoxon test}).
This test is applied to matched pairs of values $\{(a_1,b_1),\dots,(a_n,b_n)\}$ sampled from a distribution and can be used to check whether the median of $\{a_1,\dots,a_n\}$ can be said to be larger or smaller than that of $\{b_1,\dots,b_n\}$ with significance.
To compare saliency generation methods $X$ and $Y$, we calculate the pairs of the values of metrics evaluated with a certain dataset, the first of each are of the method $X$ and the second are of $Y$; then, we apply the Wilcoxon test to check the difference in the metrics.
For further details, see~\cite{woolson2007wilcoxon}.

%
% In contrast to the paired Student's $t$-test~\cite{ttest}, the Wilcoxon test does not assume that the difference between the first and the second components of the paired values are normally distributed.

% \paragraph{Conducted experiments.}
To address these RQs, we conducted the following experiments:
\begin{description}
\item[RQ1:] We used
  \textsc{RISE}~\cite{DBLP:conf/bmvc/PetsiukDS18} and
  \textsc{Grad-CAM++}~\cite{DBLP:conf/wacv/ChattopadhyaySH18} to generate
  saliency maps for the images in PascalVOC
  dataset~\cite{Everingham10}; we write $D_{\mathrm{RISE}}$ and
  $D_{\mathrm{GradCAM++}}$ for the set of saliency maps generated by
  \textsc{RISE} and \textsc{Grad-CAM++}, respectively.
  Then, we applied BOREx with these saliency maps as input; we write
  $D_{\mathrm{RISE}}^{\mathrm{BOREx}}$ (resp.,
  $D_{\mathrm{GradCAM++}}^{\mathrm{BOREx}}$) for the saliency maps
  generated using $D_{\mathrm{RISE}}$ (resp.,
  $D_{\mathrm{GradCAM++}}$) as input.
  We check whether the quality of the saliency maps in
  $D_{-}^{\mathrm{BOREx}}$ is better than $D_{-}$ by the one-sided
  Wilcoxon signed-rank test.
  If so, we can conclude that BOREx indeed improves the saliency
  map generated by other methods.
\item[RQ2:]
  We generated saliency maps for the PascalVOC dataset using Mokuwe et al.~\cite{DBLP:conf/ijcnn/MokuweBB20} presented in Algorithm~\ref{alg:gp}; we write $D_{\mathrm{BO}}$ for the generated saliency maps.
  We check if the quality of the saliency maps in
  $D_{\mathrm{RISE}}^{\mathrm{BOREx}}$ is better than
  $D_{\mathrm{BO}}$ by one-sided Wilcoxon signed-rank test.
  %
  % Since Algorithm~\ref{alg:gp} starts the Gaussian process regression
  % from all-zero $\mu$ (Line~\ref{alg:gp:initializeMu}) rather than starting from
  % an input saliency map, this comparison studies
  % the merit of giving an input saliency map to our algorithm.
  If so, we can conclude the merit of BOREx over the method by Mokuwe et al.
\item[RQ3:]
  We generated saliency maps for the dataset in Kinetics-400 using an
  extension of GradCAM++ and RISE for video classification implemented
  by us; let the set of saliency maps $D_{M,\mathrm{GradCAM++}}$ and
  $D_{M,\mathrm{RISE}}$, respectively.
  Then, we applied BOREx with these saliency maps as input; we write
  $D_{\mathrm{M,RISE}}^{\mathrm{BOREx}}$ (resp.,
  $D_{\mathrm{M,GradCAM++}}^{\mathrm{BOREx}}$) for the saliency maps
  generated using $D_{\mathrm{M,RISE}}$ (resp.,
  $D_{\mathrm{M,GradCAM++}}$) as input.
  We check whether the quality of the saliency maps in
  $D_{M,-}^{\mathrm{BOREx}}$ is better than $D_{M,-}$ by one-sided
  Wilcoxon signed-rank test.
  If so, we can conclude the merit of BOREx as an explanation method for
  a video-classification result.
\end{description}
As the model whose classification behavior to be explained, we used
ResNet-152~\cite{DBLP:conf/cvpr/HeZRS16} obtained from
torchvision.models\footnote{\url{https://pytorch.org/vision/stable/models.html}},
which is pre-trained with ImageNet~\cite{DBLP:conf/cvpr/DengDSLL009},
for RQ1 and RQ2; and i3D~\cite{DBLP:conf/cvpr/CarreiraZ17} obtained
from TensorFlow
Hub\footnote{\url{https://tfhub.dev/deepmind/i3d-kinetics-400/1}},
which is pre-trained with Kinetics-400~\cite{DBLP:journals/corr/KayCSZHVVGBNSZ17}.
Notice that the datasets PascalVOC and Kinetics-400 provide human-annotated bounding regions for each label and each image, enabling computation of the F-measure.

% As baseline saliency map generation methods, we used \textsc{RISE} by~\cite{DBLP:conf/bmvc/PetsiukDS18} and Bayesian-optimization-based method (\textsc{BO}) by~\cite{DBLP:conf/ijcnn/MokuweBB20}.
% As baseline saliency map generation methods, we used \textsc{RISE} by~\cite{DBLP:conf/bmvc/PetsiukDS18}, \textsc{Grad-CAM++} by~\cite{DBLP:conf/wacv/ChattopadhyaySH18}, and the 
% Bayesian-optimization-based method (\textsc{BO}) by~\cite{DBLP:conf/ijcnn/MokuweBB20}.

% %----------------------------------------------------------
% \begin{table*}[tbp]
%  \caption{Summary of the benchmarks we used to evaluate the saliency map generation methods for image- or -video classification models.}%
%  \label{table:benchmark_summary}
%  \small
%  \centering
%  \begin{tabular}{lcc}
%   \toprule
%   & Explained Pre-trained Model & Classified Data for Saliency Map Generation\\
%   \midrule
%   Image-classification & ResNet152 by~\cite{DBLP:conf/cvpr/HeZRS16} from torchvision.models\footnote{\url{https://pytorch.org/vision/stable/models.html}} & ImageNet by~\cite{DBLP:conf/cvpr/DengDSLL009}\\
%   Video-classification & i3D by~\cite{DBLP:conf/cvpr/CarreiraZ17} from TensorFlow Hub\footnote{\url{https://tfhub.dev/deepmind/i3d-kinetics-400/1}} & Kinetics400 by~\cite{DBLP:journals/corr/KayCSZHVVGBNSZ17}\\
%   \bottomrule
%  \end{tabular}
% \end{table*}
% %----------------------------------------------------------

\subsection{Results and Discussion}
\label{sec:results}

% \KS{Delete Wilcoxon Statistics from the experiments.}

%----------------------------------------------------------
\begin{figure}[t]
 \begin{subfigure}[t]{.32\linewidth}
  \centering
  \includegraphics[trim=39 36 35 35,width=.5\linewidth,height=.5\linewidth,clip]{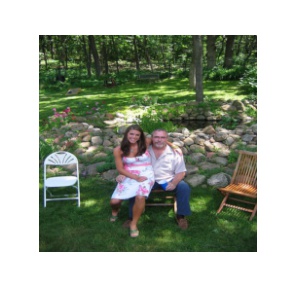}
  \caption{Input}%
  \label{fig:chair:image}
 \end{subfigure}
 \hfill
 \begin{subfigure}[t]{.32\linewidth}
  \centering
  \includegraphics[trim=77 38 117 43,width=.5\linewidth,height=.5\linewidth,clip]{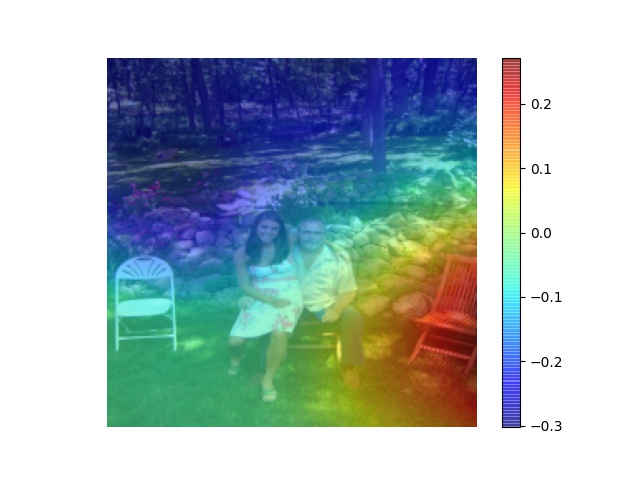}
  \caption{BOREx}%
  \label{fig:chair:borex}
 \end{subfigure}
 \hfill
 \begin{subfigure}[t]{.32\linewidth}
  \centering
  \includegraphics[trim=77 38 117 43,width=.5\linewidth,height=.5\linewidth,clip]{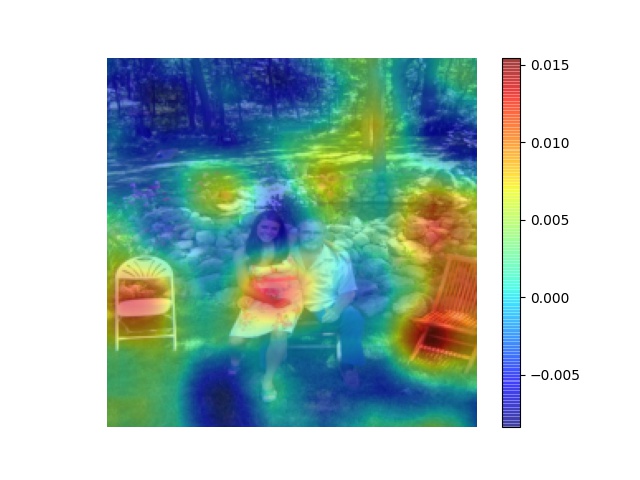}
  \caption{RISE}%
  \label{fig:chair:rise}
 \end{subfigure}
 \caption{Image of chairs and saliency maps to explain it.}%
 \label{fig:chair}
\end{figure}

% The results are shown in Table~\ref{tab:result:rq1-2}.

\paragraph{RQ1.}
\cref{tab:result:rq1-2} shows that BOREx improved the quality of the saliency maps generated by RISE and Grad-CAM++ in several metrics with statistical significance ($p < 0.001$).
Therefore, we conclude that \textbf{BOREx successfully refines an input saliency map}.
%% %
%% The statistical test also confirms that BOREx improves the quality of (1) a saliency map generated by RISE in terms of F-measure and (2) a saliency map generated by Grad-CAM++ in terms of the deletion metric.
This improvement is thanks to the Gaussian process regression that successfully captured the locality of the salient pixels.
For example, the saliency maps in \cref{fig:intro:saliency} suggest that BOREx is better at generalizing the salient pixels to the surrounding areas than RISE.

The time spent for GPR-based optimization was $9.26 \pm 0.26$ seconds in average for each image.
We believe this computation time pays off if we selectively apply BOREx to saliency maps whose quality needs to be improved.

To investigate the effect of the features of BOREx presented in Section~\ref{sec:borex} (i.e., flipped masks and the saliency-map computation from the result of GPR by weighted average in its performance), we conducted an ablation study; the result is shown in Table~\ref{tab:ablation}.
We compared BOREx with (1) a variant that does not use flipped masks (no-flip), (2) a variant that uses simple average instead of the average weighted by the inverse of the area of masks (simple-avg), and (3) a variant that does not use prior (no-prior).
The statistical test demonstrates that flipped masks and weighted averages are effective in the performance of BOREx.
However, the effectiveness over the no-prior variant is not confirmed.
This is mainly because, if the quality of a given prior is already high, the effectiveness of BOREx is limited.
Indeed, BOREx is confirmed to be effective over the no-prior case if the insertion metric of the priors is less than 0.6; see the row ``no-prior (base insertion $< 0.6$)'' in \cref{tab:ablation}.

The statistical test did not demonstrate the improvement in the deletion metric for a saliency map generated by RISE and the F-measure for a saliency map generated by Grad-CAM++.
Investigation of several images for which BOREx degrades the metrics reveals that this is partly because the current BOREx allows only square-shaped masks; this limitation degrades the deletion metric for an image with multiple objects with the target label $l$.
For example, a single square-shaped mask cannot focus on both chairs simultaneously in the image in \cref{fig:chair:image}.
For such an image, BOREx often focuses on only one of the objects, generating the saliency map in \cref{fig:chair:borex}.
Even if we mask the right chair in \cref{fig:chair:image}, we still have the left chair, and the confidence of the label ``chair'' does not significantly decrease, which degrades the deletion metric of the BOREx-generated saliency map.
\begin{table}[t]
  \footnotesize
\centering
 \small
\caption{Result of the experiments.  ``Image/Video'': The kind of the classifier; ``Compared with'': the baseline
  method; ``Metric'': evaluation metric; %``Stat.'': the value of the one-sided Wilcoxon
  %signed-rank test statistics;
  ``$p$-value'': the $p$-value.  The null hypothesis of each test
  expresses that the average of the metric of BOREx is not better than
  that of baseline.  One asterisk indicates $p < 0.05$; two asterisks
  indicates $p < 0.001$.}
\label{tab:result:rq1-2}
\begin{tabular}{lllrr}
  \toprule
  Image/Video & Compared with &  Metric & %Stat. &
 $p$-value \\
  \midrule
  Image & RISE & F-measure & \textbf{8.307e-21}${}^{**}$ \\
              && insertion & \textbf{1.016e-23}${}^{**}$ \\
              && deletion & 8.874e-01\\
              & Grad-CAM++ & F-measure & 1.000 \\
              && insertion & \textbf{5.090e-08}${}^{**}$\\
              && deletion & \textbf{6.790e-04}${}^{**}$ \\
              & BO & F-measure & \textbf{1.800e-05}${}^{**}$ \\
              && insertion & \textbf{6.630e-11}${}^{**}$ \\
              && deletion & 3.111e-01 \\
  \midrule
  Video & RISE & F-measure & \textbf{4.988e-07}${}^{**}$ \\
              && insertion & 8.974e-01 \\
              && deletion & \textbf{8.161e-18}${}^{**}$ \\
              & Grad-CAM++ & F-measure & 9.9980e-01 \\
              && insertion & 3.636e-01 \\
              && deletion & \textbf{2.983e-07}${}^{**}$ \\
  \bottomrule
\end{tabular}
% \begin{tabular}{llrr}
%   \toprule
%   Compared with &  Metric & Stat. &  $p$-value \\
%   \midrule
%   RISE & F-measure & 63841 & \textbf{8.307474e-21}${}^{**}$ \\
%               & insertion & 65484 & \textbf{1.016050e-23}${}^{**}$ \\
%               & deletion & 44608 & 8.87350e-01\\
%   Grad-CAM++ & F-measure & 16054 & 1.0 \\
%               & insertion & 54406 & \textbf{5.089895e-08}${}^{**}$\\
%               & deletion & 34083 & \textbf{6.79e-04}${}^{**}$ \\
%   % Grad-CAM++  & F-measure & 15577.0 & 8.730225e-09${}^{**}$ \\
%   % (two-sided) & insertion & 24111.0 & 6.32333e-01 \\
%   %             & deletion & 23228.0 & 3.05708e-01 \\
%   BO & F-measure & 34483 & \textbf{1.8e-05}${}^{**}$ \\
%      & insertion & 38450 & \textbf{6.629843e-11}${}^{**}$ \\
%      & deletion & 26453 & 3.11127e-01 \\
%   \bottomrule
% \end{tabular}
\end{table}
\paragraph{RQ2.}
The last three rows of \cref{tab:result:rq1-2} show that the use of an initial saliency map improved the quality of the saliency maps generated by Bayesian optimization in terms of several metrics with statistical significance compared to the case where the initial saliency map is not given ($p < 0.001$).
Therefore, we conclude that \textbf{BOREx produces a better saliency map than the one produced by Mokuwe et al. in terms of the insertion metric and F-measure.}

The improvement was not concluded in terms of the deletion metrics.
Investigation of the generated saliency maps suggests that such degradation is observed when the quality of a given initial saliency map is too low; if such a saliency map is given, it misleads an execution of BOREx, which returns a premature saliency map at the end of the prespecified number of iterations.

\begin{table}[t]
  \footnotesize
\centering
\caption{The result of ablation study.}
\label{tab:ablation}
\footnotesize
\begin{tabular}{llrr}
\toprule
Compared with &  Metric & %Statistics &
  $p$-value \\
  \midrule
  no-flip & F-measure & %419033 &
\textbf{2.987e-29}${}^{**}$ \\
          & insertion & %339830 &
\textbf{1.416e-04}${}^{**}$ \\
          & deletion & %271465 &
\textbf{2.024e-03}${}^{*}$ \\

  \hline
  simple-avg & F-measure & %331882 &
  \textbf{2.026e-03}${}^{*}$ \\
              & insertion & %451913 &
\textbf{1.184e-46}${}^{**}$ \\
              & deletion & %274508 &
                                     \textbf{4.871e-03}${}^{*}$ \\
  \hline
  no-prior & F-measure & %331882 &
                                   \textbf{4.514e-02}${}^{*}$\\
              & insertion & %451913 &
                                      3.84624e-01\\
              & deletion & %274508 &
                                     2.2194e-01\\
  \hline
  no-prior (base insertion $< 0.6$) & F-measure & %331882 &
                                                            \textbf{2.4825e-02}${}^{*}$\\
              & insertion & %451913 &
                                      \textbf{3.219e-03}${}^{*}$\\
              & deletion & %274508 &
                                     6.47929e-01\\
  \bottomrule
\end{tabular}
% \begin{tabular}{lllrr}
% \toprule
% Compared with &  Metric & Statistics &  $p$-value \\
%   \midrule
%   no-flip & F-measure & 419033 &  \textbf{2.987425e-29}${}^{**}$ \\
%           & insertion & 339830 &  \textbf{1.415743e-04}${}^{**}$ \\
%           & deletion & 271465 &  \textbf{2.024e-03}${}^{*}$ \\

%   \hline
%   simple-avg & F-measure & 331882 &  \textbf{2.026366e-03}${}^{*}$ \\
%               & insertion & 451913 &  \textbf{1.184027e-46}${}^{**}$ \\
%               & deletion & 274508 &  \textbf{4.871e-03}${}^{*}$ \\
%   \bottomrule
% \end{tabular}
\end{table}

\paragraph{RQ3.}

\Cref{tab:result:rq1-2} shows the result of the experiment for RQ3.
It shows that the saliency maps generated by the extensions of RISE and Grad-CAM++ for video classifiers are successfully refined by BOREx in terms of at least one metric with statistical significance ($p < 0.001$). 
Therefore, we conclude that \textbf{a saliency map produced by BOREx points out regions in a video that are indispensable to explain the classification result better than the other methods.}
%
% The improvement over RISE in terms of F-measure is also demonstrated.

% \begin{table}[t]
%   \footnotesize
%  \centering
% \caption{Result of the experiments for RQ3.  The meaning of the value
%   of each cell is the same as that of Table~\ref{tab:result:rq1-2}.}
% \label{tab:result:rq3}
% \begin{tabular}{llrr}
%   \toprule
%   Compared with &  Metric & %Statistics &
% $p$-value \\
%   \midrule
%   RISE & F-measure & %4092 &
% \textbf{4.988e-07}${}^{**}$ \\
%        & insertion & %2247 &
% 8.974e-01 \\
%        & deletion & %75 &
% \textbf{8.161e-18}${}^{**}$ \\
%   Grad-CAM++ & F-measure & %1544 &
% 9.9980e-01 \\
%              & insertion & %2731 &
% 3.636e-01 \\
%              & deletion & %1131 &
% \textbf{2.983e-07}${}^{**}$ \\
%   \bottomrule
% \end{tabular}
% % \begin{tabular}{llrr}
% %   \toprule
% %   Compared with &  Metric & Stat. &  $p$-value \\
% %   \midrule
% %   RISE & F-measure & 4092 & \textbf{4.988147e-07}${}^{**}$ \\
% %        & insertion & 2247 & 0.897394 \\
% %        & deletion & 75 & \textbf{8.160800e-18}${}^{**}$ \\
% %   Grad-CAM++ & F-measure & 1544 & 0.9998490 \\
% %              & insertion & 2731 & 0.363604 \\
% %              & deletion & 1131 & \textbf{2.983445e-07}${}^{**}$ \\
% %   \bottomrule
% % \end{tabular}
% \end{table}

The improvement in the insertion metric over RISE and Grad-CAM++, and in F-measure over Grad-CAM++ were not concluded.
The investigation of saliency maps whose quality is degraded by BOREx reveals that the issue is essentially the same as that of the images with multiple objects discussed above.
A mask used by BOREx occludes the same position across several frames; therefore, for a video in which an object with the target label moves around, it is difficult to occlude all occurrences of the object in different frames.
This limitation leads to a saliency map generated by BOREx that tends to point out salient regions only in a part of the frames, which causes the degradation in the insertion metric.
The improvement in deletion metric seems to be due to the mask shape of BOREx.
To improve the deletion metric for a video-classifier explanation, a saliency map must point out a salient region across several frames.
The current mask shape of BOREx is advantageous, at least for a video in which there is a single salient object that does not move around, to cover the salient object over several frames.
 
% , at least for a video in which there is a single salient object that does not move around, to cover the salient object over several frames.

% Therefore, an explanation-generating method must place masks on a salient region over several frames for a better deletion metric. RISE is not good at generating such masks because it generates them randomly; therefore, it is less probable that the masks are placed on salient regions of several frames. In contrast, the mask shape of BOREx is advantageous, at least for a video in which there is a single salient object that does not move around, to cover the salient object over several frames.

% On the contrary, the deletion metric of BOREx is improved since occluding important regions in several frames is often enough for preventing the model from inferring a correct label.

% \paragraph{Ablation study}
% \label{sec:ablation}

% We also conducted an ablation study.  We
% \KS{From here.}

% \KS{Write a paragraph for ablation study.}

%% \KS{Write this section.}

\section{Conclusion}
\label{sec:conclusion}

% \MW{We have to modify it if we omit the comparison with Grad-CAM++.}
This paper has presented BOREx, a method to refine a potentially low-quality saliency map that explains a classification result of image and video classifiers.
Our refinement of a saliency map with Bayesian optimization applies to any existing saliency-map generation method.
The experiment results demonstrate that BOREx improves the quality of the saliency maps, especially when the quality of the given saliency map is neither too high nor too low.
%
% DO NOT TALK ABOUT THE SPEED
% By combining a fast but not necessarily high-quality saliency map generation method with our refinement method, we can efficiently obtain a high-quality saliency map.
% This is especially the case when an inference of a classification model is time-consuming, e.g., the inference of a 3D CNN for video classification.
% \KS{Write conclusion.}\MW{I wrote a draft of the conclusion.}

%% Future works
% The future direction includes addressing the issues identified through the experiments.
%
We are currently looking at enhancing BOREx by investigating the optimal shape of masks to improve performance.
% for images with multiple objects is a future work.
%
Another important research task is making BOREx more robust to an input saliency map with very low quality.
% tune the Bayesian optimization so that the resulting saliency map has a good quality even if the quality of the given saliency map is very low.
%
% Extension of our saliency-map refinement for 
% RNNs with convolutional layers is also a potential future direction.

% \paragraph{Acknowledgment.}
% We thank Atsushi Nakazawa for his fruitful comments on this work.
% %
% KS is partially supported by JST, CREST Grant Number JPMJCR2012, Japan.
% %
% MW is partially supported by JST, ACT-X Grant Number JPMJAX200U, Japan.

\ifdraft
\pagelimitmarker{14}
\fi

\clearpage
% ---- Bibliography ----
%
% BibTeX users should specify bibliography style 'splncs04'.
% References will then be sorted and formatted in the correct style.
%

\iffull

\appendix

\section{The detail of experimental environment}

\sloppy
We implemented \iffull\cref{alg:gpBOREx}\else\cref{main:alg:gpBOREx}\fi using Python 3.6.12 and PyTorch 1.6.0.
The experiments on image classification are conducted on a GPU workstation with 3.60\,GHz Intel Core i7-6850K, 12 CPUs, NVIDIA Quadro P6000, and 32GB RAM that runs Ubuntu 20.04.2 LTS (64\,bit) and CUDA 11.0.
The experiments on video classification are conducted on a GPU workstation with 3.00\,GHz Intel Xeon E5-2623 v3, 16 CPUs, NVIDIA Tesla P100, and 500GB RAM that runs Ubuntu 16.04.3 LTS (64\,bit).
\KS{Explain the number of iterations in each method.}
In the implementation of \iffull\cref{alg:gpBOREx}\else\cref{main:alg:gpBOREx}\fi, we used Mat\'ern kernel defined as follows:
\begin{eqnarray}
  k((\lambda_1,r_1),(\lambda_2,r_2)) := \frac{2^{1-\nu}}{\Gamma(\nu)} {\left(\frac{\sqrt{2\nu}d'}{l}\right)}^\nu K_{\nu}\left({\frac{\sqrt{2\nu}d'}{l}}\right), \nonumber
\end{eqnarray}
where $d'$ is the Euclidean distance between $(\lambda_1,r_1)$ and $(\lambda_2,r_2)$ (i.e., $\sqrt{d(\lambda_1,\lambda_2)^2 + (r_1 - r_2)^2}$ where $d(\lambda_1,\lambda_2)$ is the Euclidean distance between $\lambda_1$ and $\lambda_2$ in the input image); $\nu$ and $l$ are positive parameters that control the shape of the function; $\Gamma$ is the gamma function; and $K_{\nu}$ is a modified Bessel function~\cite{AbraSteg72}.  Mokuwe et al.~\cite{DBLP:conf/ijcnn/MokuweBB20} used $\nu=2.5$ and $l=12$ in their implementation.
We use the Mat\'ern kernel with $\nu=1.5$ and $l=12$.

\section{Results of additional experiments}

\begin{table}[t]\small
\centering
\caption{Comparison of the results produced by different number of $N$ in Algorithm~\iffull\ref{alg:gpBOREx}\else\ref{main:alg:gpBOREx}\fi.  Each result uses an input saliency map generated by RISE with $100$ masks.  ``BO$n$'' represents Algorithm~\iffull\ref{alg:gpBOREx}\else\ref{main:alg:gpBOREx}\fi with $N=n$.  Each column represents the following: ``Metric'' for the metric; ``Base.'' for the baseline; ``Comp.'' for the compared method; ``$p$-val.'' for the $p$-value.  One asterisk indicates $p < 0.05$; two asterisks indicates $p < 0.001$.}
\label{tab:BOIterations}
\begin{tabular}{lllr}
\toprule
Metric & Base. & Comp. &  $p$-val. \\
\midrule
     Insertion & BO10 &  BO50 & \textbf{7.042e-36}${}^{**}$ \\
          & BO10 &  BO80 & \textbf{4.023e-38}${}^{**}$ \\
          & BO10 &  BO100 & \textbf{2.261e-38}${}^{**}$ \\
          & BO50 &  BO80 & 8.156e-02 \\
          & BO50 &  BO100 & 7.542e-02 \\
          & BO80 & BO100 & 3.066e-01 \\
     \hline
     Deletion & BO10 &  BO50 & \textbf{2.937e-32}${}^{**}$ \\
          & BO10 &  BO80 & \textbf{1.838e-38}${}^{**}$ \\
          & BO10 &  BO100 & \textbf{2.007e-41}${}^{**}$ \\
          & BO50 &  BO80 & \textbf{4.454e-03}${}^{*}$ \\
          & BO50 &  BO100 & \textbf{1.132e-04}${}^{**}$\\
          & BO80 & BO100 & \textbf{3.721e-02}${}^{*}$ \\
     \hline
     F-measure & BO10 &  BO50 & \textbf{4.825e-32}${}^{**}$ \\
          & BO10 &  BO80 & \textbf{1.442e-34}${}^{**}$ \\
          & BO10 &  BO100 & \textbf{1.010e-34}${}^{**}$ \\
          & BO50 &  BO80 & \textbf{1.606e-03}${}^{*}$ \\
          & BO50 &  BO100 & \textbf{7.387e-04}${}^{**}$ \\
          & BO80 & BO100 & 5.806e-01 \\
     \bottomrule
\end{tabular}
\end{table}

\begin{table}[t]\small
\centering
\caption{Two-sided test to compare the results produced by different number of $N$ in Algorithm~\iffull\ref{alg:gpBOREx}\else\ref{main:alg:gpBOREx}\fi.  Each result uses an input saliency map generated by RISE with $100$ masks.}
\label{tab:BOIterations:twosided}
\begin{tabular}{lllr}
\toprule
Metric & Base. & Comp. &  $p$-val. \\
\midrule
     Insertion & BO10 &  BO50 & \textbf{1.408e-35}${}^{**}$ \\
          & BO10 &  BO80 & \textbf{8.046e-38}${}^{**}$ \\
          & BO10 &  BO100 & \textbf{4.522e-38}${}^{**}$ \\
          & BO50 &  BO80 & 1.631e-01 \\
          & BO50 &  BO100 &  1.508e-01 \\
          & BO80 & BO100 & 6.131e-01 \\
     \hline
     Deletion & BO10 &  BO50 & \textbf{5.875e-32}${}^{**}$ \\
          & BO10 &  BO80 & \textbf{3.677e-38}${}^{**}$ \\
          & BO10 &  BO100 & \textbf{4.013e-41}${}^{**}$ \\
          & BO50 &  BO80 & \textbf{8.907e-03}${}^{*}$ \\
          & BO50 &  BO100 & \textbf{2.264e-04}${}^{**}$ \\
          & BO80 & BO100 & 7.442e-02 \\
     \hline
     F-measure & BO10 &  BO50 & \textbf{9.650e-32}${}^{**}$ \\
          & BO10 &  BO80 & \textbf{2.884e-34}${}^{**}$ \\
          & BO10 &  BO100 & \textbf{2.020e-34}${}^{**}$ \\
          & BO50 &  BO80 & \textbf{3.212e-03}${}^{*}$ \\
          & BO50 &  BO100 & \textbf{1.477e-03}${}^{*}$ \\
          & BO80 & BO100 & 8.388e-01 \\
     \bottomrule
\end{tabular}
\end{table}

\begin{table}[t]\small
\centering
\caption{Comparison of the results of Algorithm~\iffull\ref{alg:gpBOREx}\else\ref{main:alg:gpBOREx}\fi with input saliency maps produced by RISE with different number of masks.  $N$ is set to $50$ in each execution of Algorithm~\iffull\ref{alg:gpBOREx}\else\ref{main:alg:gpBOREx}\fi.  ``RISE$n$'' represents input salinecy maps produced by RISE with $n$ masks are used.  Each column represents the following: ``Metric'' for the metric; ``Base.'' for the baseline; ``Comp.'' for the compared method; ``$p$-val.'' for the $p$-value.  One asterisk indicates $p < 0.05$; two asterisks indicates $p < 0.001$.}
\label{tab:RISEIterations}
\begin{tabular}{lllr}
\toprule
  Metric & Base. & Comp. &  $p$-val. \\
  \midrule
  Deletion & RISE0 &  RISE100 & 8.453e-01 \\
  & RISE0 &  RISE300 & 6.630e-01 \\
  & RISE0 &  RISE500 & 6.202e-01 \\
  & RISE0 & RISE1000 & 3.491e-01 \\
  & RISE100 &  RISE300 & 3.976e-01 \\
  & RISE100 &  RISE500 & 3.277e-01 \\
  & RISE100 & RISE1000 & 1.773e-01 \\
  & RISE300 &  RISE500 & 4.381e-01 \\
  & RISE300 & RISE1000 & 2.786e-01 \\
  & RISE500 & RISE1000 & 2.085e-01 \\
  \hline
  F-meas. & RISE0 &  RISE100 & 1.123e-01 \\
  & RISE0 &  RISE300 & \textbf{1.131e-02}${}^{*}$ \\
  & RISE0 &  RISE500 & 1.81777e-01 \\
  & RISE0 & RISE1000 & 1.636e-01 \\
  & RISE100 &  RISE300 & 4.176e-01 \\
  & RISE100 &  RISE500 & 7.307e-01 \\
  & RISE100 & RISE1000 & 5.568e-01 \\
  & RISE300 &  RISE500 & 7.514e-01 \\
  & RISE300 & RISE1000 & 9.475e-01 \\
  & RISE500 & RISE1000 & 3.592e-01 \\
  \hline
  Insertion & RISE0 &  RISE100 & 4.525e-02 \\
  & RISE0 &  RISE300 & \textbf{3.566e-03}${}^{*}$ \\
  & RISE0 &  RISE500 & \textbf{3.518e-03}${}^{*}$ \\
  & RISE0 & RISE1000 & \textbf{1.101e-03}${}^{*}$ \\
  & RISE100 &  RISE300 & 2.698e-01 \\
  & RISE100 &  RISE500 & 1.473e-01 \\
  & RISE100 & RISE1000 & 7.709e-02 \\
  & RISE300 &  RISE500 & 3.091e-01 \\
  & RISE300 & RISE1000 & 1.492e-01 \\
  & RISE500 & RISE1000 & 2.151e-01 \\
  \bottomrule
\end{tabular}
\end{table}

\begin{table}[t]\small
\centering
\caption{Two-sided test to compare Algorithm~\iffull\ref{alg:gpBOREx}\else\ref{main:alg:gpBOREx}\fi with input saliency maps produced by RISE with different number of masks.}
\label{tab:RISEIterations:twosided}
\begin{tabular}{lllr}
\toprule
  Metric & Base. & Comp. & $p$-val. \\
  \midrule
  Deletion & RISE0 &  RISE100 & 3.09308e-01 \\
         & RISE0 &  RISE300 & 6.73922e-01 \\
         & RISE0 &  RISE500 & 7.59515e-01 \\
         & RISE0 & RISE1000 & 6.98164e-01 \\
         & RISE100 &  RISE300 & 7.95128e-01 \\
         & RISE100 &  RISE500 & 6.55416e-01 \\
         & RISE100 & RISE1000 & 3.54525e-01 \\
         & RISE300 &  RISE500 & 8.76144e-01 \\
         & RISE300 & RISE1000 & 5.57116e-01 \\
         & RISE500 & RISE1000 & 4.17039e-01 \\
  \hline
  F-meas. & RISE0 &  RISE100 & 2.24636e-01 \\
         & RISE0 &  RISE300 & \textbf{2.2610e-02}${}^{*}$ \\
         & RISE0 &  RISE500 & 3.63555e-01 \\
         & RISE0 & RISE1000 & 3.27184e-01 \\
         & RISE100 &  RISE300 & 8.35268e-01 \\
         & RISE100 &  RISE500 & 5.38648e-01 \\
         & RISE100 & RISE1000 & 8.86484e-01 \\
         & RISE300 &  RISE500 & 4.97256e-01 \\
         & RISE300 & RISE1000 & 1.04953e-01 \\
         & RISE500 & RISE1000 & 7.18478e-01 \\
  \hline
  Insertion & RISE0 &  RISE100 & \textbf{4.5251e-02}${}^{*}$ \\
         & RISE0 &  RISE300 & \textbf{3.566e-03}${}^{*}$ \\
         & RISE0 &  RISE500 & \textbf{3.518e-03}${}^{*}$ \\
         & RISE0 & RISE1000 & \textbf{1.101e-03}${}^{*}$ \\
         & RISE100 &  RISE300 & 2.698e-01 \\
         & RISE100 &  RISE500 & 1.473e-01 \\
         & RISE100 & RISE1000 & 7.709e-02 \\
         & RISE300 &  RISE500 & 3.091e-01 \\
         & RISE300 & RISE1000 & 1.492e-01 \\
         & RISE500 & RISE1000 & 2.151e-01 \\
  \bottomrule
\end{tabular}
\end{table}

%%% Local Variables:
%%% mode: japanese-latex
%%% TeX-master: "supplementary"
%%% End:

\subsection{Effect of increasing $N$ in Algorithm~\iffull\ref{alg:gpBOREx}\else\ref{main:alg:gpBOREx}\fi}

We executed Algorithm~\iffull\ref{alg:gpBOREx}\else\ref{main:alg:gpBOREx}\fi with different $N$, the number of iterations for Gaussian process regression.
With more iterations, the estimated $\mu$ is expected to be more precise.
However, the time spent in one iteration in the later iterations tends to be longer because there are more observations to fit.

% Table~\ref{tab:BOIterations} shows the result.
% %
% We observe that $N=80$ and $100$ produce better saliency maps than $N=50$ in any metrics; however, we cannot conclude that $N=100$ is better than $N=80$.
% %
% Although the optimal $N$ varies depending on the setting, this result suggest that there is an optimal $N$ such that executing Algorithm~\iffull\ref{alg:gpBOREx}\else\ref{main:alg:gpBOREx}\fi with more iterations than $N$ is not very effective for the quality of the output.

\KS{Rewrite this paragraph.}
Tables~\ref{tab:BOIterations} and \ref{tab:BOIterations:twosided} show the result.
We observe the significant improvement in the insertion metric only when we increased $N$ from 10 to the other values; in the other cases, improvement nor degradation is not concluded (Table~\ref{tab:BOIterations:twosided}).
For the other metrics, increasing $N$ from 50 to the other values is also concluded to be effective.
In any metrics, increasing $N$ from 80 to 100 was not concluded to be effective.
%
% We observe that $N=80$ and $100$ produce better saliency maps than $N=50$ in any metrics; however, we cannot conclude that $N=100$ is better than $N=80$.
%
% Although the optimal $N$ varies depending on the setting, this result suggest that there is an optimal $N$ such that executing Algorithm~\iffull\ref{alg:gpBOREx}\else\ref{main:alg:gpBOREx}\fi with more iterations than $N$ is not very effective for the quality of the output.
This result suggests that increasing $N$, which incurs time for executing Algorithm~\ref{tab:BOIterations}, is effective; however, the merit of increasing the number beyond certain number (here 50) is limited.

\subsection{Effect of the quality of an input saliency map on the output of Algorithm~\iffull\ref{alg:gpBOREx}\else\ref{main:alg:gpBOREx}\fi}

We executed Algorithm~\iffull\ref{alg:gpBOREx}\else\ref{main:alg:gpBOREx}\fi with input maps generated by RISE with different numbers of masks.
This experiment is to study the effect of the quality of an input saliency map on the output because the quality of an input saliency map is expected to be higher with more masks.

\KS{Rewrite this paragraph.}
Table~\ref{tab:RISEIterations} shows the result.
Significant improvement is observed (1) in the insertion metric when we increased the number of masks from 0 to 300 or more and (2) in the F-measure when we increased the number of masks from 0 to 300.
We cannot conclude the significant improvement in the other cases.
% We cannot conclude the effect of using more masks for generating an input saliency map in the deletion and the F-measure metrics.
%
The two-sided test we conducted, whose result is presented in Table~\ref{tab:RISEIterations:twosided}, does not conclude that the quality of the saliency maps measured in these metrics differ among the input saliency maps generated by RISE with different numbers of masks, which implies that the quality is not degraded by increasing the number of masks.
These results back our conclusion of BOREx is effective to improve a low-quality saliency map in terms of the insertion metric.

% \subsection{Ablation study}

% \paragraph{Flipped mask.}

% \paragraph{Square masks vs. rectangular masks.}

% \paragraph{Simple average vs. weighted average in saliency-map computation from $\mu$.}

\section{Examples of saliency maps}

\subsection{Examples in which BOREx successfully improves input images}

\begin{figure}[t]
  \begin{subfigure}[t]{.32\linewidth}\centering
    \includegraphics[trim=39 36 35 35,width=\imagewidth,height=\imagewidth,clip]{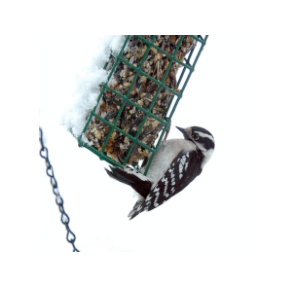}\\
   \includegraphics[trim=39 36 35 35,width=\imagewidth,height=\imagewidth,clip]{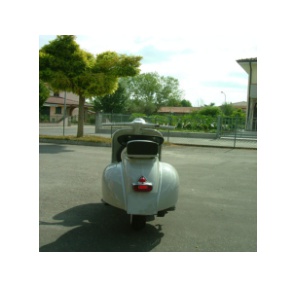}\\
   \includegraphics[trim=39 36 35 35,width=\imagewidth,height=\imagewidth,clip]{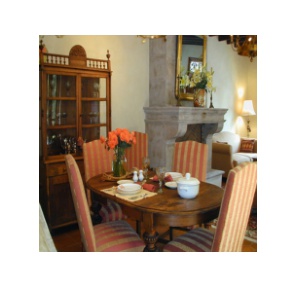}\\
   \includegraphics[trim=39 36 35 35,width=\imagewidth,height=\imagewidth,clip]{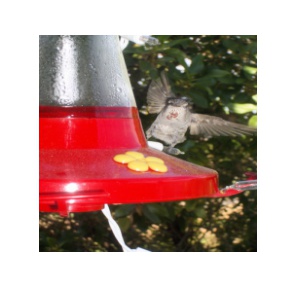}\\
   \includegraphics[trim=39 36 35 35,width=\imagewidth,height=\imagewidth,clip]{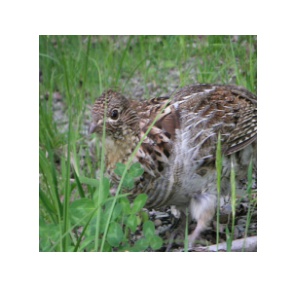}\\
   \includegraphics[trim=39 36 35 35,width=\imagewidth,height=\imagewidth,clip]{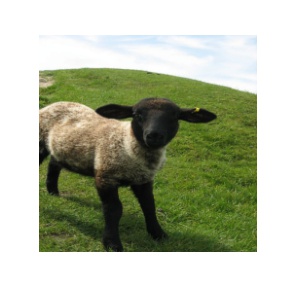}\\
   \includegraphics[trim=39 36 35 35,width=\imagewidth,height=\imagewidth,clip]{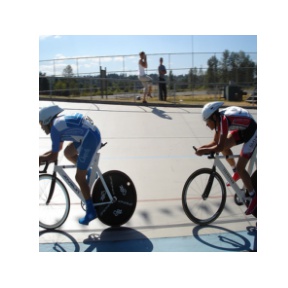}
   \caption{Input}%
   % \label{fig:motorbike:image}
 \end{subfigure}
 \hfill
 \begin{subfigure}[t]{.32\linewidth}\centering
   \includegraphics[trim=77 38 117 43,width=\imagewidth,height=\imagewidth,clip]{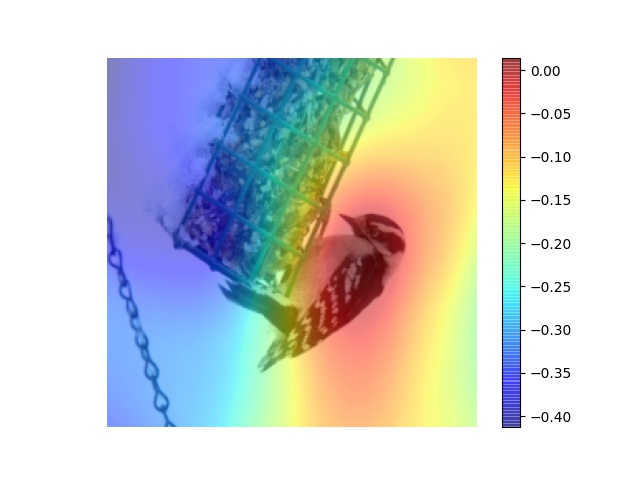}
   \includegraphics[trim=77 38 117 43,width=\imagewidth,height=\imagewidth,clip]{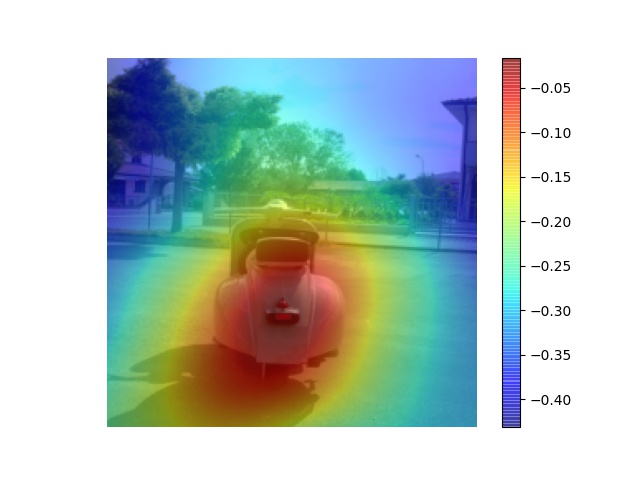}
   \includegraphics[trim=77 38 117 43,width=\imagewidth,height=\imagewidth,clip]{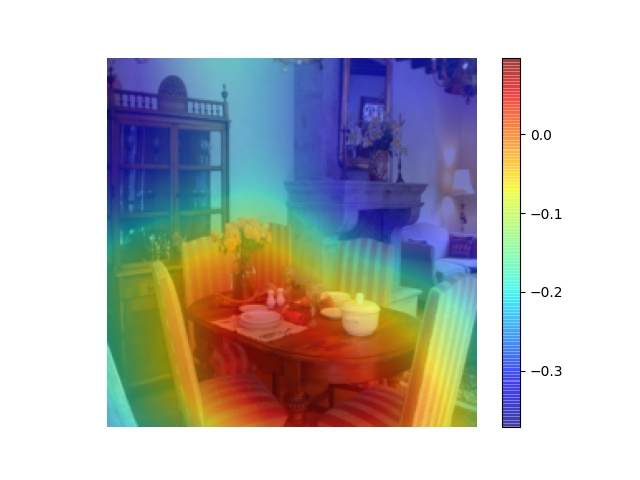}
   \includegraphics[trim=77 38 117 43,width=\imagewidth,height=\imagewidth,clip]{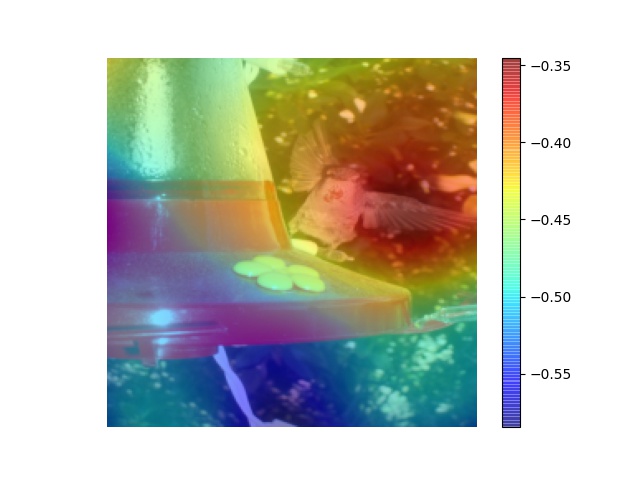}
   \includegraphics[trim=77 38 117 43,width=\imagewidth,height=\imagewidth,clip]{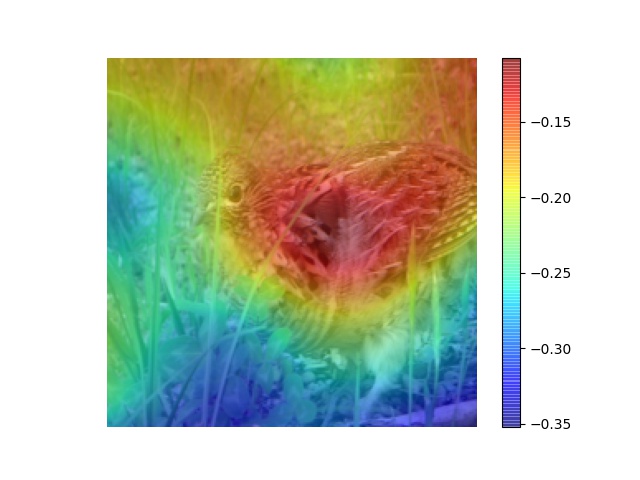}
   \includegraphics[trim=77 38 117 43,width=\imagewidth,height=\imagewidth,clip]{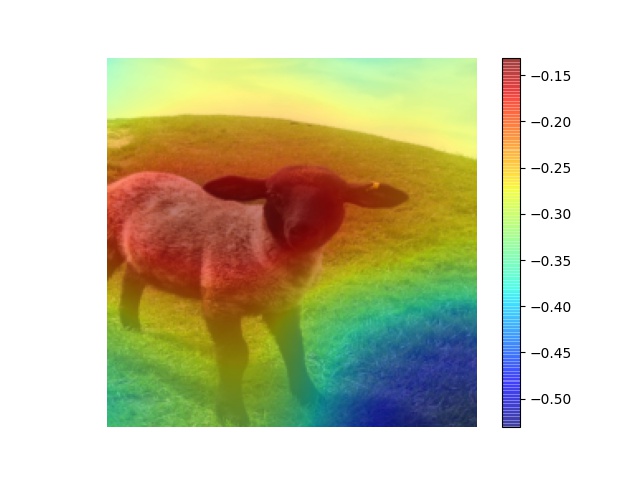}
   \includegraphics[trim=77 38 117 43,width=\imagewidth,height=\imagewidth,clip]{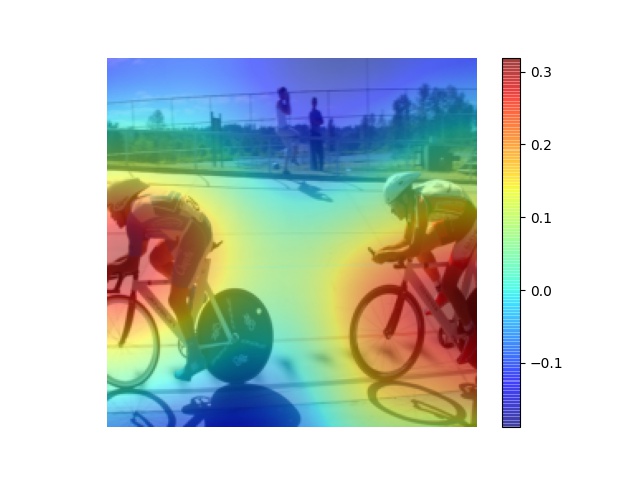}
   \caption{BOREx}%
   % \label{fig:motorbike:borex}
 \end{subfigure}
 \hfill
 \begin{subfigure}[t]{.32\linewidth}\centering
   \includegraphics[trim=77 38 117 43,width=\imagewidth,height=\imagewidth,clip]{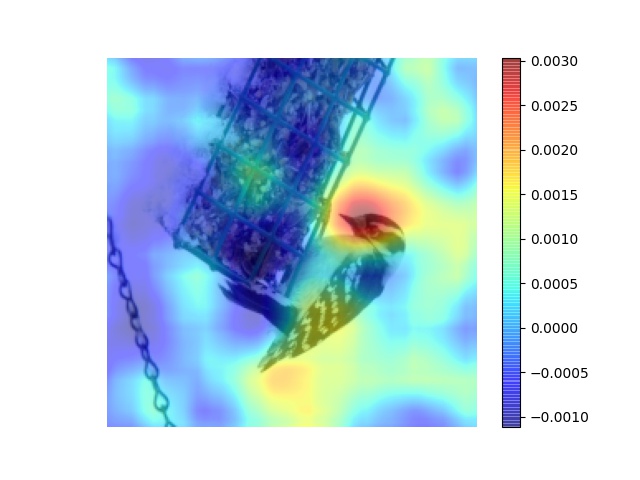}
   \includegraphics[trim=77 38 117 43,width=\imagewidth,height=\imagewidth,clip]{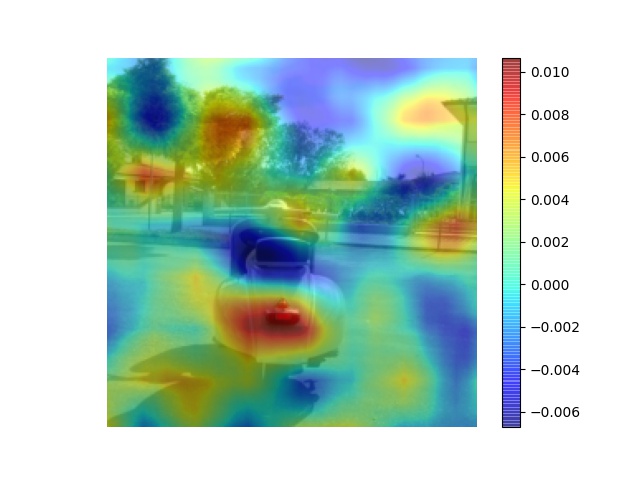}
   \includegraphics[trim=77 38 117 43,width=\imagewidth,height=\imagewidth,clip]{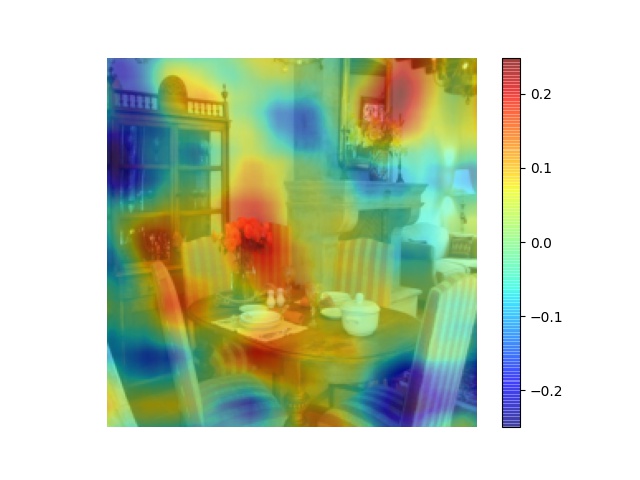}
   \includegraphics[trim=77 38 117 43,width=\imagewidth,height=\imagewidth,clip]{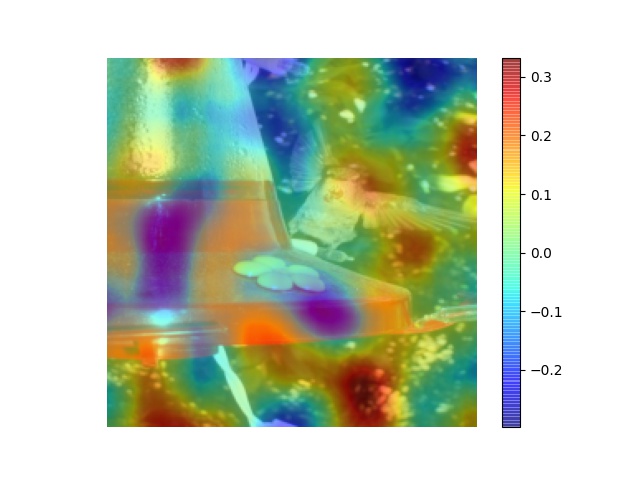}
   \includegraphics[trim=77 38 117 43,width=\imagewidth,height=\imagewidth,clip]{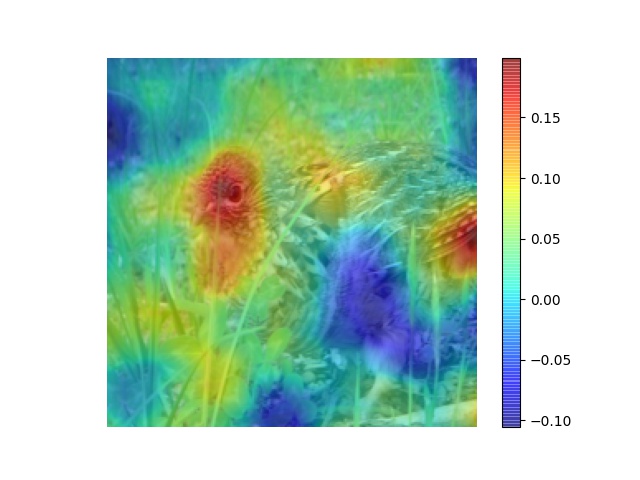}
   \includegraphics[trim=77 38 117 43,width=\imagewidth,height=\imagewidth,clip]{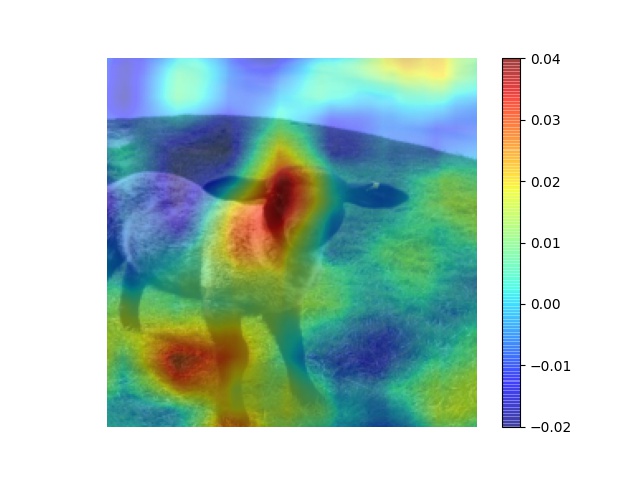}
   \includegraphics[trim=77 38 117 43,width=\imagewidth,height=\imagewidth,clip]{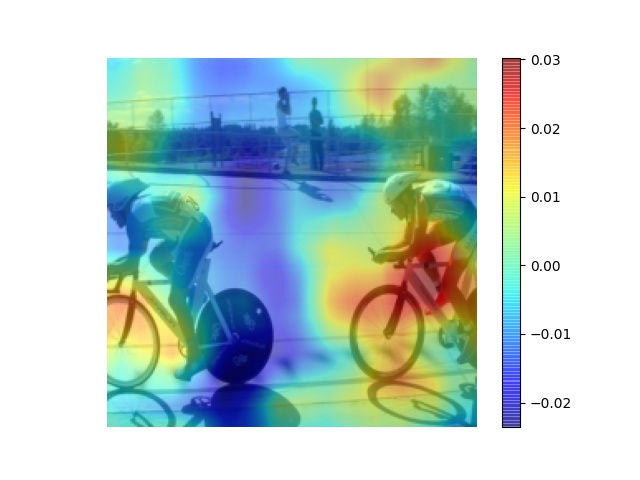}
   \caption{RISE}%
   % \label{fig:motorbike:rise}
 \end{subfigure}
 \caption{Examples of saliency maps that are successfully refined by BOREx.  The labels used in each explanation are: ``bird'', ``motor bike'', ``dining table'', ``bird'', ``bird'', ``sheep'', and ``bicycle'', from the first row.}%
 \label{fig:good:1}
\end{figure}

\begin{figure}[t]
 \begin{subfigure}[t]{.32\linewidth}\centering
   \includegraphics[trim=39 36 35 35,width=\imagewidth,height=\imagewidth,clip]{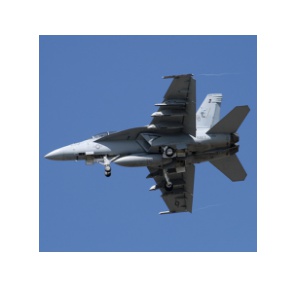}\\
   \includegraphics[trim=39 36 35 35,width=\imagewidth,height=\imagewidth,clip]{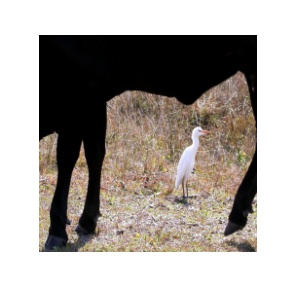}\\
   \includegraphics[trim=39 36 35 35,width=\imagewidth,height=\imagewidth,clip]{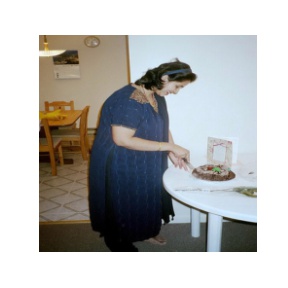}\\
   \includegraphics[trim=39 36 35 35,width=\imagewidth,height=\imagewidth,clip]{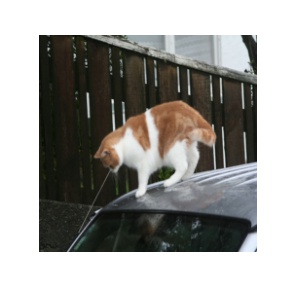}\\
   \includegraphics[trim=39 36 35 35,width=\imagewidth,height=\imagewidth,clip]{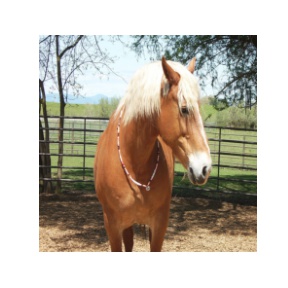}\\
   \includegraphics[trim=39 36 35 35,width=\imagewidth,height=\imagewidth,clip]{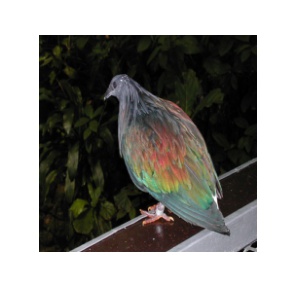}\\
   \includegraphics[trim=39 36 35 35,width=\imagewidth,height=\imagewidth,clip]{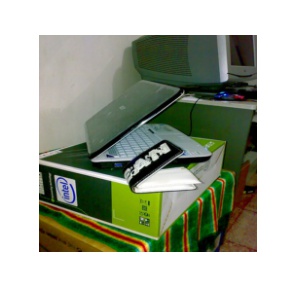}
   \caption{Input}%
   % \label{fig:motorbike:image}
 \end{subfigure}
 \hfill
 \begin{subfigure}[t]{.32\linewidth}\centering
   \includegraphics[trim=77 38 117 43,width=\imagewidth,height=\imagewidth,clip]{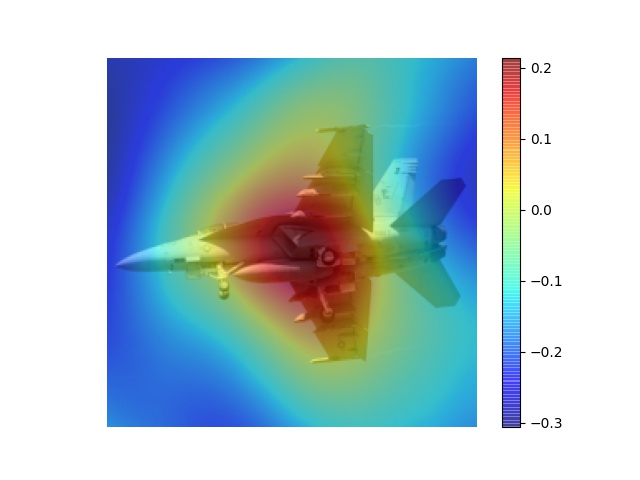}
   \includegraphics[trim=77 38 117 43,width=\imagewidth,height=\imagewidth,clip]{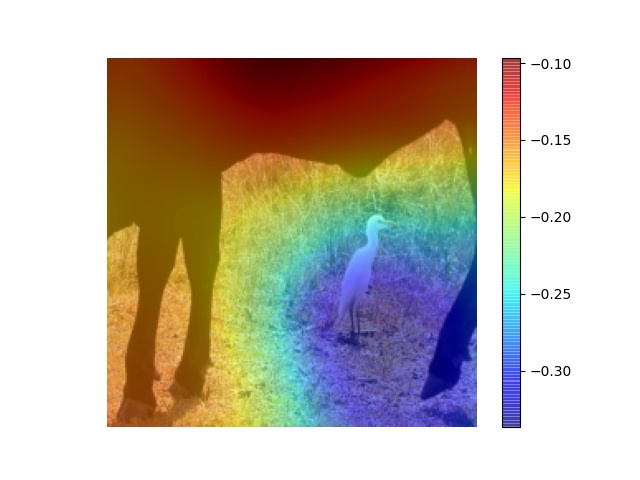}
   \includegraphics[trim=77 38 117 43,width=\imagewidth,height=\imagewidth,clip]{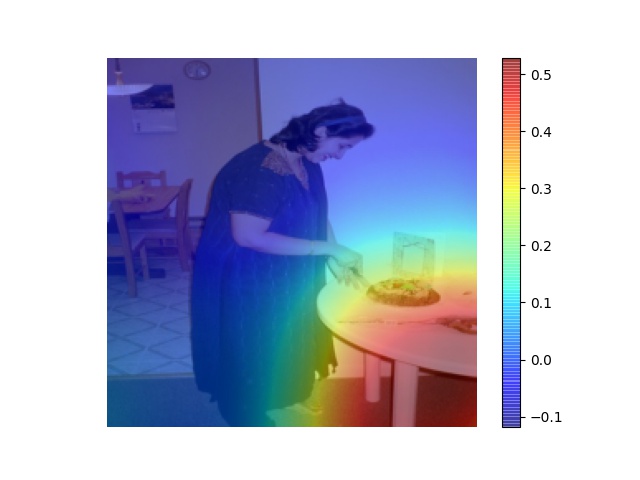}
   \includegraphics[trim=77 38 117 43,width=\imagewidth,height=\imagewidth,clip]{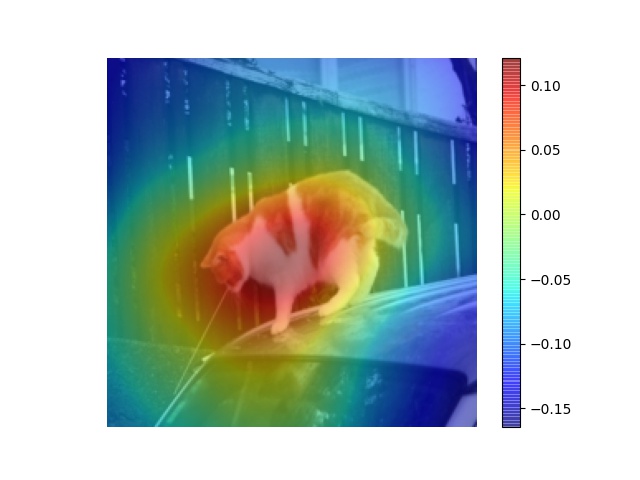}
   \includegraphics[trim=77 38 117 43,width=\imagewidth,height=\imagewidth,clip]{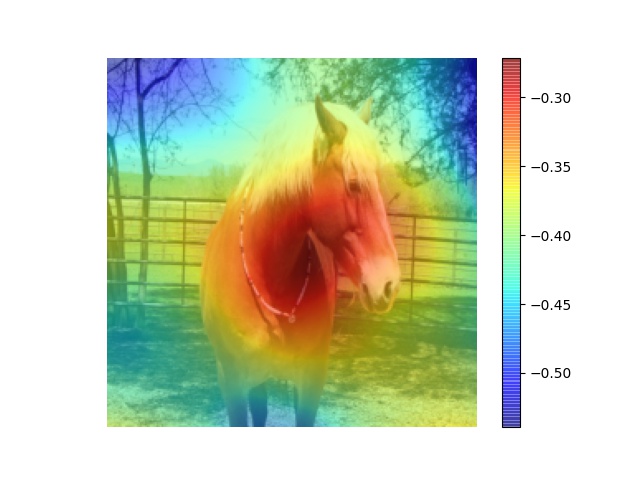}
   \includegraphics[trim=77 38 117 43,width=\imagewidth,height=\imagewidth,clip]{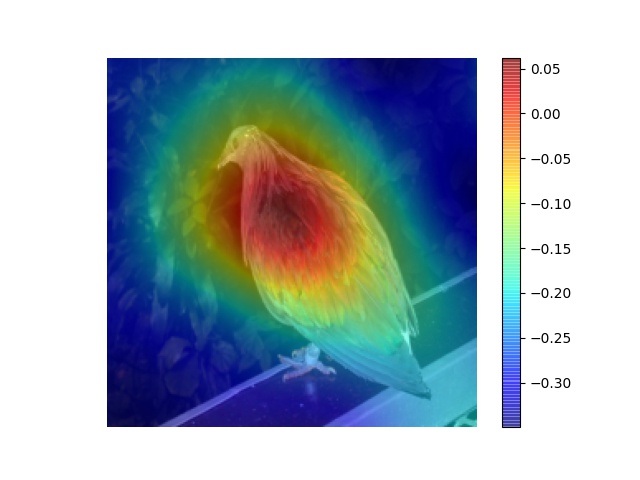}
   \includegraphics[trim=77 38 117 43,width=\imagewidth,height=\imagewidth,clip]{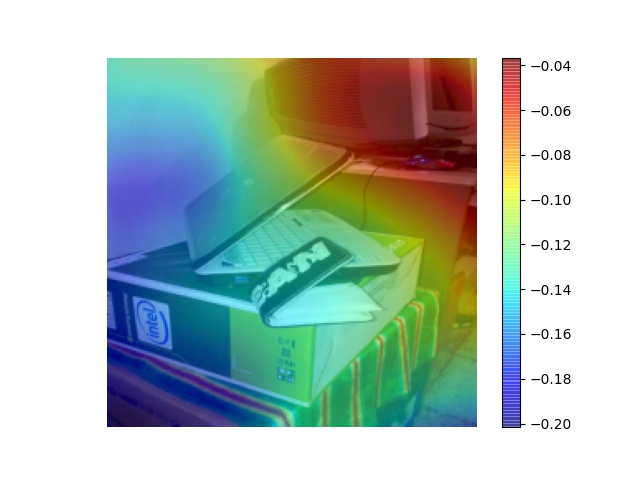}
   \caption{BOREx}%
   % \label{fig:motorbike:borex}
 \end{subfigure}
 \hfill
 \begin{subfigure}[t]{.32\linewidth}\centering
   \includegraphics[trim=77 38 117 43,width=\imagewidth,height=\imagewidth,clip]{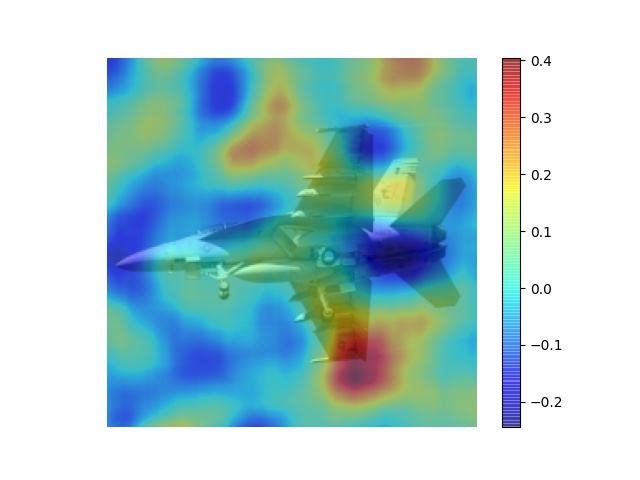}
   \includegraphics[trim=77 38 117 43,width=\imagewidth,height=\imagewidth,clip]{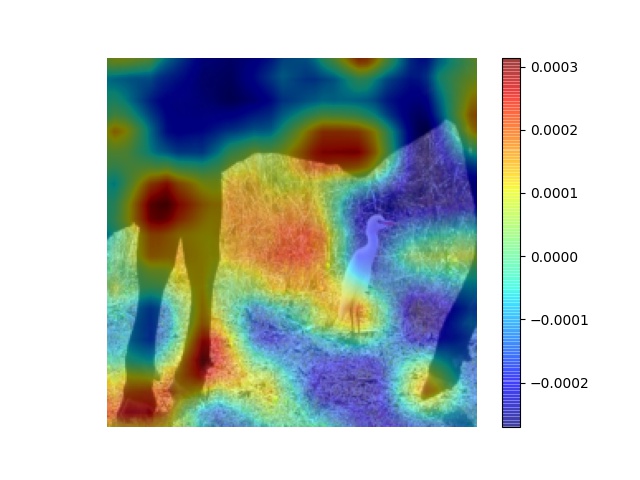}
   \includegraphics[trim=77 38 117 43,width=\imagewidth,height=\imagewidth,clip]{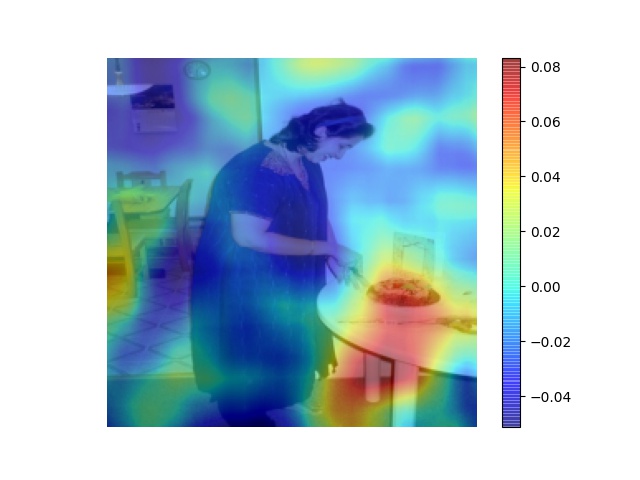}
   \includegraphics[trim=77 38 117 43,width=\imagewidth,height=\imagewidth,clip]{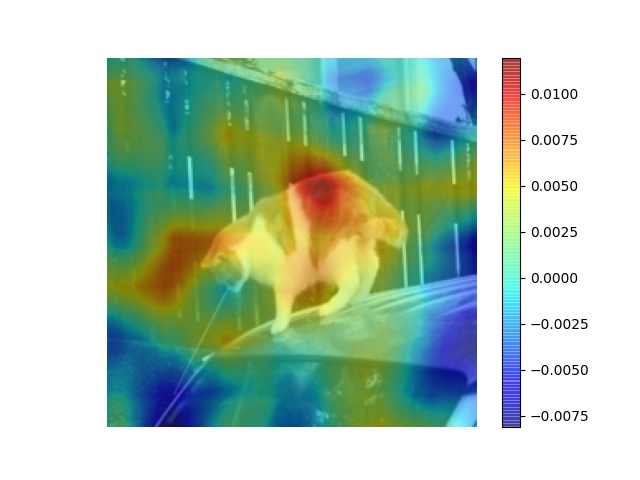}
   \includegraphics[trim=77 38 117 43,width=\imagewidth,height=\imagewidth,clip]{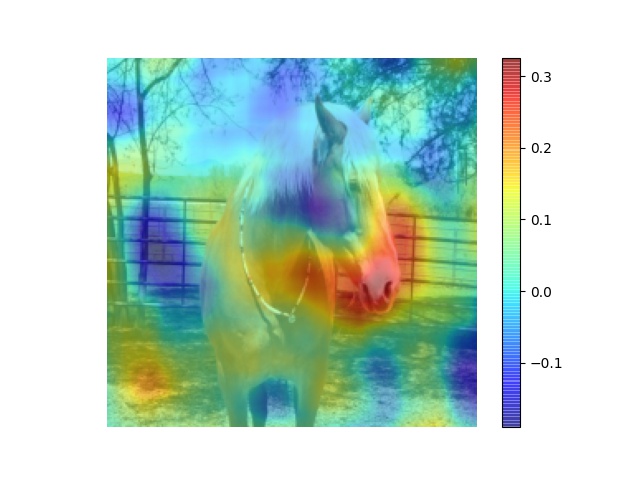}
   \includegraphics[trim=77 38 117 43,width=\imagewidth,height=\imagewidth,clip]{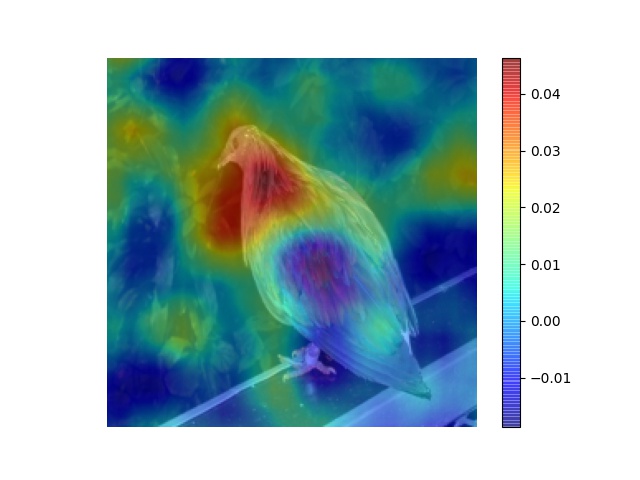}
   \includegraphics[trim=77 38 117 43,width=\imagewidth,height=\imagewidth,clip]{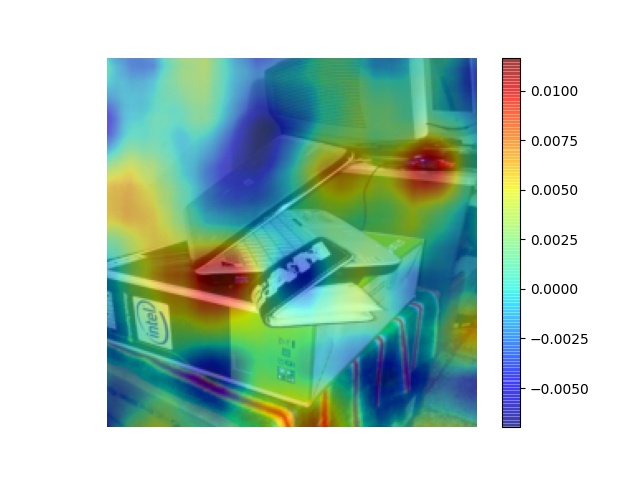}
   \caption{RISE}%
   % \label{fig:motorbike:rise}
 \end{subfigure}
 \caption{Examples of saliency maps that are successfully refined by BOREx.  The labels used in each explanation are: ``aeroplane'', ``cow'', ``dining table'', ``cat'', ``horse'', ``bird'', and ``TV monitor'', from the first row.}%
 \label{fig:good:2}
\end{figure}

\begin{figure}[t]
 \begin{subfigure}[t]{.32\linewidth}\centering
   \includegraphics[trim=39 36 35 35,width=\imagewidth,height=\imagewidth,clip]{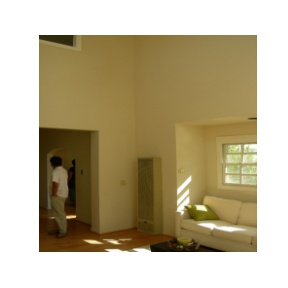}\\
   \includegraphics[trim=39 36 35 35,width=\imagewidth,height=\imagewidth,clip]{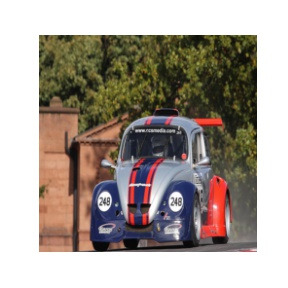}\\
   \includegraphics[trim=39 36 35 35,width=\imagewidth,height=\imagewidth,clip]{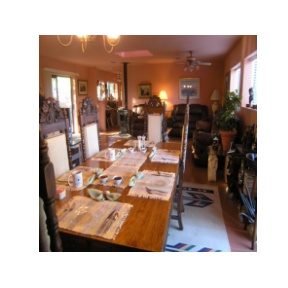}\\
   \includegraphics[trim=39 36 35 35,width=\imagewidth,height=\imagewidth,clip]{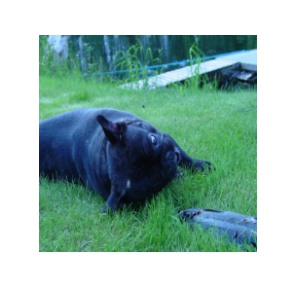}\\
   \includegraphics[trim=39 36 35 35,width=\imagewidth,height=\imagewidth,clip]{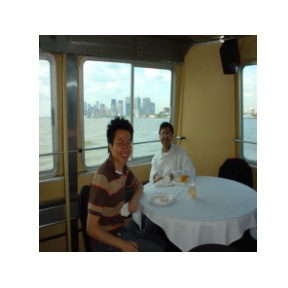}\\
   \includegraphics[trim=39 36 35 35,width=\imagewidth,height=\imagewidth,clip]{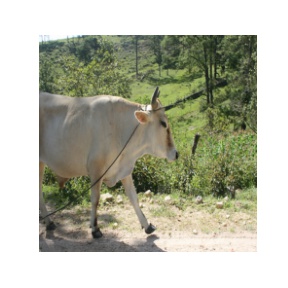}\\
   \includegraphics[trim=39 36 35 35,width=\imagewidth,height=\imagewidth,clip]{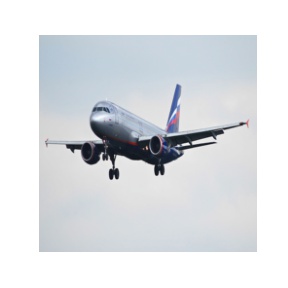}
   \caption{Input}%
   % \label{fig:motorbike:image}
 \end{subfigure}
 \hfill
 \begin{subfigure}[t]{.32\linewidth}\centering
   \includegraphics[trim=77 38 117 43,width=\imagewidth,height=\imagewidth,clip]{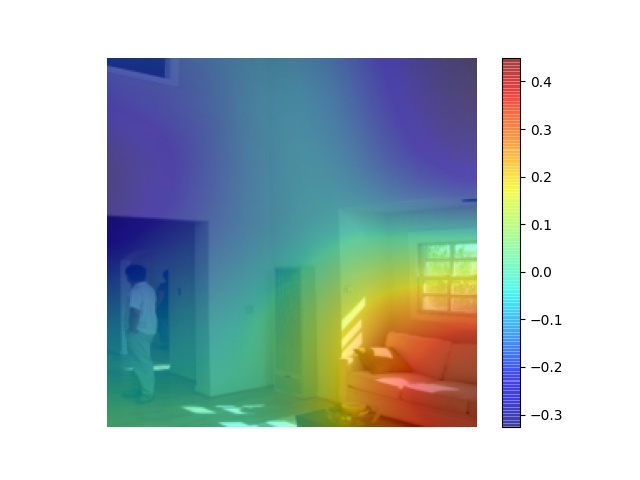}
   \includegraphics[trim=77 38 117 43,width=\imagewidth,height=\imagewidth,clip]{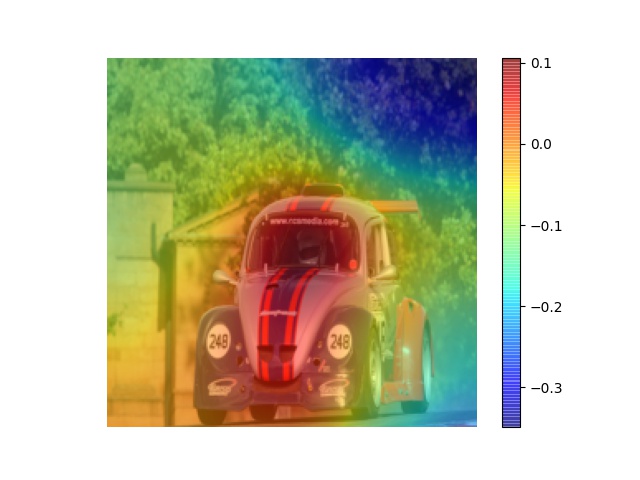}
   \includegraphics[trim=77 38 117 43,width=\imagewidth,height=\imagewidth,clip]{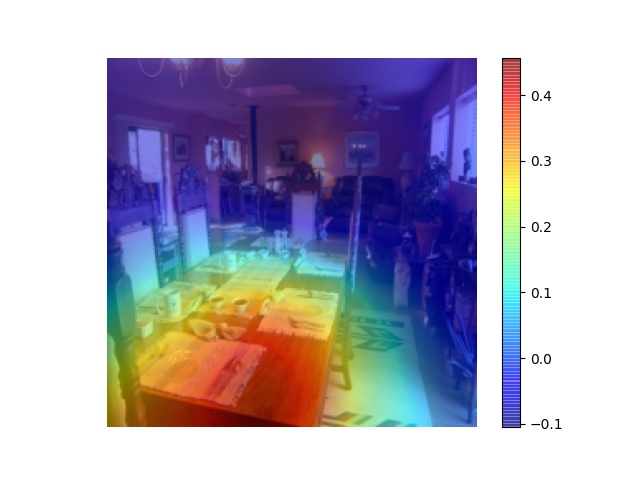}
   \includegraphics[trim=77 38 117 43,width=\imagewidth,height=\imagewidth,clip]{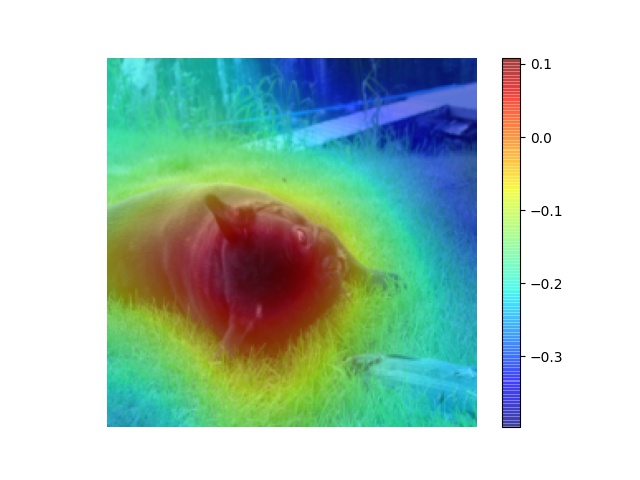}
   \includegraphics[trim=77 38 117 43,width=\imagewidth,height=\imagewidth,clip]{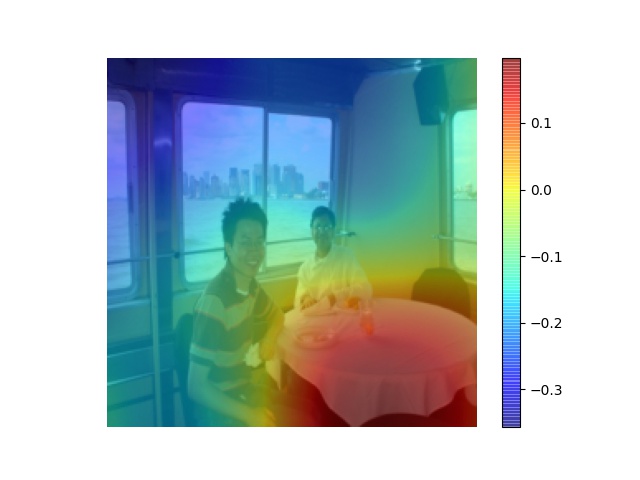}
   \includegraphics[trim=77 38 117 43,width=\imagewidth,height=\imagewidth,clip]{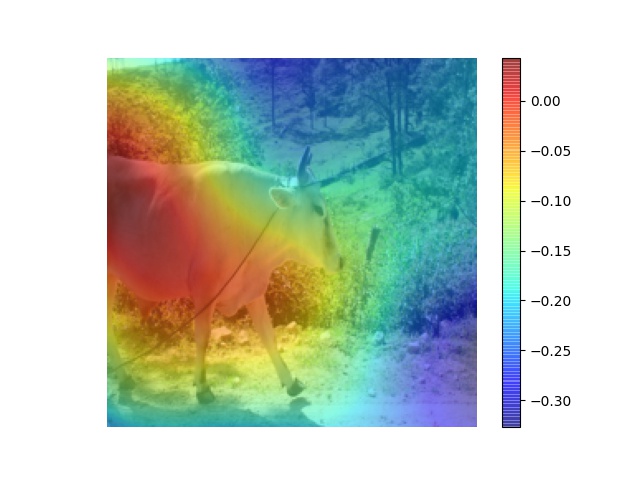}
   \includegraphics[trim=77 38 117 43,width=\imagewidth,height=\imagewidth,clip]{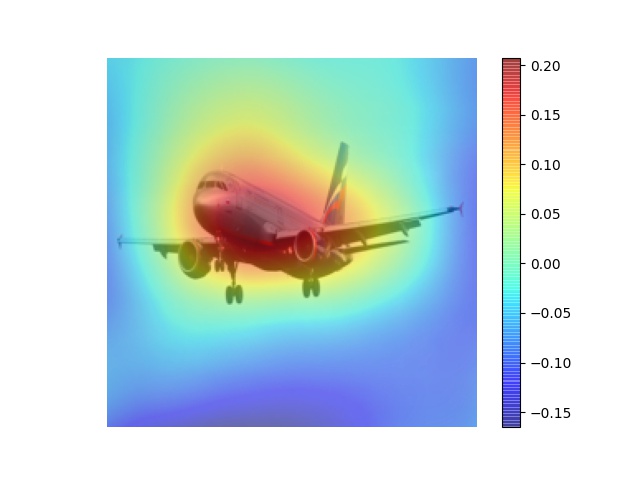}
   \caption{BOREx}%
   % \label{fig:motorbike:borex}
 \end{subfigure}
 \hfill
 \begin{subfigure}[t]{.32\linewidth}\centering
   \includegraphics[trim=77 38 117 43,width=\imagewidth,height=\imagewidth,clip]{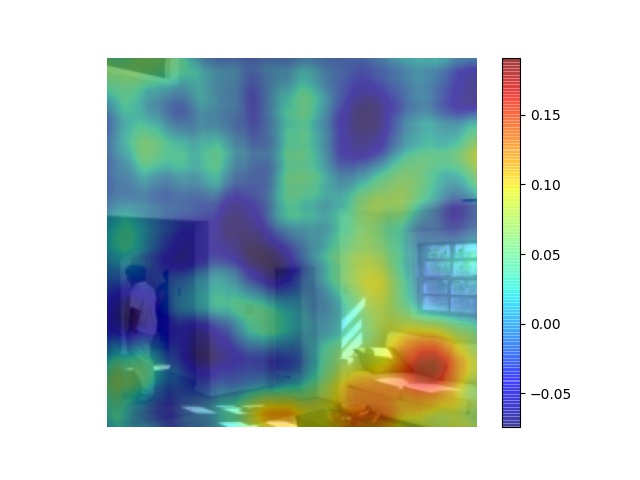}
   \includegraphics[trim=77 38 117 43,width=\imagewidth,height=\imagewidth,clip]{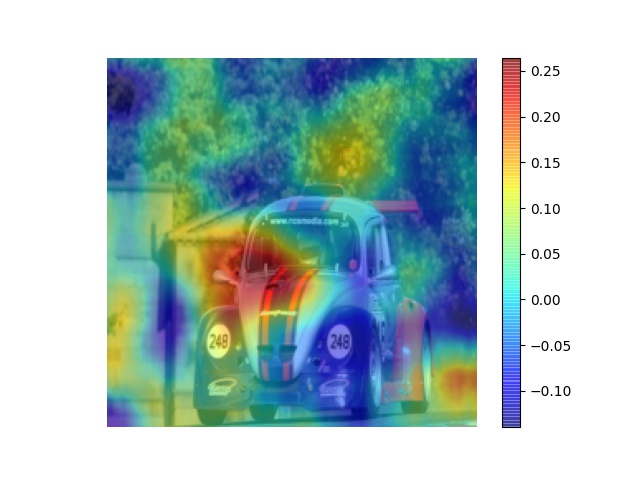}
   \includegraphics[trim=77 38 117 43,width=\imagewidth,height=\imagewidth,clip]{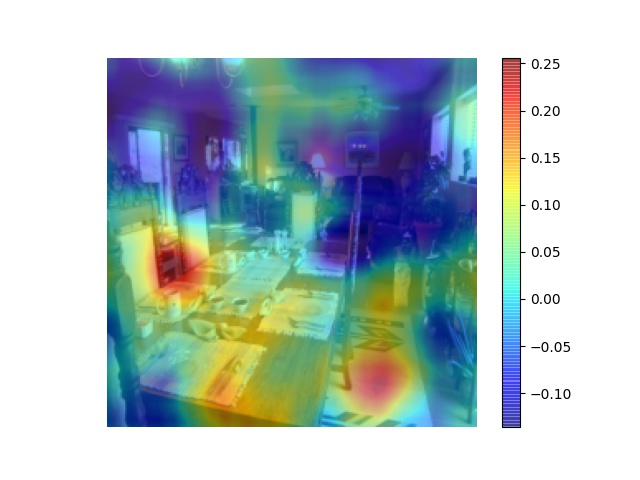}
   \includegraphics[trim=77 38 117 43,width=\imagewidth,height=\imagewidth,clip]{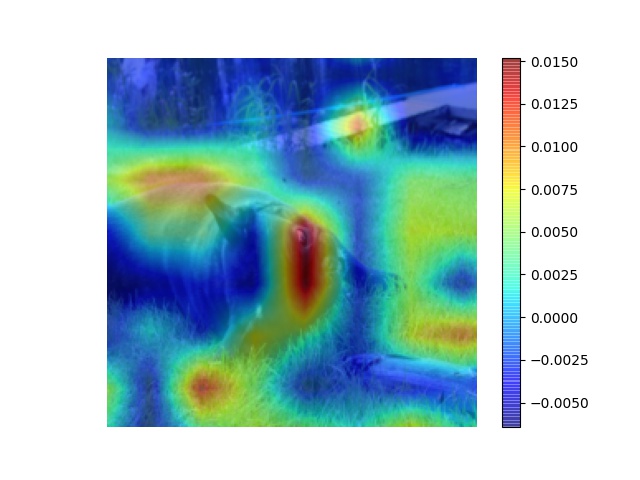}
   \includegraphics[trim=77 38 117 43,width=\imagewidth,height=\imagewidth,clip]{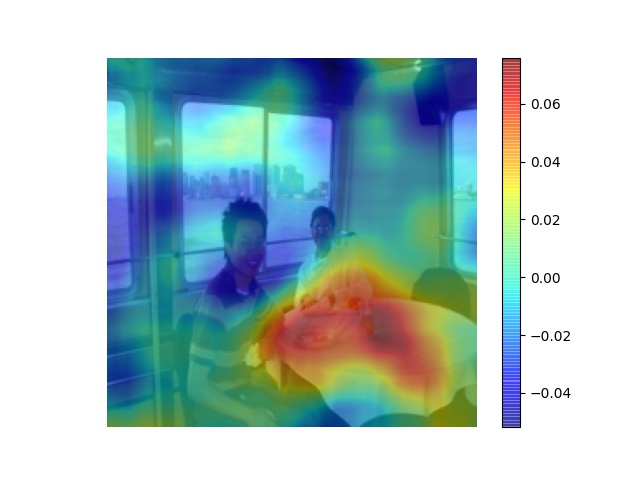}
   \includegraphics[trim=77 38 117 43,width=\imagewidth,height=\imagewidth,clip]{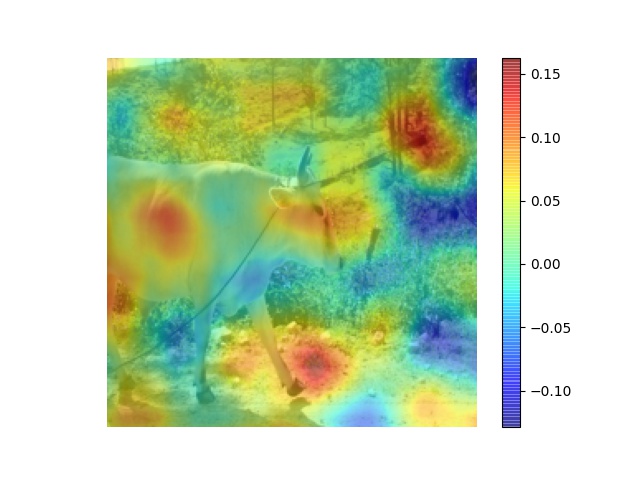}
   \includegraphics[trim=77 38 117 43,width=\imagewidth,height=\imagewidth,clip]{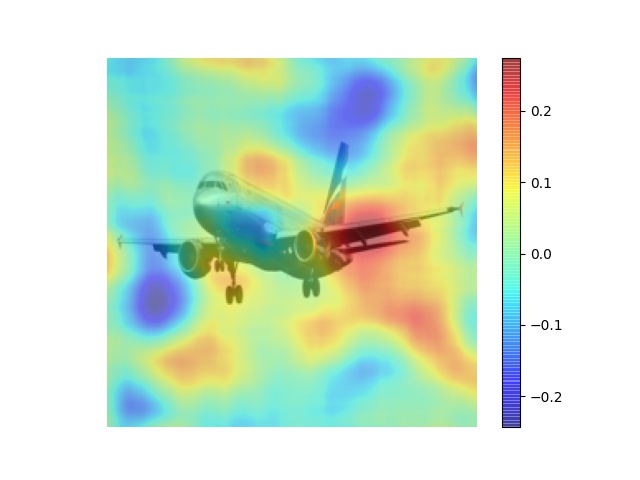}
   \caption{RISE}%
   % \label{fig:motorbike:rise}
 \end{subfigure}
 \caption{Examples of saliency maps that are successfully refined by BOREx.  The labels used in each explanation are: ``sofa'', ``car'', ``dining table'', ``dog'', ``dining table'', ``cow'', and ``aero plane'', from the first row.}%
 \label{fig:good:3}
\end{figure}

\begin{figure}[t]
 \begin{subfigure}[t]{.32\linewidth}\centering
   \includegraphics[trim=39 36 35 35,width=\imagewidth,height=\imagewidth,clip]{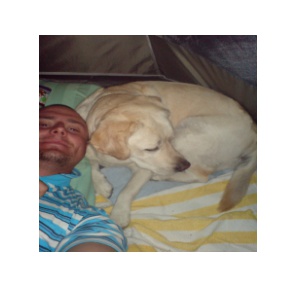}\\
   \includegraphics[trim=39 36 35 35,width=\imagewidth,height=\imagewidth,clip]{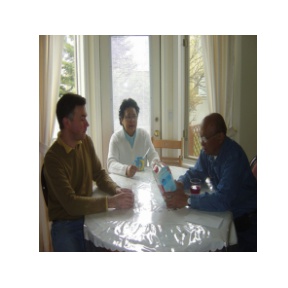}\\
   \includegraphics[trim=39 36 35 35,width=\imagewidth,height=\imagewidth,clip]{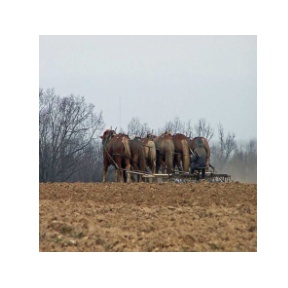}\\
   \includegraphics[trim=39 36 35 35,width=\imagewidth,height=\imagewidth,clip]{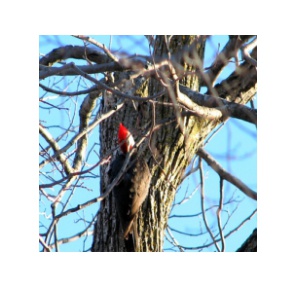}\\
   \includegraphics[trim=39 36 35 35,width=\imagewidth,height=\imagewidth,clip]{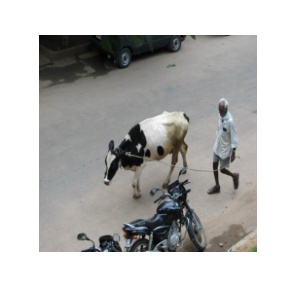}
   \caption{Input}%
   % \label{fig:motorbike:image}
 \end{subfigure}
 \hfill
 \begin{subfigure}[t]{.32\linewidth}\centering
   \includegraphics[trim=77 38 117 43,width=\imagewidth,height=\imagewidth,clip]{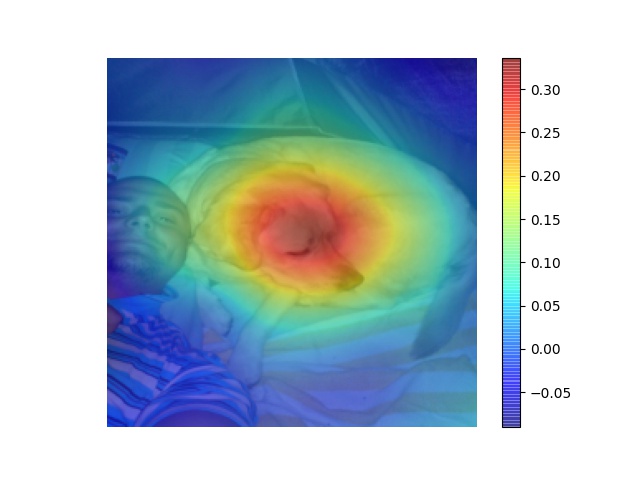}
   \includegraphics[trim=77 38 117 43,width=\imagewidth,height=\imagewidth,clip]{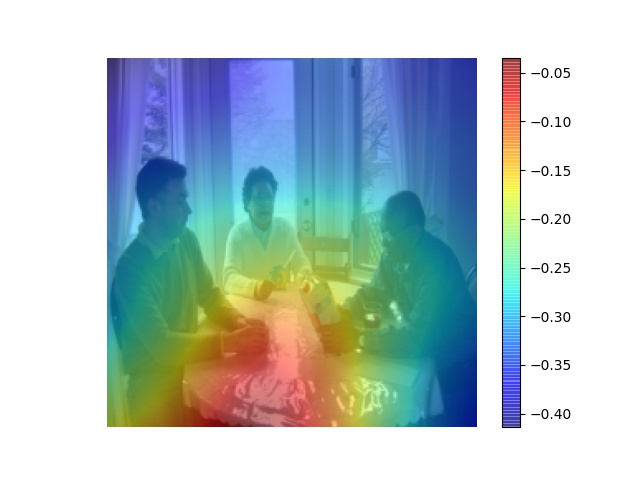}
   \includegraphics[trim=77 38 117 43,width=\imagewidth,height=\imagewidth,clip]{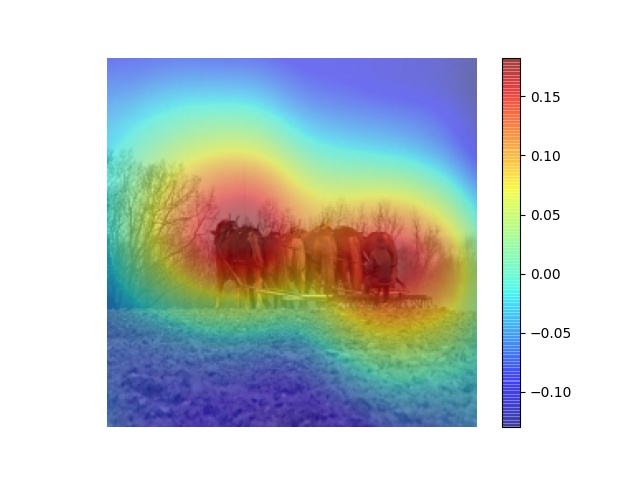}
   \includegraphics[trim=77 38 117 43,width=\imagewidth,height=\imagewidth,clip]{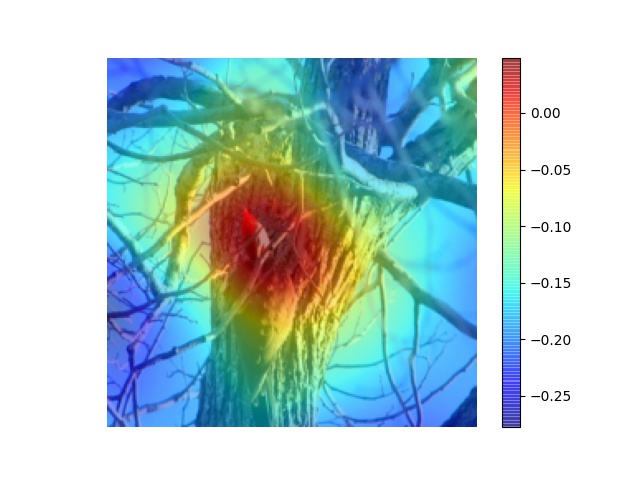}
   \includegraphics[trim=77 38 117 43,width=\imagewidth,height=\imagewidth,clip]{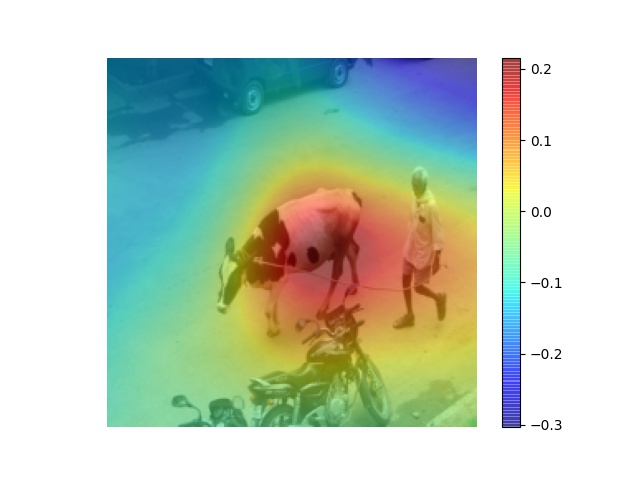}
   \caption{BOREx}%
   % \label{fig:motorbike:borex}
 \end{subfigure}
 \hfill
 \begin{subfigure}[t]{.32\linewidth}\centering
   \includegraphics[trim=77 38 117 43,width=\imagewidth,height=\imagewidth,clip]{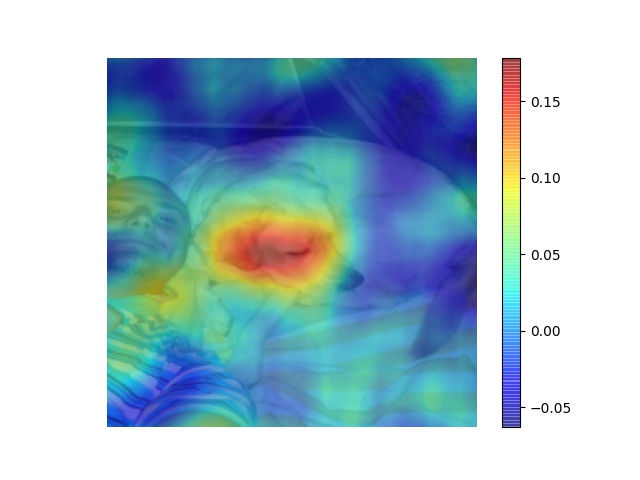}
   \includegraphics[trim=77 38 117 43,width=\imagewidth,height=\imagewidth,clip]{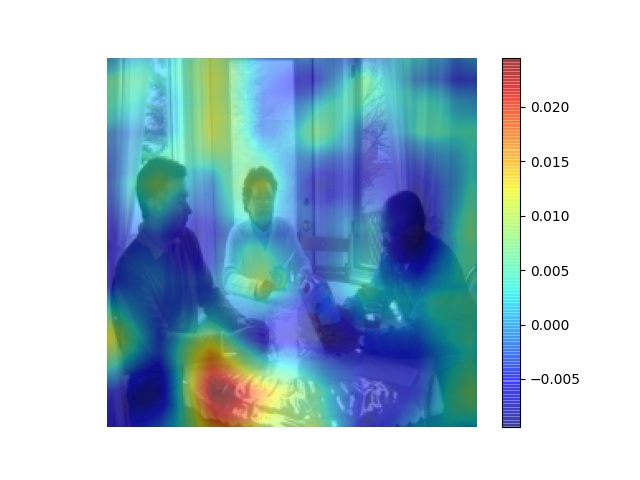}
   \includegraphics[trim=77 38 117 43,width=\imagewidth,height=\imagewidth,clip]{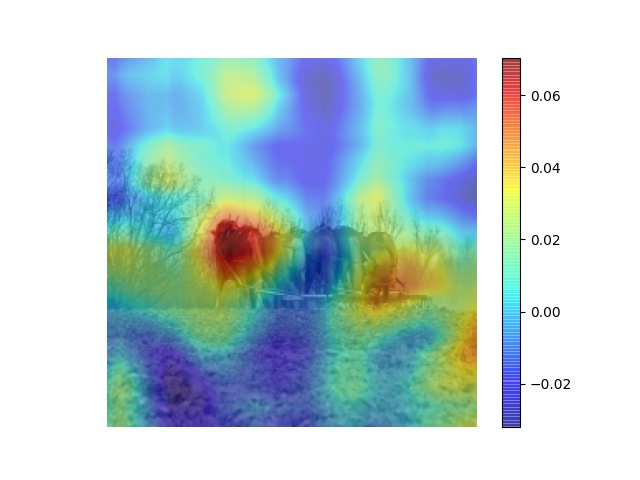}
   \includegraphics[trim=77 38 117 43,width=\imagewidth,height=\imagewidth,clip]{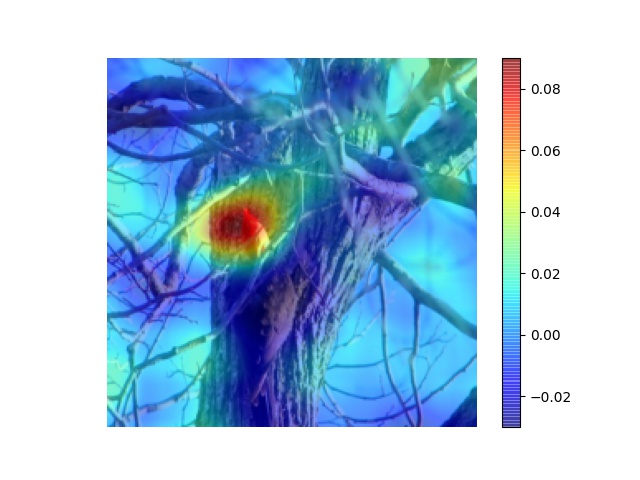}
   \includegraphics[trim=77 38 117 43,width=\imagewidth,height=\imagewidth,clip]{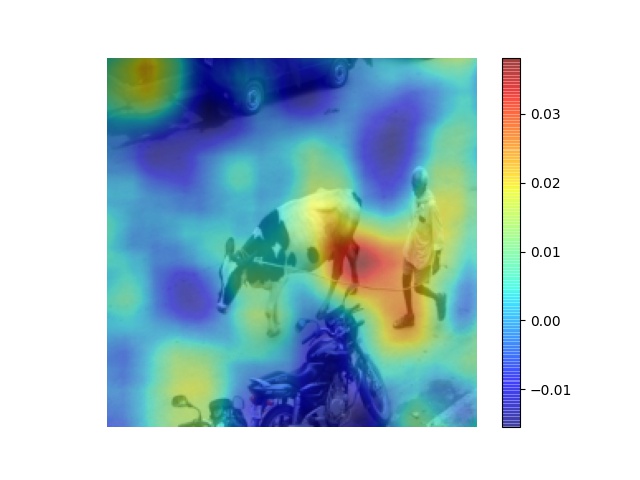}
   \caption{RISE}%
   % \label{fig:motorbike:rise}
 \end{subfigure}
 \caption{Examples of saliency maps that are successfully refined by BOREx.  The labels used in each explanation are: ``dog'', ``dining table'', ``horse'', ``bird'', and ``cow'', from the first row.}%
 \label{fig:good:4}
\end{figure}

Figures~\ref{fig:good:1}--\ref{fig:good:4} present examples of saliency maps generated by BOREx.
We picked several examples in which BOREx successfully refines the quality of input images measured in the quantitative metrics.
Compared to the input images generated by RISE, the saliency maps generated by BOREx localize the important regions better and less noisy.

\subsection{Examples in which BOREx degraded the quality of input images}

Figure~\ref{fig:bad:1} presents the examples in which BOREx degraded the input images measured in the quantitative metrics.
We add explanations for each example.

\begin{figure}[t]
  \begin{subfigure}[t]{.32\linewidth}\centering
    \includegraphics[trim=39 36 35 35,width=\imagewidth,height=\imagewidth,clip]{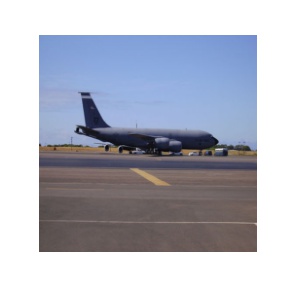}\\
    \includegraphics[trim=39 36 35 35,width=\imagewidth,height=\imagewidth,clip]{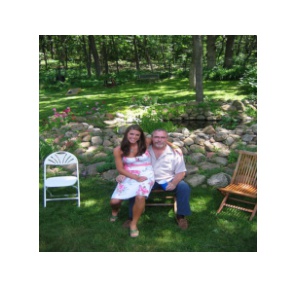}\\
    \includegraphics[trim=39 36 35 35,width=\imagewidth,height=\imagewidth,clip]{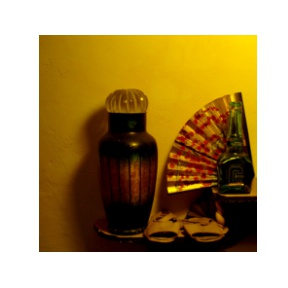}\\
    \caption{Input}%
    % \label{fig:motorbike:image}
 \end{subfigure}
 \hfill
 \begin{subfigure}[t]{.32\linewidth}\centering
   \includegraphics[trim=77 38 117 43,width=\imagewidth,height=\imagewidth,clip]{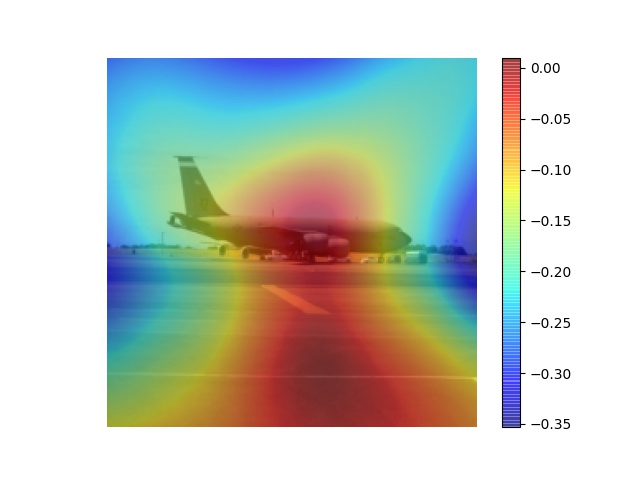}\\
   \includegraphics[trim=77 38 117 43,width=\imagewidth,height=\imagewidth,clip]{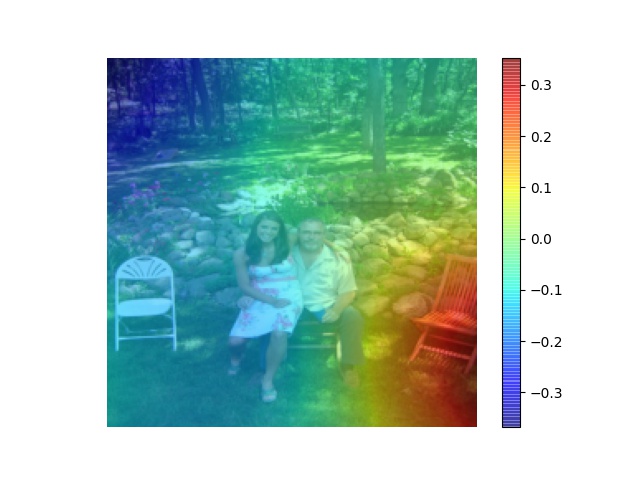}\\
   \includegraphics[trim=77 38 117 43,width=\imagewidth,height=\imagewidth,clip]{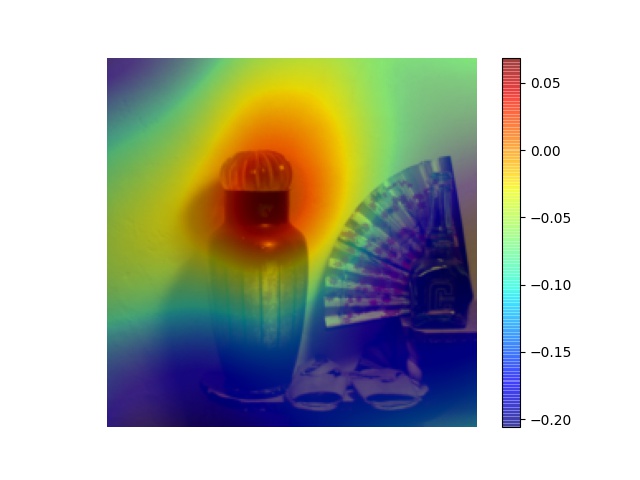}\\
   \caption{BOREx}%
   % \label{fig:motorbike:borex}
 \end{subfigure}
 \hfill
 \begin{subfigure}[t]{.32\linewidth}\centering
   \includegraphics[trim=77 38 117 43,width=\imagewidth,height=\imagewidth,clip]{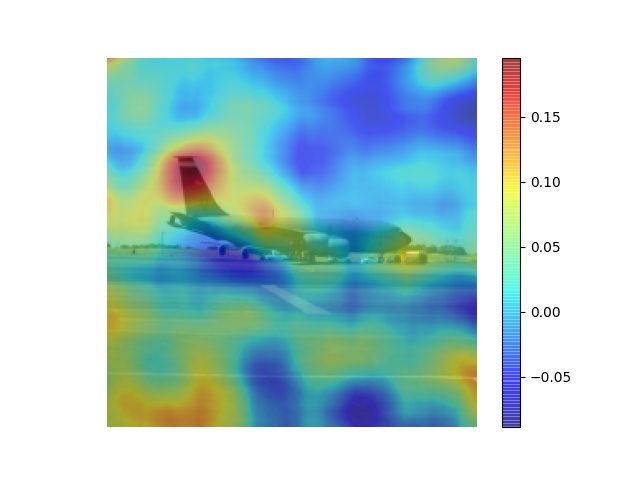}\\
   \includegraphics[trim=77 38 117 43,width=\imagewidth,height=\imagewidth,clip]{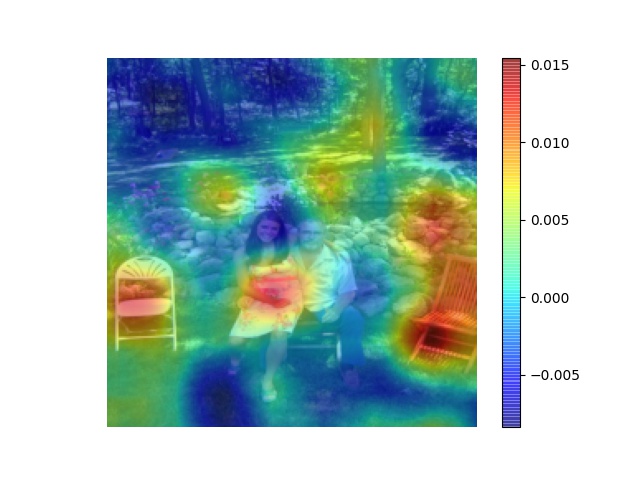}\\
   \includegraphics[trim=77 38 117 43,width=\imagewidth,height=\imagewidth,clip]{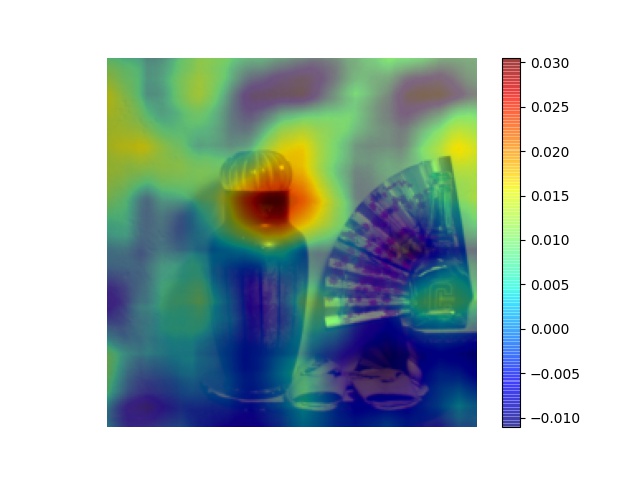}\\
   \caption{RISE}%
   % \label{fig:motorbike:rise}
 \end{subfigure}
 \caption{Examples of saliency maps that BOREx degraded the quantitative metric of the input image synthesized by RISE.  The labels used in each explanation are: ``warplane'', ``park bench'', and ``water bottle'', from the first row.}%
 \label{fig:bad:1}
\end{figure}

\KS{Fix the label names looking at result1.csv.}

The first row in Figure~\ref{fig:bad:1}, which BOREx degraded the insertion metric, presents an example in which RISE successfully identifies the aeroplane, whereas BOREx wrongly identifies the ground in addition to the aeroplane.
This is caused by the saliency map produced by RISE used as the prior; in the prior, the saliency of the ground in the image is high, which misled~\iffull\ref{alg:gpBOREx}\else\ref{main:alg:gpBOREx}\fi.

The saliency maps in the second row of Figure~\ref{fig:bad:1} are generated by the label ``chair''.
BOREx identifies one of the chairs in the image, but not the other chair, degrading the insertion metric because identifying two chairs are needed to recognize a park bench.
This is due to the issue of the limited shape of the masks used by BOREx discussed in \iffull\cref{sec:results}\else\cref{main:sec:results}\fi.
RISE looks successfully identifies both chairs.

BOREx degraded the F-measure metric for the image in the third row of Figure~\ref{fig:bad:1}.
The important region identified by BOREx concentrates around the lid of the bottle, whereas the region identified by RISE exists also on the body of the bottle.
The PascalVOC dataset specifies the entire bottle as the correct answer, which leads to the poor value in the F-measure metric for the saliency map generated by BOREx.
Deciding from the insertion and the deletion metrics, we guess that the model indeed considers the lid part as the salient region.

\section{Comparison with Grad-CAM++}

\begin{figure}[t]
  \begin{subfigure}[t]{.32\linewidth}\centering
    \includegraphics[trim=39 36 35 35,width=\imagewidth,height=\imagewidth,clip]{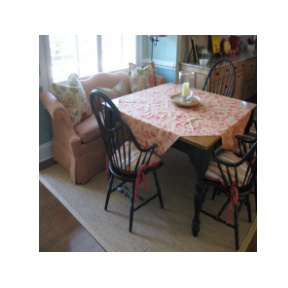}
    \includegraphics[trim=39 36 35 35,width=\imagewidth,height=\imagewidth,clip]{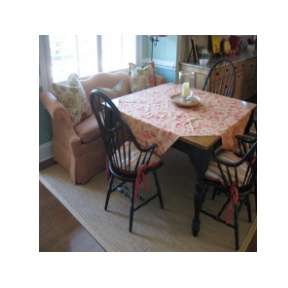}
    \includegraphics[trim=39 36 35 35,width=\imagewidth,height=\imagewidth,clip]{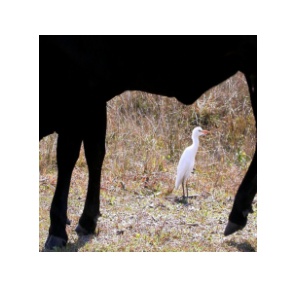}
    \includegraphics[trim=39 36 35 35,width=\imagewidth,height=\imagewidth,clip]{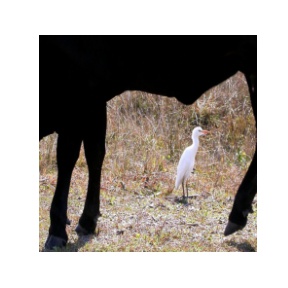}
    \includegraphics[trim=39 36 35 35,width=\imagewidth,height=\imagewidth,clip]{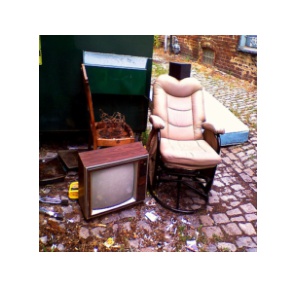}
    \includegraphics[trim=39 36 35 35,width=\imagewidth,height=\imagewidth,clip]{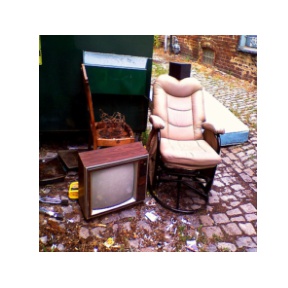}
     \includegraphics[trim=39 36 35 35,width=\imagewidth,height=\imagewidth,clip]{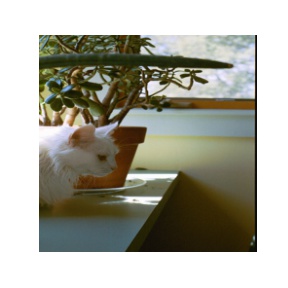}
   \caption{Input}%
   % \label{fig:motorbike:image}
 \end{subfigure}
 \hfill
 \begin{subfigure}[t]{.32\linewidth}\centering
    \includegraphics[trim=77 38 117 43,width=\imagewidth,height=\imagewidth,clip]{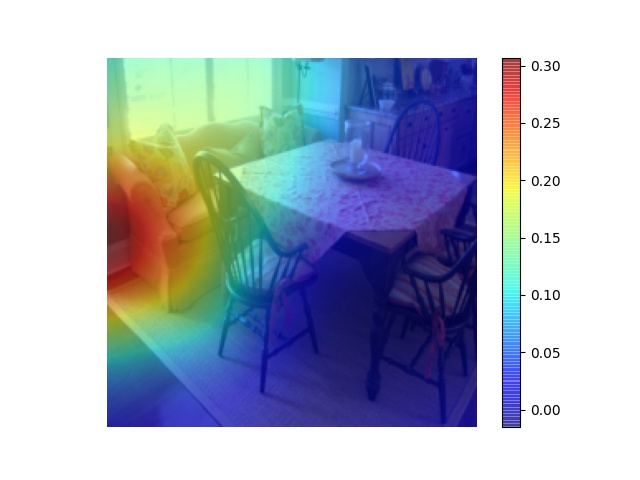}
    \includegraphics[trim=77 38 117 43,width=\imagewidth,height=\imagewidth,clip]{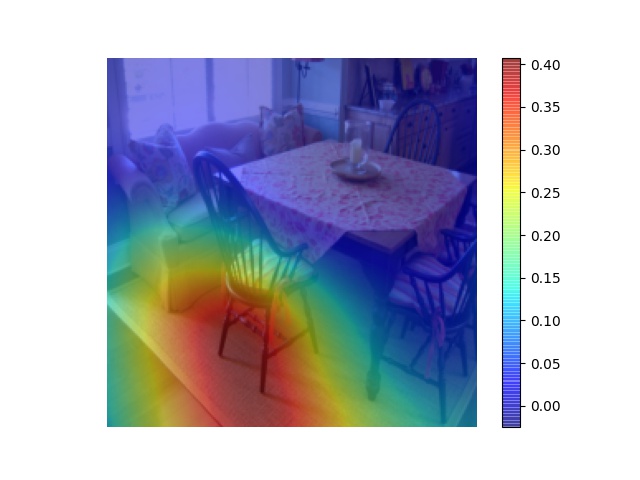}
    \includegraphics[trim=77 38 117 43,width=\imagewidth,height=\imagewidth,clip]{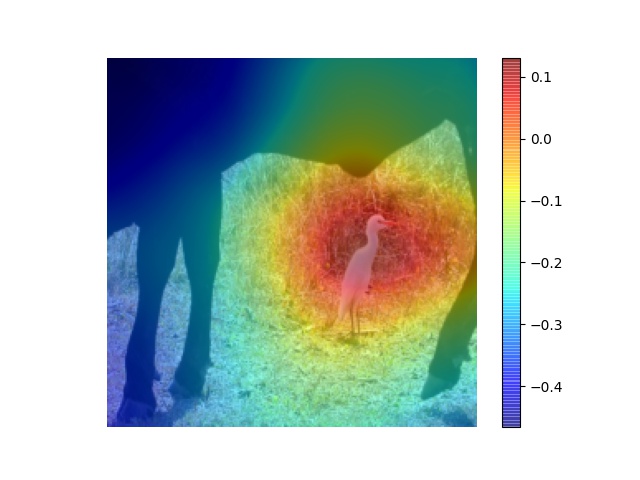}
    \includegraphics[trim=77 38 117 43,width=\imagewidth,height=\imagewidth,clip]{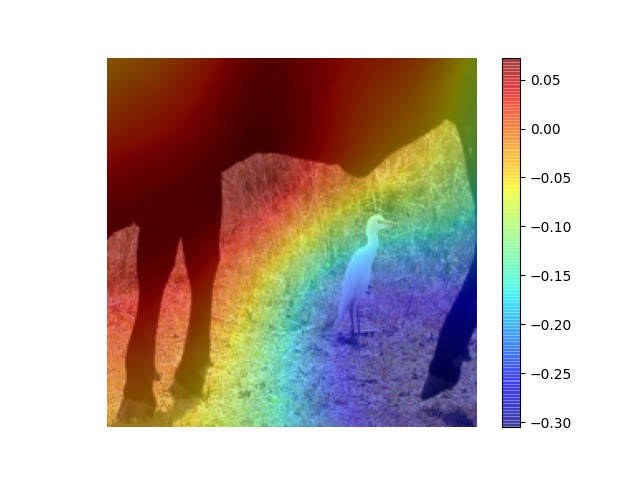}
    \includegraphics[trim=77 38 117 43,width=\imagewidth,height=\imagewidth,clip]{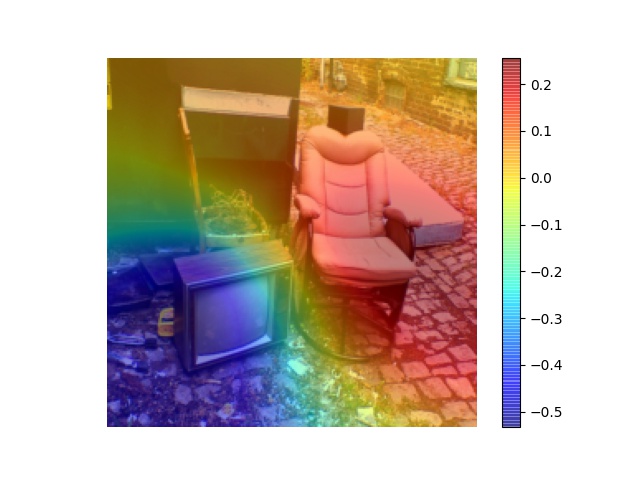}
    \includegraphics[trim=77 38 117 43,width=\imagewidth,height=\imagewidth,clip]{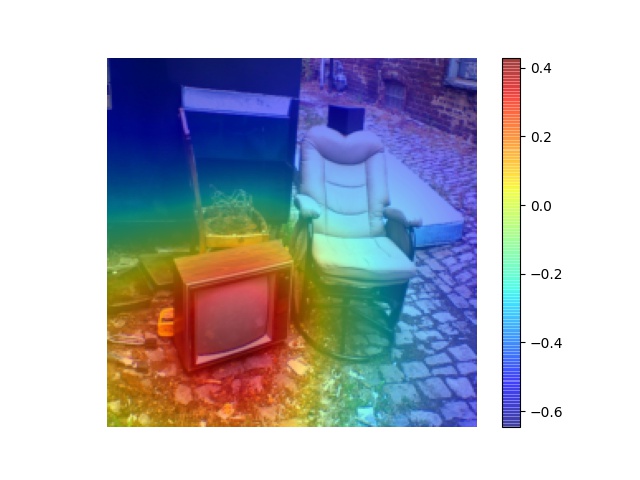}
    \includegraphics[trim=77 38 117 43,width=\imagewidth,height=\imagewidth,clip]{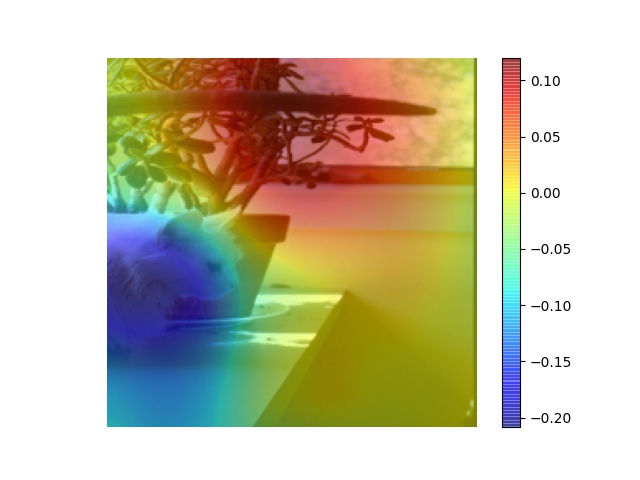}    \caption{BOREx}%
    % \label{fig:motorbike:borex}
 \end{subfigure}
 \hfill
 \begin{subfigure}[t]{.32\linewidth}\centering
    \includegraphics[trim=77 38 117 43,width=\imagewidth,height=\imagewidth,clip]{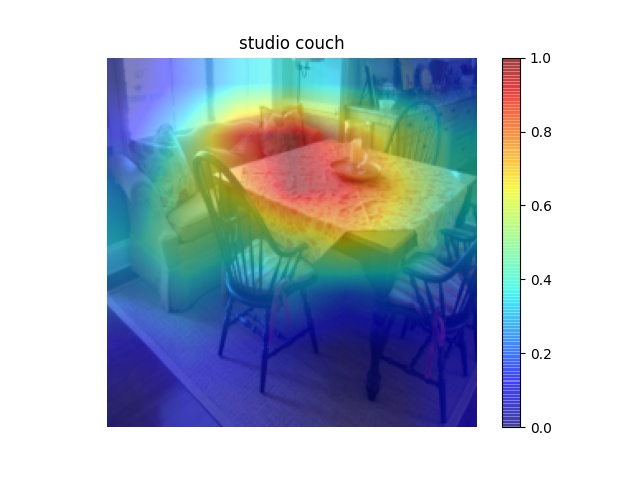}
    \includegraphics[trim=77 38 117 43,width=\imagewidth,height=\imagewidth,clip]{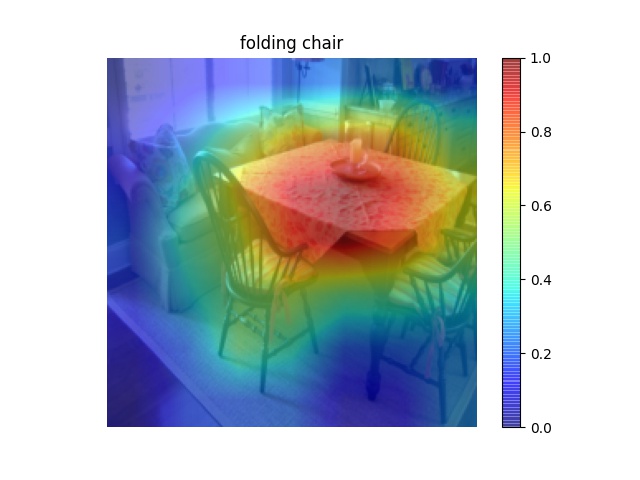}
    \includegraphics[trim=77 38 117 43,width=\imagewidth,height=\imagewidth,clip]{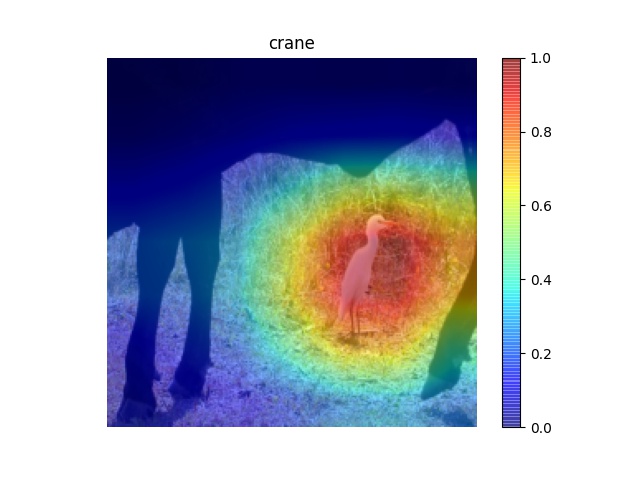}
    \includegraphics[trim=77 38 117 43,width=\imagewidth,height=\imagewidth,clip]{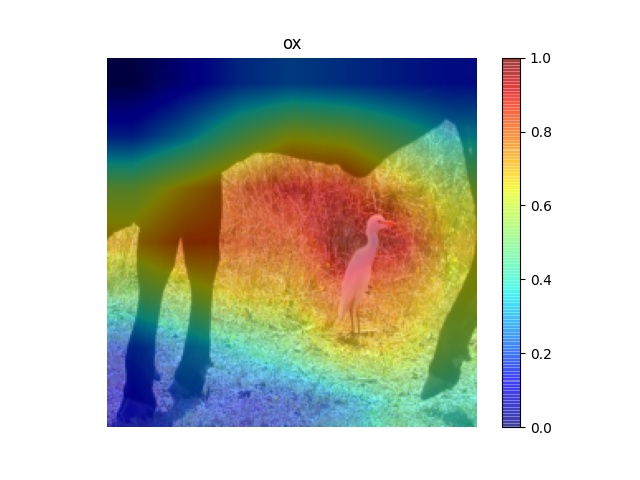}
    \includegraphics[trim=77 38 117 43,width=\imagewidth,height=\imagewidth,clip]{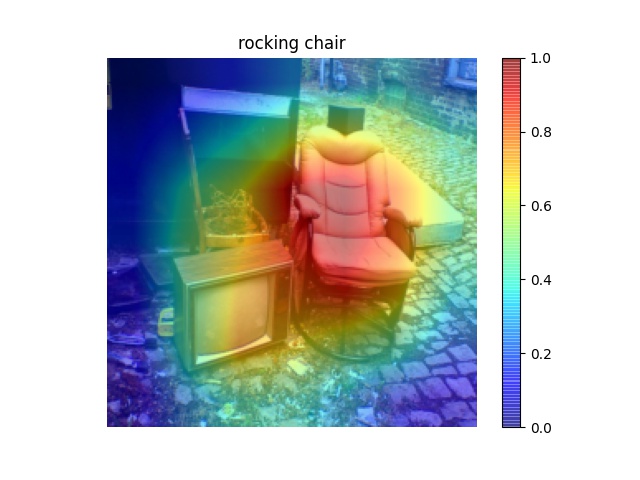}
    \includegraphics[trim=77 38 117 43,width=\imagewidth,height=\imagewidth,clip]{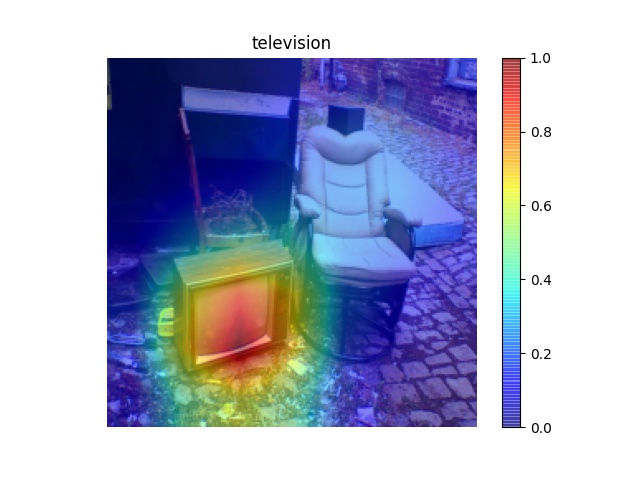}
    \includegraphics[trim=77 38 117 43,width=\imagewidth,height=\imagewidth,clip]{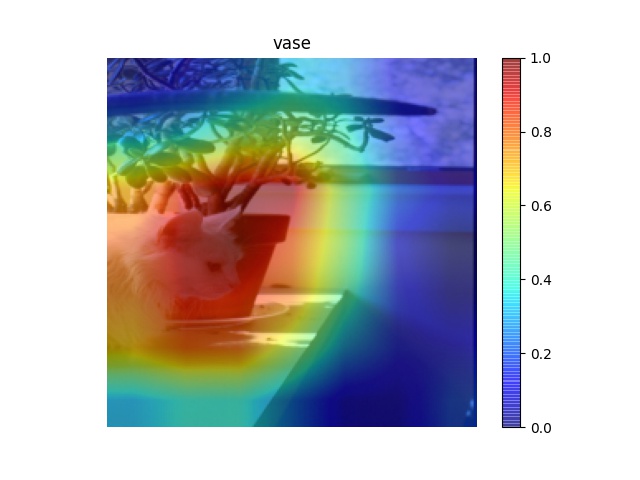}       \caption{Grad-CAM++}%
    % \label{fig:motorbike:borex}
 \end{subfigure}
 \caption{Examples of saliency maps in which BOREx outperforms GradCAM++.  The labels used in each explanation are: ``sofa'', ``chair'', ``bird'', ``cow'', ``chair'', ``TV monitor'', and ``potted plant'', from the first row.}%
 \label{fig:good:gradcampp:insertion}
\end{figure}

Although the statistical tests in Section~\iffull\ref{sec:experiments}\else\ref{main:sec:experiments}\fi do not conclude the effectiveness of BOREx measured in the insertion and the F-measure metrics to refine the saliency map produced by Grad-CAM++, there are some instances that indeed benefit from BOREx.
Figure~\ref{fig:good:gradcampp:insertion} presents several examples of the saliency maps in which BOREx outperforms Grad-CAM++.

Interestingly, if we let BOREx and Grad-CAM++ produce saliency maps for the same image with different labels, the saliency maps produced by Grad-CAM++ are often less sensitive to the change in the label than BOREx.
For example, the first (resp., the second) row in Figure~\ref{fig:good:gradcampp:insertion} are the saliency maps produced by BOREx and Grad-CAM++ with label ``sofa'' (resp., ``chair'').
The saliency maps produced by BOREx correctly identify the region that is important for classifying the image to the given label; however, the saliency maps produced by Grad-CAM++ is less focused to the given label than BOREx.

\section{Examples of saliency maps for video classifiers}
\KS{More examples of video saliency maps.}

\begin{figure*}[t]
  \begin{minipage}[t]{.50\linewidth}
    \begin{subfigure}[t]{\linewidth}\centering
      \includegraphics[width=\linewidth]{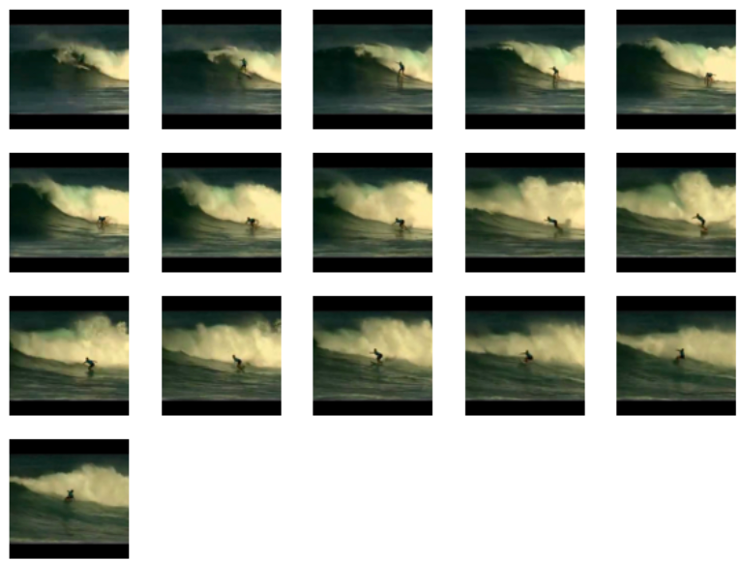}
      \caption{Input}%
      % \label{fig:motorbike:image}
    \end{subfigure}\\
    \begin{subfigure}[t]{\linewidth}\centering
      \includegraphics[width=\linewidth]{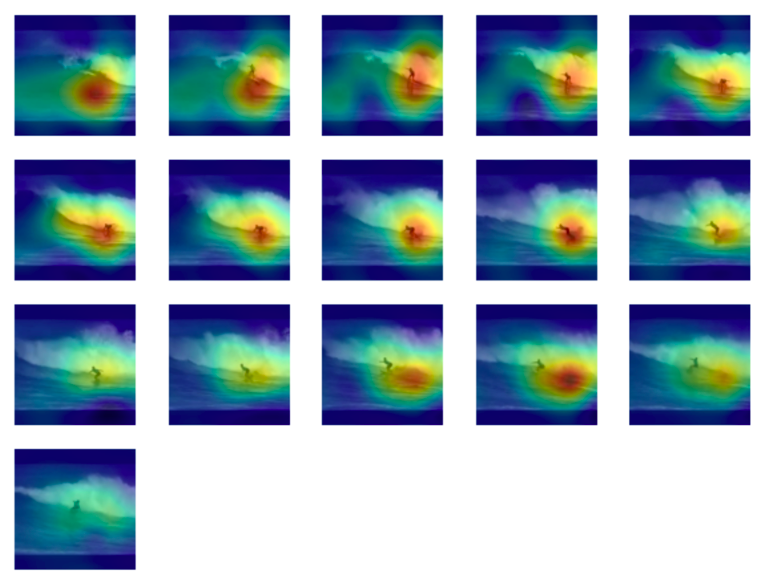}
      \caption{BOREx}%
      % \label{fig:motorbike:borex}
    \end{subfigure}\\
  \end{minipage}
  \begin{minipage}[t]{.50\linewidth}
    \begin{subfigure}[t]{\linewidth}\centering
      \includegraphics[width=\linewidth]{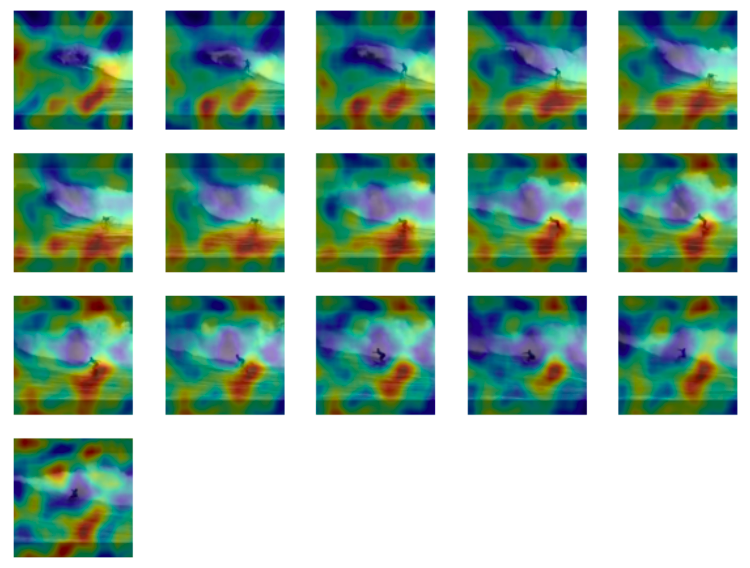}
      \caption{RISE}%
      % \label{fig:motorbike:borex}
    \end{subfigure}
    \begin{subfigure}[t]{\linewidth}\centering
      \includegraphics[trim=90 60 65 50,width=\linewidth,height=.75\linewidth,clip]{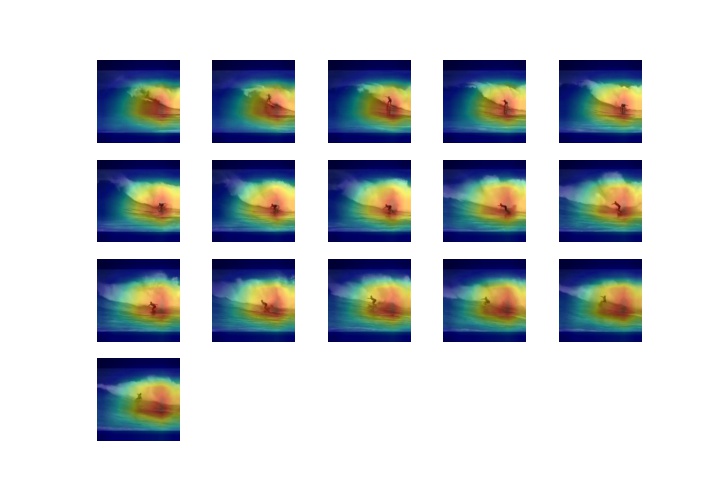}
      \caption{Grad-CAM++}%
      % \label{fig:motorbike:borex}
    \end{subfigure}
  \end{minipage}
  \caption{Examples of saliency maps for video classifiers with label ``surfing''.}%
  \label{fig:movie:surfing}
\end{figure*}

\begin{figure*}[t]
  \begin{minipage}[t]{.50\linewidth}
    \begin{subfigure}[t]{\linewidth}\centering
      \includegraphics[trim=70 40 70 40,width=\linewidth,clip]{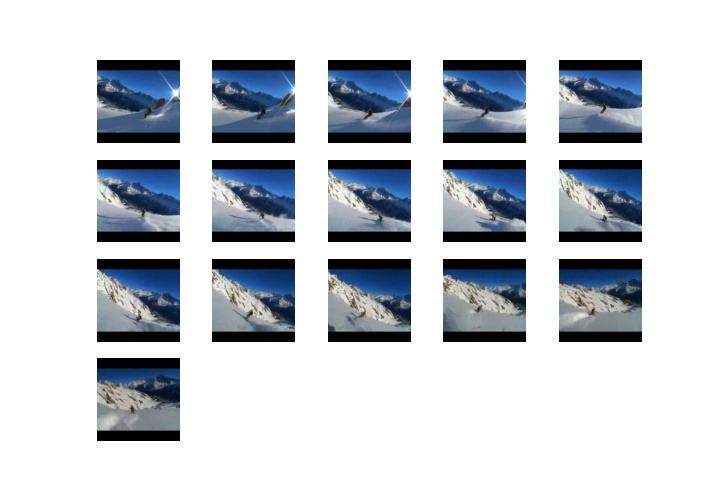}
      \caption{Input}%
      % \label{fig:mot&orbike:image}
    \end{subfigure}
    \begin{subfigure}[t]{\linewidth}\centering
      \includegraphics[trim=70 40 70 40,width=\linewidth,clip]{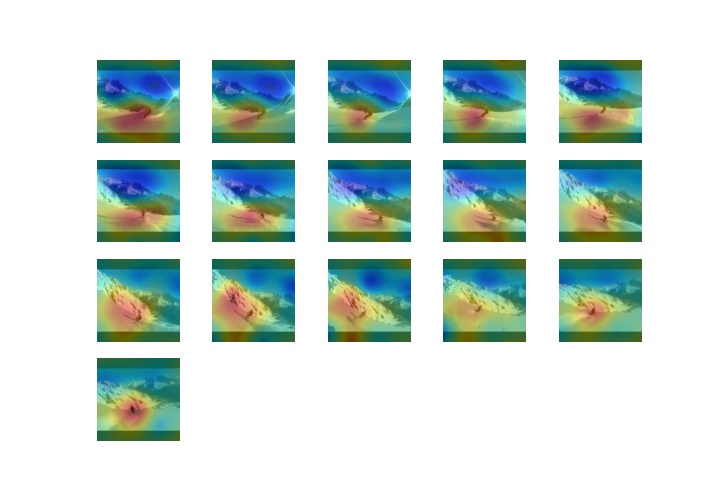}
      \caption{BOREx}%
      % \label{fig:motorbike:borex}
    \end{subfigure}
  \end{minipage}
  \begin{minipage}[t]{.50\linewidth}
    \begin{subfigure}[t]{\linewidth}\centering
      \includegraphics[trim=70 40 70 40,width=\linewidth,clip]{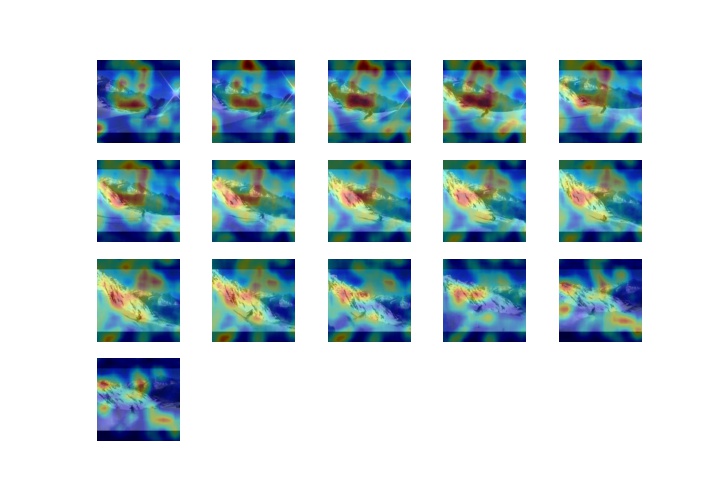}
      \caption{RISE}%
      % \label{fig:motorbike:borex}
    \end{subfigure}
    \begin{subfigure}[t]{\linewidth}\centering
      \includegraphics[trim=70 40 70 40,width=\linewidth,clip]{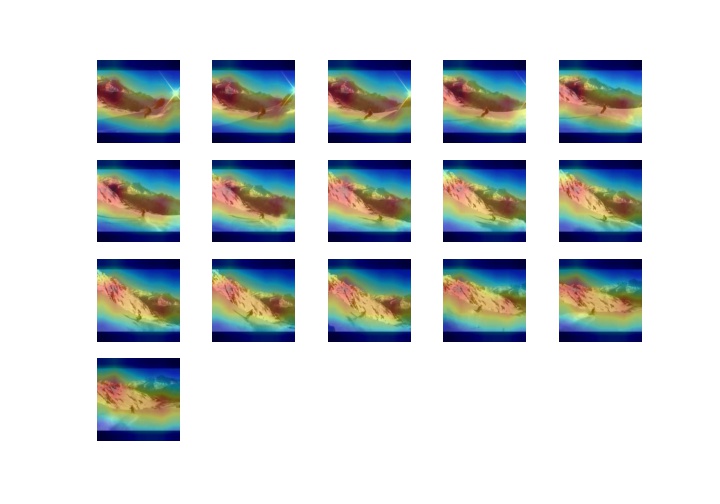}
      \caption{Grad-CAM++}%
      % \label{fig:motorbike:borex}
    \end{subfigure}
  \end{minipage}
  \caption{Examples of saliency maps for video classifiers with label ``skiing''.}%
  \label{fig:movie:skiing}
\end{figure*}

\begin{figure*}[t]
  \begin{minipage}[t]{.47\linewidth}
    \begin{subfigure}[t]{\linewidth}\centering
      \includegraphics[trim=70 40 70 40,width=\linewidth,clip]{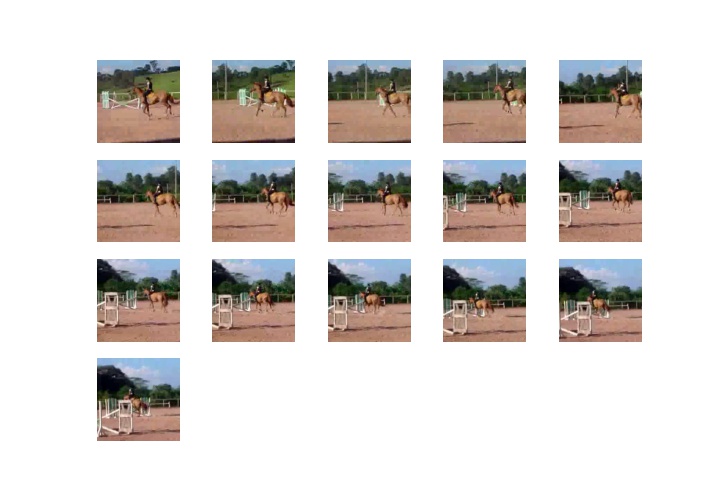}
      \caption{Input}%
      % \label{fig:motorbike:image}
    \end{subfigure}
    \begin{subfigure}[t]{\linewidth}\centering
      \includegraphics[trim=70 40 70 40,width=\linewidth,clip]{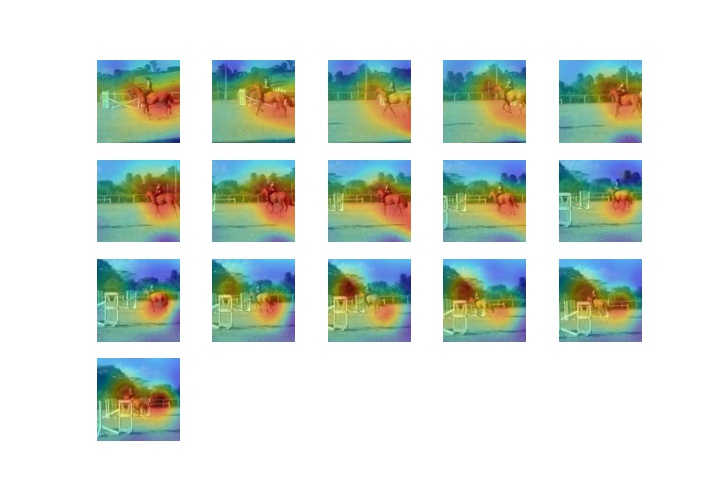}
      \caption{BOREx}%
      % \label{fig:motorbike:borex}
    \end{subfigure}
  \end{minipage}
  \begin{minipage}[t]{.47\linewidth}
    \begin{subfigure}[t]{\linewidth}\centering
      \includegraphics[trim=70 40 70 40,width=\linewidth,clip]{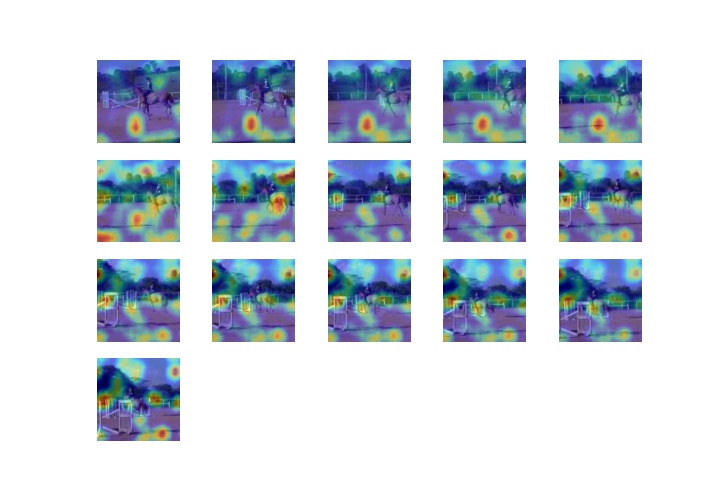}
      \caption{RISE}%
   % \label{fig:motorbike:borex}
    \end{subfigure}
    \begin{subfigure}[t]{\linewidth}\centering
      \includegraphics[trim=70 40 70 40,width=\linewidth,clip]{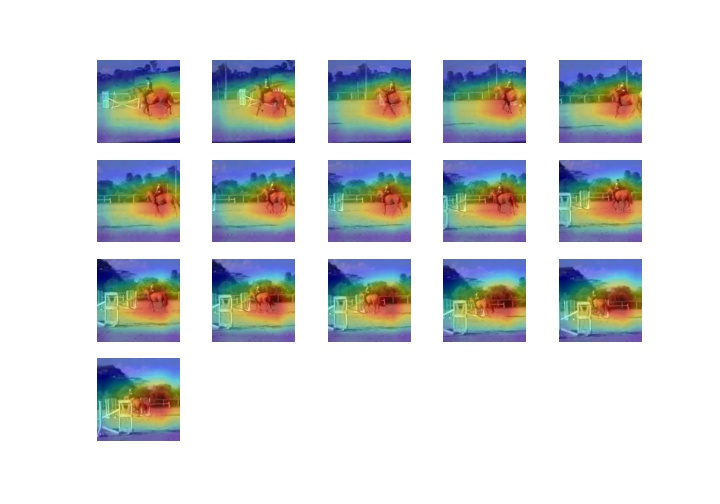}
      \caption{Grad-CAM++}%
      % \label{fig:motorbike:borex}
    \end{subfigure}
  \end{minipage}
  \caption{Examples of saliency maps for video classifiers with label ``horseriding''.}%
  \label{fig:movie:horseriding}
\end{figure*}

Figures~\ref{fig:movie:surfing}--\ref{fig:movie:horseriding} present examples of saliency maps for a video classifier produced by BOREx and a naive extension of RISE.
%
% BOREx produces a saliency map that is more focused on the important region in each frame compared to the one produced by RISE.
Compared with the saliency maps produced by RISE, the saliency maps generated by BOREx better localize the important parts.
It is also observed that the salinecy maps produced by BOREx are comparable to those produced by Grad-CAM++ in spite of the black-box nature of BOREx.
In Figure~\ref{fig:movie:skiing}, we can observe that BOREx follows the skier better than Grad-CAM++.

% \subsection{On RQ1}

% \MW{Maybe we can omit from here.}
% \cref{fig:motorbike:image} shows another example with two motorbikes.
% Since BOREx does not generate very small masks, the masks covering the left and the right motorbikes tend to have an intersecting area in between two motorbikes.
% In such a case, the pixels in the intersecting area is estimated to be salient, which results in the saliency map in \cref{fig:motorbike:borex}.\MW{Maybe we can omit by here.}
% These two behaviors are purely because of the limited search space of the mask generation, which is one of the hyper-parameters in BOREx.
% One of our future works is to improve the flexibility of the mask generation, e.g., to allow generation of multiple rectangles or increase the variation of the mask sizes.

\KS{Discuss computation time?}

% \section{On the definition of saliency}

% \pagelimitmarker{8}

% \begin{table}\small
%   \centering
%   \caption{}
%   \label{tab:AbrationStudy}
%   \begin{tabular}{lllr}
%   \toprule
%   Metric & Base. & Comp. & $p$-val. \\
%   \midrule
%     Deletion & no\_flip &  normal & 0.004182 \\
%          & square & normal & 0.348812 \\
%          & no\_normalize & normal & 0.383979 \\
%     \hline
%     F-meas. & no\_flip &  normal & 1.171645e-22 \\
%          & square & normal & 8.650028e-01 \\
%          & no\_normalize & normal & 0.112701 \\
%     \hline
%     Insertion & no\_flip &  normal & 9.170745e-01 \\
%          & square & normal & 1.0 \\
%          & no\_normalize & normal & 1.439060e-18 \\
%     \bottomrule
%   \end{tabular}
% \end{table}

% \begin{table}
%   \centering
%   \caption{}
%   \label{tab:newBorexResult}
%   \begin{tabular}{llr}
%     \toprule
%     Compared w/ &  Metric & $p$-val. \\
%     \midrule
%     RISE & F-meas. & 8.307474e-21 \\
%                 & ins. & 1.016050e-23 \\
%                 & del. & 0.887350 \\
%     Grad-CAM++ & F-meas. & 1.0 \\
%                 & ins. & 0.363604 \\
%                 & del. & 5.089895e-08 \\
%     \bottomrule
%   \end{tabular}
%   \end{table}

\else
\fi

\bibliographystyle{splncs04}
\bibliography{main}

\end{document}